\documentclass[10pt,twocolumn,letterpaper]{article}

\usepackage{cvpr}              

\newcommand{\TODO}[1]{\textbf{\color{red}[TODO: #1]}}
\renewcommand{\TODO}[1]{}

\usepackage[accsupp]{axessibility}

\usepackage{algorithm}
\usepackage{algorithmic}
\IfFileExists{dblfloatfix.sty}{\usepackage{dblfloatfix}}{}
\usepackage{tikz}
\usetikzlibrary{arrows.meta,calc,fit,positioning,shapes.geometric}
\usepackage{placeins}
\usepackage{array}
\usepackage{makecell}
\usepackage[T1]{fontenc}
\makeatletter
\@ifpackageloaded{lineno}{\setlength{\linenumbersep}{0.75cm}}{}
\makeatother
\makeatletter
\IfFileExists{titlesec.sty}{
  \usepackage{titlesec}
  \titleformat{\section}{\large\bfseries}{\thesection}{0.6em}{}
  \titleformat{\subsection}{\normalsize\bfseries}{\thesubsection}{0.6em}{}
}{
  \renewcommand\section{\@startsection{section}{1}{\z@}%
    {-3.5ex \@plus -1ex \@minus -.2ex}%
    {2.3ex \@plus .2ex}%
    {\normalfont\large\bfseries}}
  \renewcommand\subsection{\@startsection{subsection}{2}{\z@}%
    {-3.25ex\@plus -1ex \@minus -.2ex}%
    {1.5ex \@plus .2ex}%
    {\normalfont\normalsize\bfseries}}
}
\makeatother

\definecolor{cvprblue}{rgb}{0.21,0.49,0.74}
\usepackage[pagebackref,breaklinks,colorlinks,allcolors=cvprblue]{hyperref}

\def\paperID{15} 
\def\confName{CVPR}
\def\confYear{2026}

\title{\bfseries CatalogStitch: Dimension-Aware and Occlusion-Preserving\\Object Compositing for Catalog Image Generation}

\author{
Sanyam Jain\textsuperscript{1}\quad
Pragya Kandari\textsuperscript{1}\quad
Manit Singhal\textsuperscript{1}\quad
He Zhang\textsuperscript{1}\quad
Soo Ye Kim\textsuperscript{1}\\
\textsuperscript{1}Adobe\\
{\tt\small \{sanyjain, pkandari, manits, hezhan, sooyek\}@adobe.com}
}

\begin{document}
\maketitle


\begin{abstract}
\itshape
Generative object compositing methods have shown remarkable ability to seamlessly insert objects into scenes. However, when applied to real-world catalog image generation, these methods require tedious manual intervention: users must carefully adjust masks when product dimensions differ, and painstakingly restore occluded elements post-generation. We present \textbf{CatalogStitch}, a set of model-agnostic techniques that automate these corrections, enabling user-friendly content creation. Our \textit{dimension-aware mask computation} algorithm automatically adapts the target region to accommodate products with different dimensions; users simply provide a product image and background, without manual mask adjustments. Our \textit{occlusion-aware hybrid restoration} method guarantees pixel-perfect preservation of occluding elements, eliminating post-editing workflows. We additionally introduce \textbf{CatalogStitch-Eval}, a 58-example benchmark covering aspect-ratio mismatch and occlusion-heavy catalog scenarios, together with supplementary PDF and HTML viewers. We evaluate our techniques with three state-of-the-art compositing models (ObjectStitch, OmniPaint, and InsertAnything), demonstrating consistent improvements across diverse catalog scenarios. By reducing manual intervention and automating tedious corrections, our approach transforms generative compositing into a practical, human-friendly tool for production catalog workflows. Our project page is at \url{https://catalogstitch.github.io}.
\end{abstract}
    

\section{Introduction}
\label{sec:intro}

Product catalog imagery is essential for modern e-commerce and retail marketing. Creating high-quality catalog images traditionally requires expensive photoshoots, coordination of photographers and stylists, and weeks of production time. For businesses managing thousands of SKUs with seasonal updates, this process is prohibitively expensive and slow.

Recent advances in generative AI, particularly diffusion-based object compositing methods like ObjectStitch~\cite{song2023objectstitch}, Paint-by-Example~\cite{yang2023paint}, and AnyDoor~\cite{chen2024anydoor}, offer a promising foundation. These methods can seamlessly insert objects into scenes, handling geometry adjustment, color harmonization, and shadow generation in a unified framework. However, when we attempt to apply these methods to real-world catalog production workflows, critical gaps emerge that prevent practical deployment.

\subsection{The Gap Between Research and Production}

\begin{figure}[t]
    \centering
    \setlength{\tabcolsep}{2pt}
    \resizebox{\linewidth}{!}{%
    \begin{tabular}{@{}ccccc@{}}
        \textbf{Background} & \textbf{Product} & \textbf{Freeform} & \textbf{BBox} & \textbf{Dim-Aware} \\
        \includegraphics[width=2.5cm]{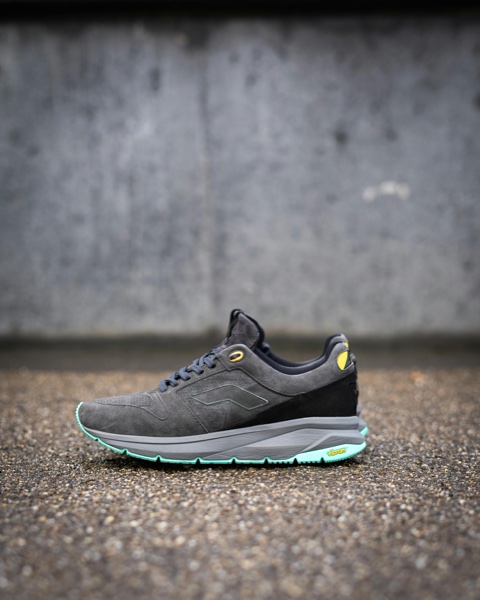} &
        \includegraphics[width=2.5cm]{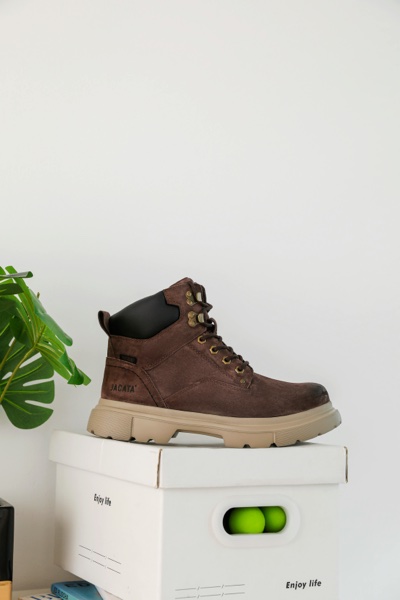} &
        \includegraphics[width=2.5cm]{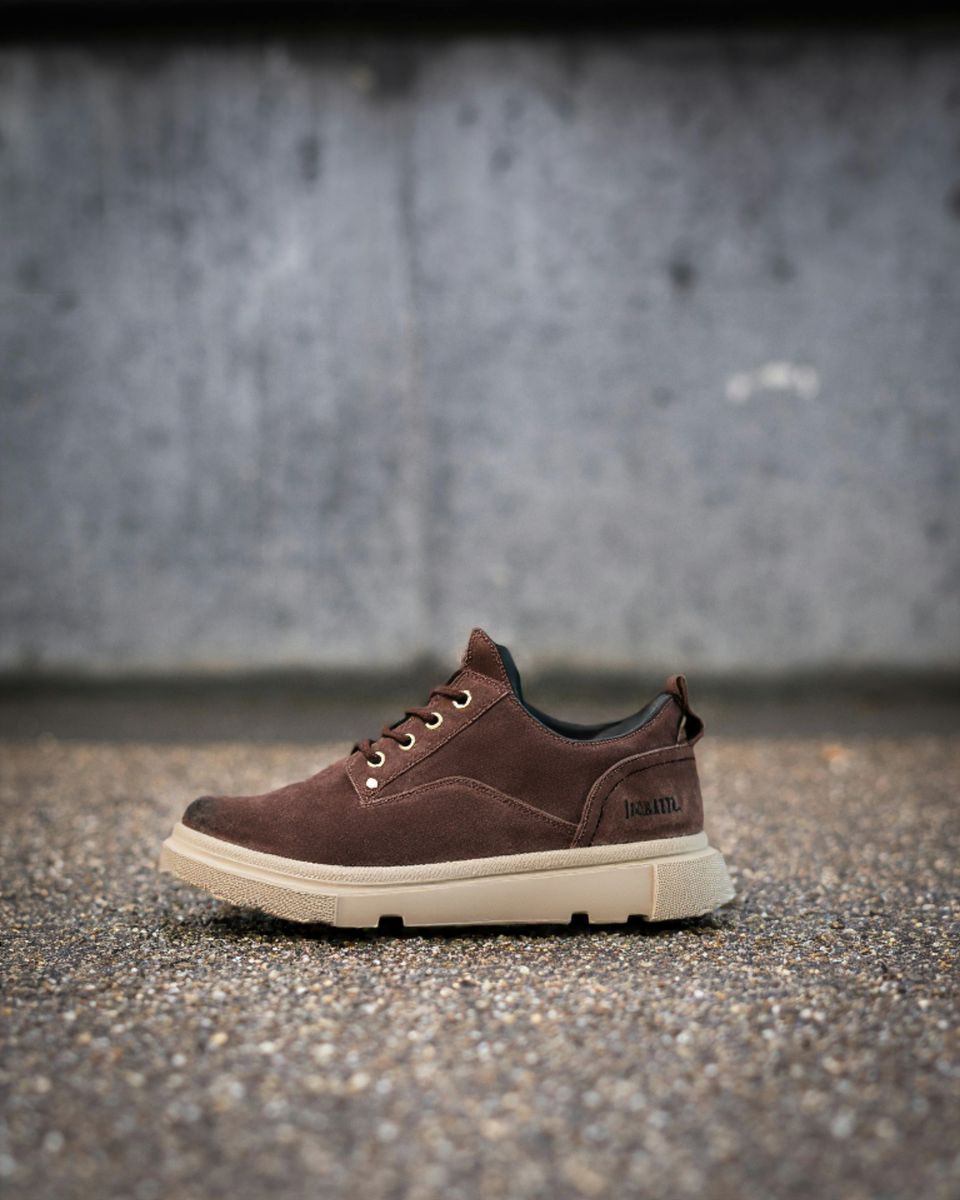} &
        \includegraphics[width=2.5cm]{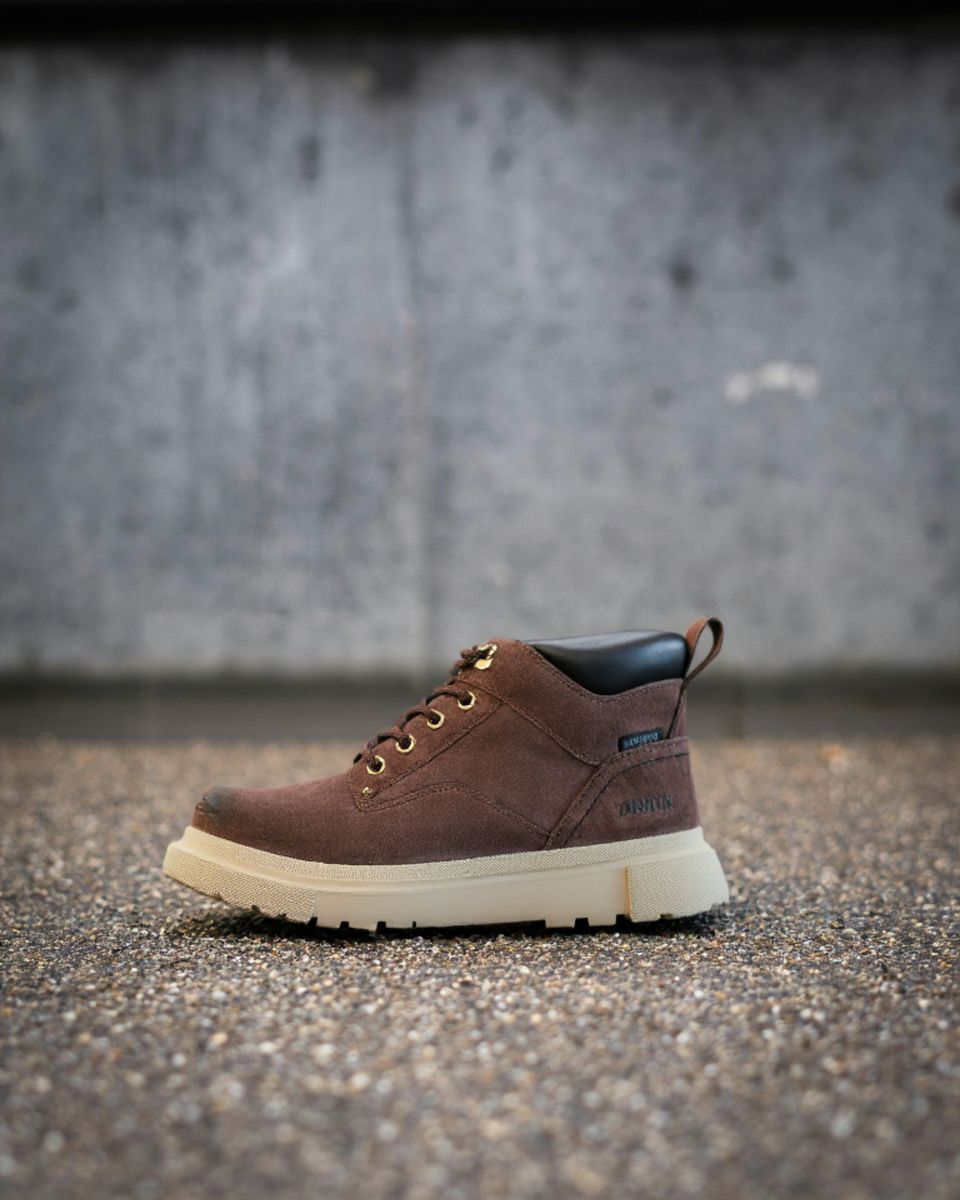} &
        \includegraphics[width=2.5cm]{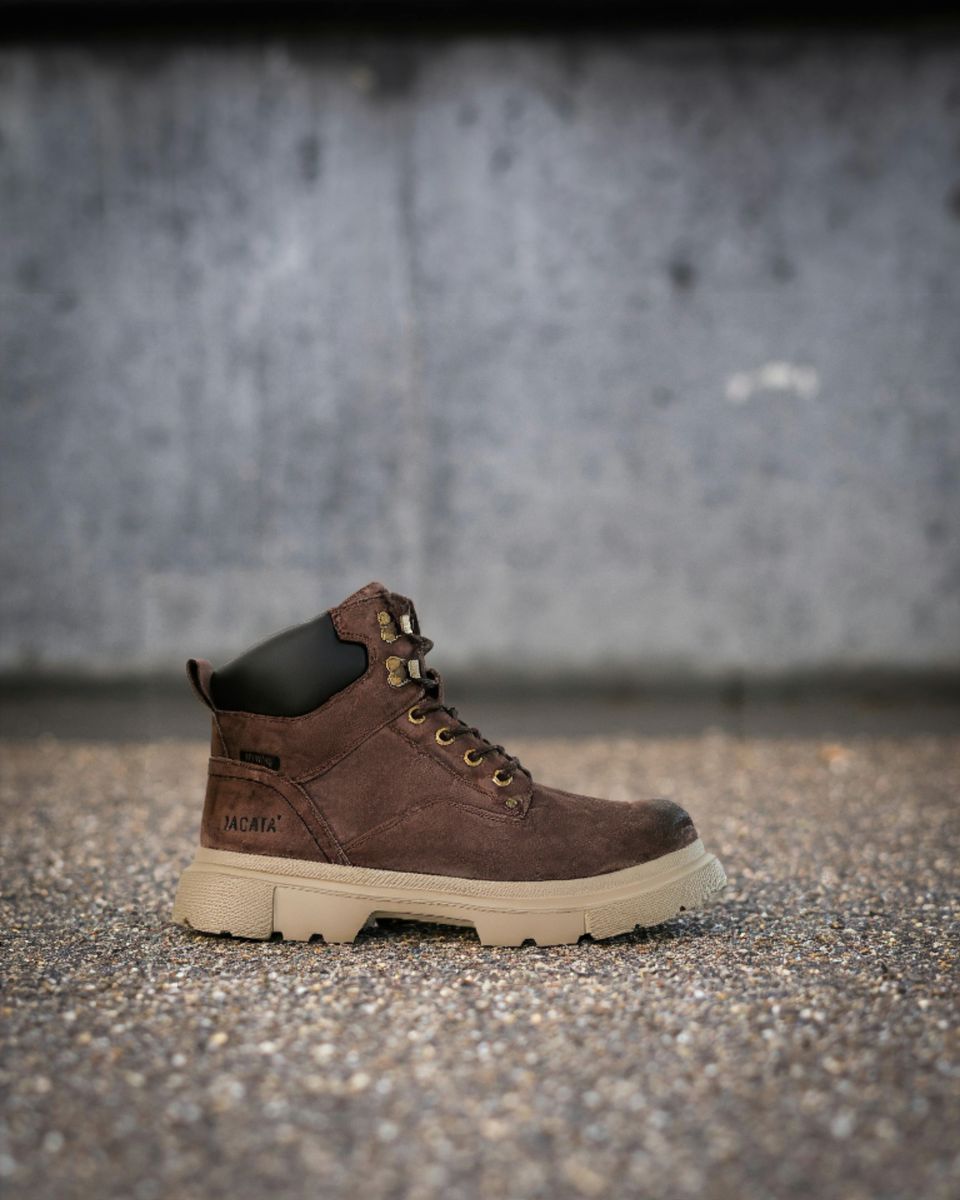} \\
    \end{tabular}%
    }
    \caption{\textbf{Challenge 1: Product Dimension Mismatch.} When replacing a product with different proportions, freeform and bounding box masks distort the product shape. Our dimension-aware mask preserves correct proportions.}
    \label{fig:challenge_dimension}
\end{figure}

Existing compositing methods excel in controlled settings where: (1) replacement objects have similar proportions to originals, and (2) scenes are simple with no overlapping elements. Real-world catalog imagery violates both assumptions:

\noindent\textbf{Challenge 1: Product Dimension Mismatch.}
When replacing one product with another, different proportions cause distortion when the new product is forced into the original target mask. For example, replacing a shoe in a scene with a differently proportioned shoe leads to shape distortion: the product is stretched or squashed to fit the original mask dimensions (Figure~\ref{fig:challenge_dimension}).

\begin{figure}[t]
    \centering
    \setlength{\tabcolsep}{2pt}
    \resizebox{\linewidth}{!}{%
    \begin{tabular}{@{}ccccc@{}}
        \textbf{Background} & \textbf{Product} & \textbf{Occluders} & \textbf{Before Restore} & \textbf{After Restore} \\
        \includegraphics[width=2.5cm]{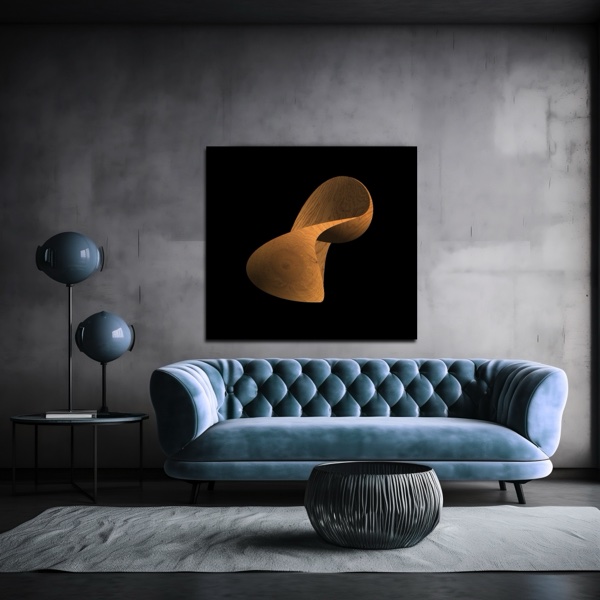} &
        \includegraphics[width=2.5cm]{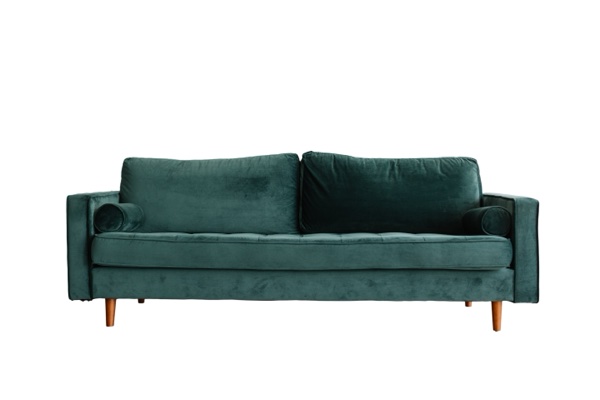} &
        \includegraphics[width=2.5cm]{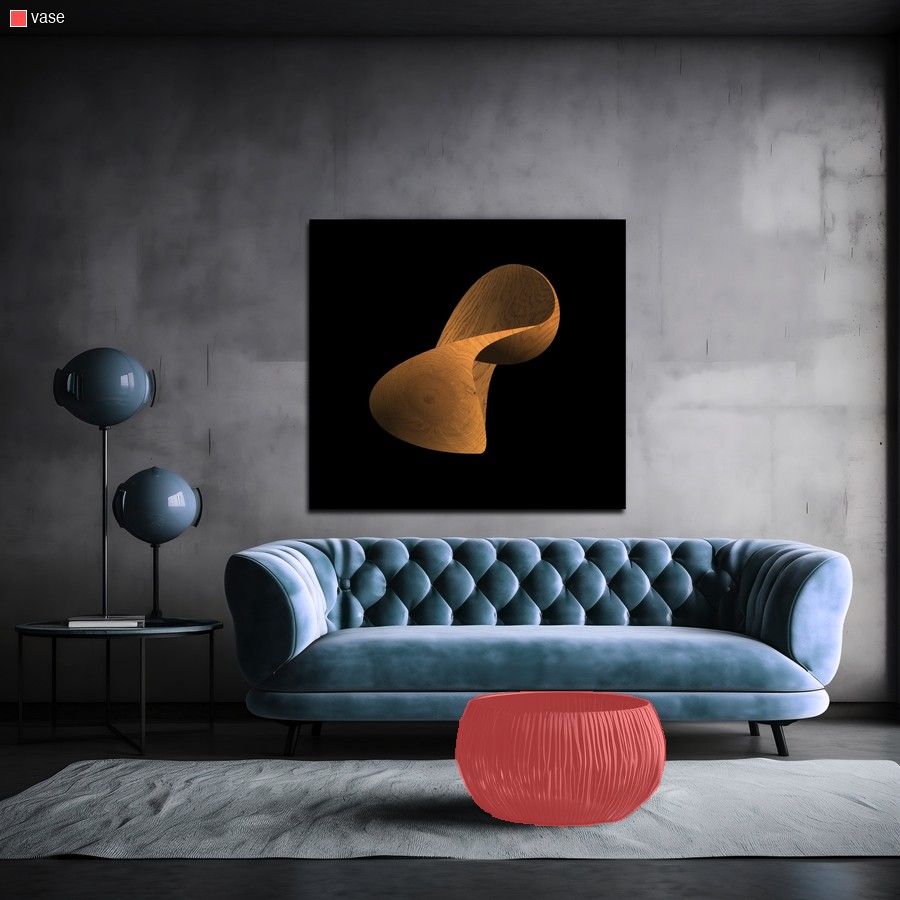} &
        \includegraphics[width=2.5cm]{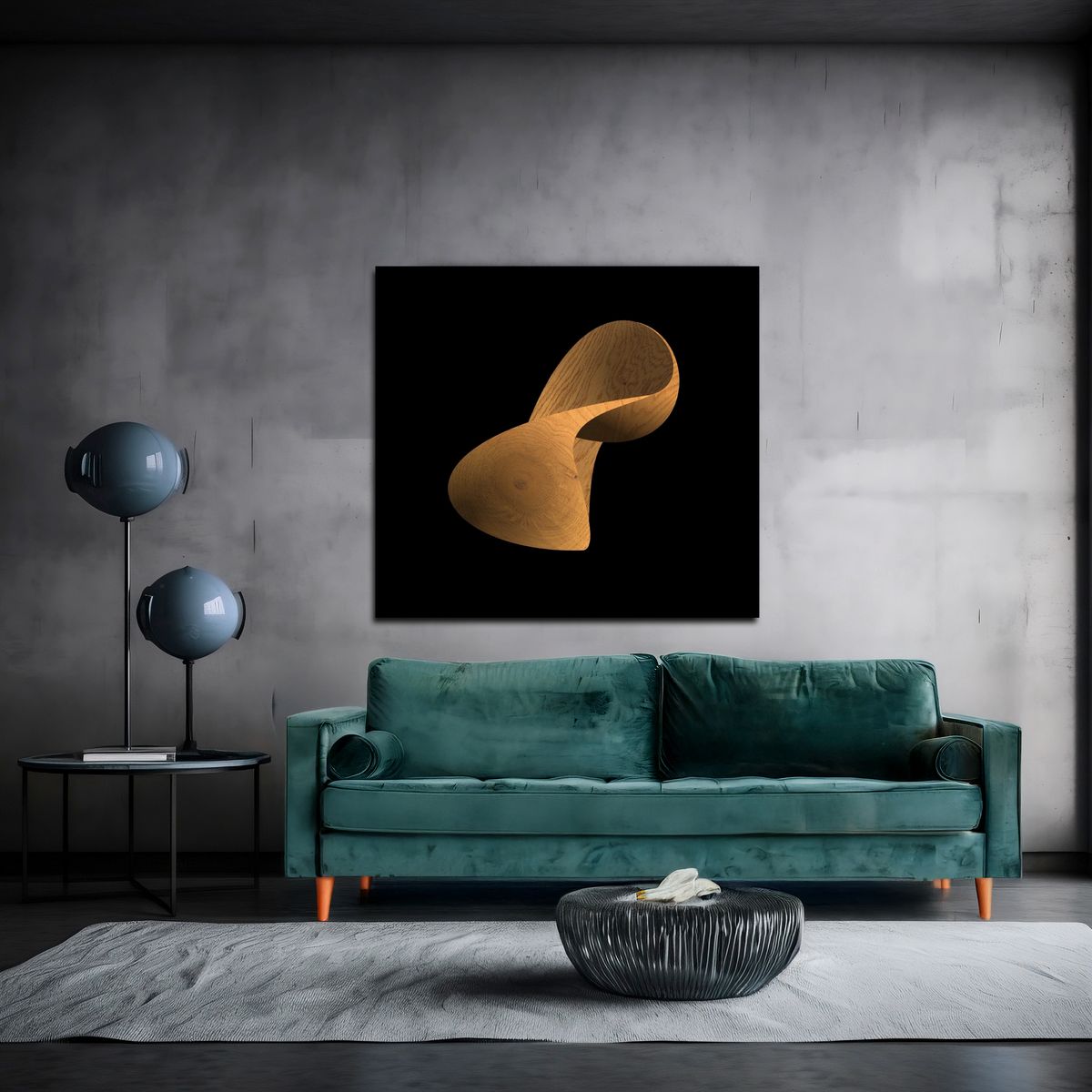} &
        \includegraphics[width=2.5cm]{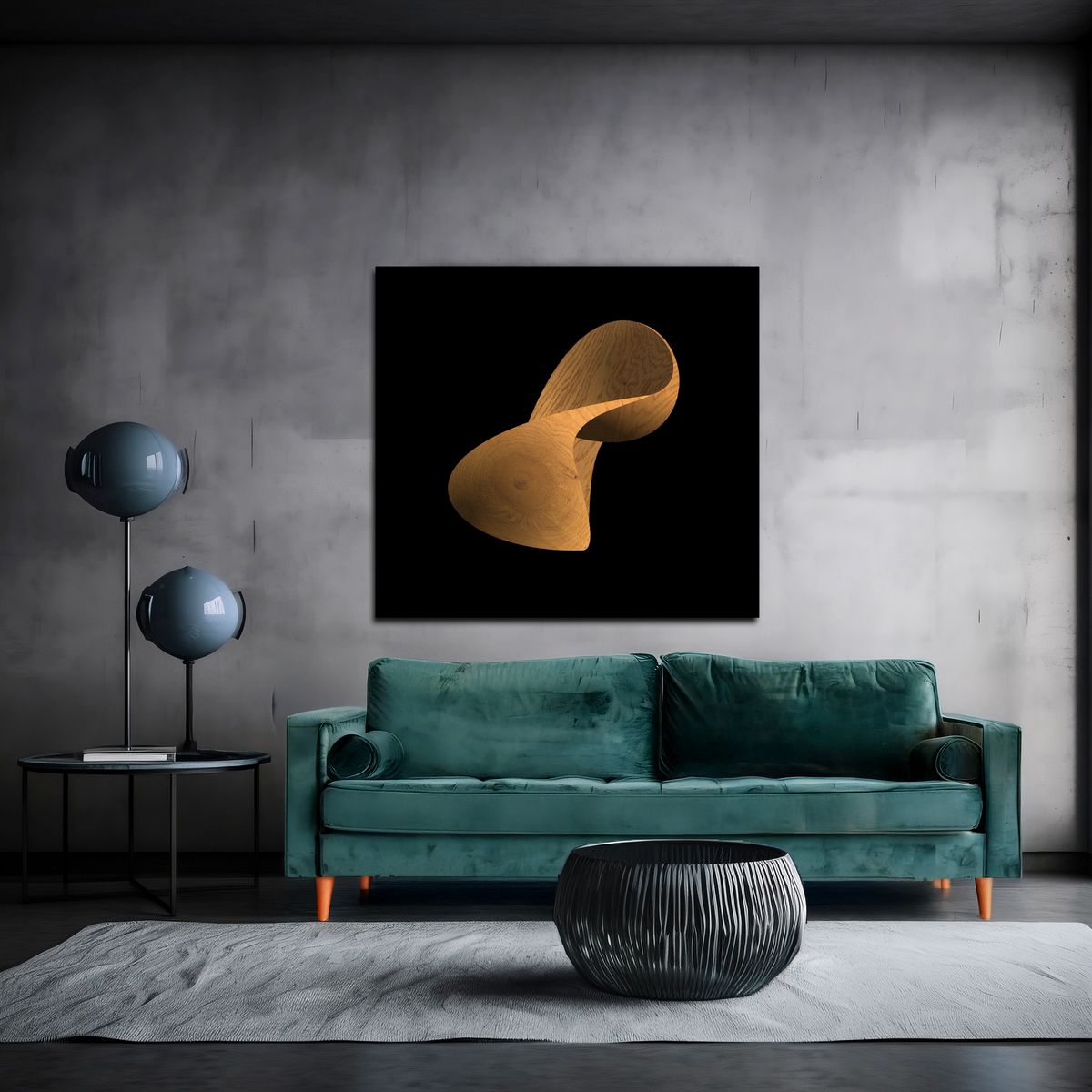} \\
    \end{tabular}%
    }
    \caption{\textbf{Challenge 2: Occlusion Destruction.} Foreground elements (table/decor items) occluding the target product are corrupted during compositing. Our hybrid restoration preserves them pixel-perfectly.}
    \label{fig:challenge_occlusion}
\end{figure}

\noindent\textbf{Challenge 2: Occlusion Destruction.}
Real catalog images feature products partially occluded by foreground elements. When the target region is replaced, these overlapping objects are destroyed or distorted by current methods. In Figure~\ref{fig:challenge_occlusion}, foreground furniture in front of the target sofa is corrupted during compositing.

\subsection{Our Contributions}

We present \textbf{CatalogStitch}, a collection of model-agnostic techniques that extend existing compositing methods for real-world catalog image generation:

\begin{enumerate}
    \item \textbf{Dimension-Aware Mask Computation:} An algorithm that automatically computes an expanded target mask accommodating the new product's aspect ratio while staying centered on the original location. This preserves correct product proportions without manual intervention.
    
    \item \textbf{Occlusion-Aware Hybrid Restoration:} A five-step approach that detects occluding objects, caches their exact pixels, applies generative inpainting to remove occluders and expose a clean background, performs generative compositing on that clean background, then restores the cached occluder pixels---guaranteeing pixel-perfect preservation of overlapping elements.
    
    \item \textbf{CatalogStitch-Eval Benchmark:} A 58-example evaluation benchmark for catalog compositing, with 35 dimension-mismatch scenes, 23 occlusion scenes, masks, source metadata, PDF result summaries, and supplementary HTML viewers for rapid inspection. The benchmark package is publicly released\footnote{\url{https://github.com/adobe-research/CatalogStitch}}.

    \item \textbf{Cross-Model Evaluation:} An empirical study across three state-of-the-art compositing models (ObjectStitch, OmniPaint, InsertAnything) showing that the proposed preprocessing and postprocessing steps consistently improve realism and structural fidelity without changing the underlying generators.
\end{enumerate}

Our key insight is that a \textit{hybrid design}, combining generative AI for tasks such as harmonization and shadow generation with deterministic operations for tasks requiring exact fidelity, delivers both the flexibility of AI compositing and the reliability required for production workflows.

\subsection{Enabling User-Friendly Compositing}

Current compositing workflows demand significant manual effort: users must carefully craft masks when product dimensions differ, experiment with scaling parameters, and painstakingly restore occluded elements that get destroyed during generation. This technical overhead prevents non-expert users from leveraging generative compositing and slows down production pipelines even for experienced operators.

Our techniques shift the human-AI interaction from low-level pixel manipulation to high-level creative decisions. Users simply provide a product image and a background scene, while the system automatically handles mask adaptation and occlusion preservation. This \textit{simplified interaction model} is crucial for democratizing AI-powered content creation and allowing marketers and designers to generate catalog imagery without specialized technical skills.

Furthermore, our model-agnostic design means these user-friendly capabilities can be added to any existing compositing model, future-proofing workflows as underlying generative models continue to improve.


\section{Related Work}
\label{sec:related}

\noindent\textbf{Generative Image Compositing.}
Recent diffusion-based methods have achieved impressive results in object compositing. ObjectStitch~\cite{song2023objectstitch} uses a content adaptor to inject object features into a pre-trained inpainting model, handling geometry adjustment and harmonization jointly. Paint-by-Example~\cite{yang2023paint} conditions image generation on reference images using CLIP embeddings. AnyDoor~\cite{chen2024anydoor} employs ID tokens and high-frequency maps to preserve object identity during compositing. ControlCom~\cite{zhang2023controlcom} adds explicit control over object location and transformation. More recently, OmniPaint~\cite{yu2025omnipaint} and InsertAnything~\cite{song2025insertanything} have pushed the boundaries of compositing quality. Our work complements these methods by addressing preprocessing (mask computation) and postprocessing (occlusion restoration) challenges that are orthogonal to the core compositing model.

\noindent\textbf{Object Placement and Harmonization.}
Prior work has studied where to place objects in scenes~\cite{zhang2020learning} and how to harmonize composited objects with backgrounds~\cite{cong2020dovenet,cong2022high}. The OPA dataset~\cite{liu2021opa} provides benchmarks for evaluating object placement plausibility. Unlike these works that focus on placement selection or post-hoc harmonization, we address the mask adaptation problem that arises when the replacement object has fundamentally different dimensions than the original.

\noindent\textbf{Amodal Completion and Occlusion.}
Amodal completion methods~\cite{zhan2020self,ozguroglu2024pix2gestalt} aim to predict the full extent of partially occluded objects. Layered image representations~\cite{zhang2024transparent} decompose scenes into separate object layers. While related, our occlusion handling takes the opposite approach: rather than predicting occluded content, we \textit{preserve} existing occluder pixels that would otherwise be destroyed during compositing. This deterministic approach guarantees fidelity that generative methods cannot match.


\section{Method}
\label{sec:method}

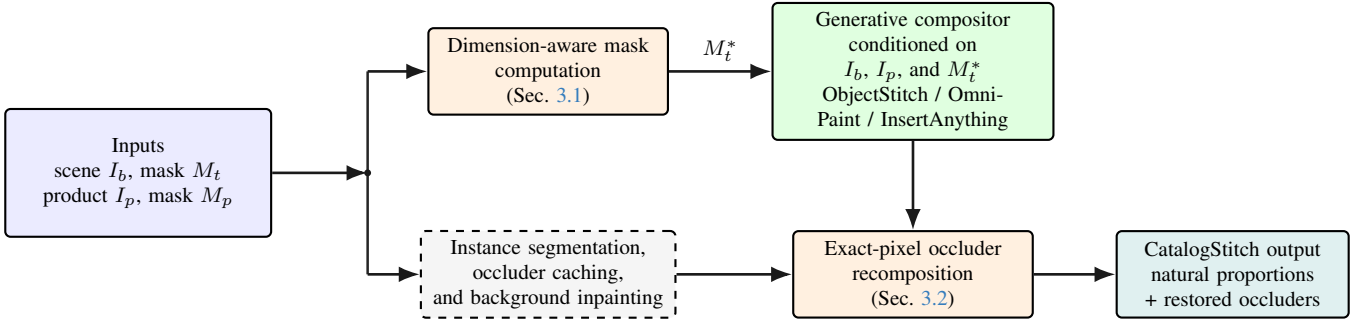
\begin{figure*}[t]
    \centering
    \begin{tikzpicture}[
        font=\footnotesize,
        >=Latex,
        arrow/.style={draw=black!90, line width=1.0pt, -{Latex[length=2.8mm,width=2.1mm]}, rounded corners=2pt},
        link/.style={draw=black!85, line width=0.95pt},
        input/.style={draw, rounded corners=2pt, thick, fill=blue!8, minimum height=1.7cm, text width=3.3cm, align=center},
        process/.style={draw, rounded corners=2pt, thick, fill=orange!12, minimum height=1.15cm, text width=2.95cm, align=center},
        model/.style={draw, rounded corners=2pt, thick, fill=green!12, minimum height=1.2cm, text width=3.5cm, align=center},
        output/.style={draw, rounded corners=2pt, thick, fill=teal!12, minimum height=1.15cm, text width=2.85cm, align=center},
        aux/.style={draw, rounded corners=2pt, dashed, thick, fill=gray!8, minimum height=1.0cm, text width=3.15cm, align=center}
    ]
        \node[input] (assets) at (0, -0.45) {Inputs\\scene $I_b$, mask $M_t$\\product $I_p$, mask $M_p$};
        \node[process] (dim) at (5.45, 0.9) {Dimension-aware mask\\computation\\(Sec.~\ref{sec:dimension_mask})};
        \node[aux] (occ) at (5.45, -1.8) {Instance segmentation,\\occluder caching,\\and background inpainting};
        \node[model] (compose) at (10.3, 0.9) {Generative compositor\\conditioned on $I_b$, $I_p$, and $M_t^*$\\ObjectStitch / OmniPaint / InsertAnything};
        \node[process] (restore) at (10.3, -1.8) {Exact-pixel occluder\\recomposition\\(Sec.~\ref{sec:occlusion})};
        \node[output] (final) at (14.55, -1.8) {CatalogStitch output\\natural proportions\\+ restored occluders};

        \coordinate (junction) at (3.05, -0.45);
        \coordinate (topbranch) at (3.05, 0.9);
        \coordinate (botbranch) at (3.05, -1.8);

        \draw[arrow] (assets.east) -- (junction);
        \draw[link] (topbranch) -- (botbranch);
        \fill[black!90] (junction) circle (1.3pt);
        \draw[arrow] (topbranch) -- (dim.west);
        \draw[arrow] (botbranch) -- (occ.west);
        \draw[arrow] (dim.east) -- node[above] {$M_t^*$} (compose.west);
        \draw[arrow] (compose.south) -- (restore.north);
        \draw[arrow] (occ.east) -- (restore.west);
        \draw[arrow] (restore.east) -- (final.west);
    \end{tikzpicture}
    \caption{\textbf{CatalogStitch overview.} Two lightweight, model-agnostic modules wrap a baseline compositor. The target mask is adapted to the replacement product ratio; occluders are segmented, cached, and inpainted away before compositing, then restored from the original pixels at the final step.}
    \label{fig:catalogstitch_overview}
\end{figure*}

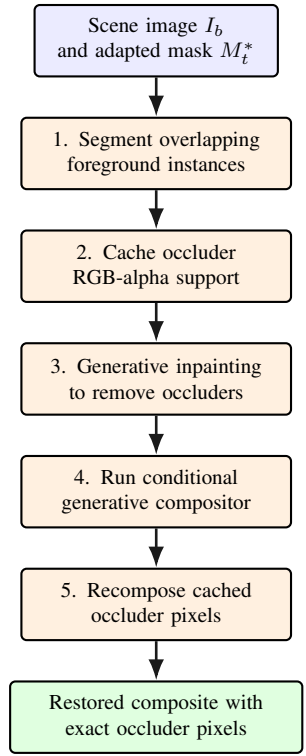
\begin{figure*}[!t]
    \centering
    \begin{tikzpicture}[
        font=\footnotesize,
        >=Latex,
        arrow/.style={draw=black!90, line width=1.0pt, -{Latex[length=2.8mm,width=2.1mm]}, rounded corners=2pt},
        state/.style={draw, rounded corners=2pt, thick, fill=blue!8, minimum height=0.95cm, text width=2.75cm, align=center},
        process/.style={draw, rounded corners=2pt, thick, fill=orange!12, minimum height=0.95cm, text width=3.05cm, align=center},
        decision/.style={draw, diamond, aspect=2.2, thick, fill=yellow!18, align=center, inner sep=1.3pt, text width=2.6cm},
        output/.style={draw, rounded corners=2pt, thick, fill=green!12, minimum height=0.95cm, text width=3.15cm, align=center}
    ]
        \node[font=\bfseries\small] at (-4.2, 2.5) {Dimension-Aware Mask Computation};
        \node[state] (din) at (-4.2, 1.55) {Input masks\\$M_t$ and $M_p$};
        \node[process, text width=3.4cm] (measure) at (-4.2, 0.1) {Measure bbox size, centroid,\\and aspect ratios $AR_t$, $AR_p$};
        \node[decision] (check) at (-4.2, -1.55) {$|AR_t - AR_p| \le \tau$?};
        \node[process, text width=3.2cm] (expand) at (-0.4, -1.55) {Expand bbox around\\the original centroid};
        \node[output, text width=3.4cm] (dout) at (-4.2, -3.2) {Output mask $M_t^*$\\(original or expanded)};

        \draw[arrow] (din.south) -- (measure.north);
        \draw[arrow] (measure.south) -- (check.north);
        \draw[arrow] (check.south) -- node[left, pos=0.52] {yes} (dout.north);
        \draw[arrow] (check.east) -- node[above, pos=0.48] {no} (expand.west);
        \draw[arrow] (expand.south) |- (dout.east);

        \node[font=\bfseries\small] at (4.35, 2.5) {Occlusion-Aware Hybrid Restoration};
        \node[state, text width=3.0cm] (oin) at (4.35, 1.55) {Scene image $I_b$\\and adapted mask $M_t^*$};
        \node[process, text width=3.35cm] (detect) at (4.35, 0.05) {1. Segment overlapping\\foreground instances};
        \node[process, text width=3.35cm] (copy) at (4.35, -1.45) {2. Cache occluder\\RGB-alpha support};
        \node[process, text width=3.35cm] (inpaint) at (4.35, -2.95) {3. Generative inpainting\\to remove occluders};
        \node[process, text width=3.35cm] (gen) at (4.35, -4.45) {4. Run conditional\\generative compositor};
        \node[process, text width=3.35cm] (restore) at (4.35, -5.95) {5. Recompose cached\\occluder pixels};
        \node[output, text width=3.65cm] (ofinal) at (4.35, -7.45) {Restored composite with\\exact occluder pixels};

        \draw[arrow] (oin.south) -- (detect.north);
        \draw[arrow] (detect.south) -- (copy.north);
        \draw[arrow] (copy.south) -- (inpaint.north);
        \draw[arrow] (inpaint.south) -- (gen.north);
        \draw[arrow] (gen.south) -- (restore.north);
        \draw[arrow] (restore.south) -- (ofinal.north);
    \end{tikzpicture}
    \caption{\textbf{Module-level flow charts.} Left: dimension-aware mask computation preserves the original placement when the mismatch is small and otherwise expands the target region around the original centroid. Right: occlusion-aware restoration detects overlapping entities, caches their exact pixels, applies generative inpainting to remove occluders and expose a clean background, runs the compositor on the clean background, and finally restores the cached occluder pixels to preserve original geometry, texture, and scene-consistent lighting.}
    \label{fig:module_flows}
\end{figure*}

We present two complementary techniques that can be applied as preprocessing and postprocessing steps around any generative compositing model. Our design philosophy prioritizes \textit{user simplicity}: the user provides only a product image and a background scene, and our system automatically handles mask adaptation and occlusion preservation. Figure~\ref{fig:catalogstitch_overview} summarizes the full end-to-end pipeline, while Figure~\ref{fig:module_flows} breaks down the two core modules.

\subsection{Dimension-Aware Mask Computation}
\label{sec:dimension_mask}

\noindent\textbf{Problem Formulation.}
Given a background image $I_b$ with a target region mask $M_t$ indicating where to place the new product, and a product image $I_p$ with product mask $M_p$, we need to compute an optimal target mask $M_t^*$ that accommodates the product's proportions.

Let $(w_t, h_t)$ be the bounding box dimensions of $M_t$ and $(w_p, h_p)$ be the dimensions of $M_p$. The aspect ratios are:
\begin{equation}
    AR_t = \frac{w_t}{h_t}, \quad AR_p = \frac{w_p}{h_p}
\end{equation}

When $|AR_t - AR_p| > \tau$ (we use $\tau = 0.06$), the product cannot fit naturally into the target region without distortion. Figure~\ref{fig:module_flows} (left) illustrates the decision process.

\noindent\textbf{Algorithm.}
Our dimension-aware mask computation proceeds as follows:

\begin{enumerate}
    \item \textbf{Extract Centroids:} Compute the centroid $(c_x, c_y)$ of the original target region $M_t$ to preserve spatial positioning.
    
    \item \textbf{Compare Aspect Ratios:} If $|AR_t - AR_p| < \tau$, the original mask is sufficient; use $M_t^* = M_t$.
    
    \item \textbf{Compute Optimal Dimensions:} When aspect ratios differ significantly:
    \begin{align}
        h^* &= h_t \\
        w^* &= h^* \cdot AR_p
    \end{align}
    If $w^* < w_t$, we instead anchor on width:
    \begin{align}
        w^* &= w_t \\
        h^* &= w^* / AR_p
    \end{align}
    
    \item \textbf{Center on Original Location:}
    \begin{align}
        x^* &= \max(0, c_x - w^*/2) \\
        y^* &= \max(0, c_y - h^*/2)
    \end{align}
    
    \item \textbf{Generate Mask:} Create $M_t^*$ as a rectangular mask with computed dimensions, clipped to image boundaries.
\end{enumerate}

This ensures the dimension-aware mask expands to fit the new product's proportions while staying centered on the original target location, maintaining scene coherence.

\subsection{Occlusion-Aware Hybrid Restoration via Exact Occluder Compositing}
\label{sec:occlusion}

\noindent\textbf{Problem Formulation.}
In professionally styled catalog images, products are often partially occluded by decorative elements (plants, vases, side tables). Let $\mathcal{O} = \{O_1, O_2, ..., O_k\}$ be the set of objects whose masks overlap with the target region $M_t$. When generative compositing replaces the target region, these occluders are typically destroyed or distorted.

\noindent\textbf{Key Insight.}
Occluding objects are already observed in the original image with exact RGB values, scene-consistent illumination, shadows, and high-frequency structure. Instead of regenerating these regions, we cache their visible support and recompose them deterministically after generation. Figure~\ref{fig:module_flows} (right) summarizes the four restoration stages.

\noindent\textbf{Five-Stage Hybrid Restoration Pipeline.}

\textbf{Step 1: Segment Overlapping Foreground Instances.}
Using entity segmentation on the background image $I_b$, we identify candidate foreground instances and their masks. For each detected entity $E_i$ with bounding box $B_i$, we compute overlap with the target region:
\begin{equation}
    \text{IoU}(B_i, B_t) = \frac{|B_i \cap B_t|}{|B_i \cup B_t|}
\end{equation}
If $\text{IoU} > \tau_{occ}$ (we use $\tau_{occ} = 0.01$ to capture minimal overlaps), $E_i$ is marked as an occluder.

\textbf{Step 2: Cache Exact Occluder Support.}
For each occluding entity $O_i$:
\begin{itemize}
    \item Extract the exact visible pixel support from the original image: $P_i = I_b \odot M_{O_i}$
    \item Cache $(P_i, M_{O_i}, \text{coords}_i)$ for deterministic reuse
    \item No filtering or synthesis is applied; the cached region is preserved verbatim
\end{itemize}

\textbf{Step 3: Generative Inpainting of Occluder Regions.}
With occluder positions cached, we remove their visual interference from the background by applying generative inpainting within the union of occluder masks $\bigcup_i M_{O_i}$:
\begin{equation}
    I_b^{\text{inp}} = \text{Inpaint}\!\left(I_b,\;\bigcup\nolimits_i M_{O_i}\right)
\end{equation}
This yields a clean background $I_b^{\text{inp}}$ in which the target region is fully unoccluded, allowing the compositor to integrate the replacement product without residual occluder interference.

\textbf{Step 4: Run Conditional Generative Compositing.}
Apply the compositing model (ObjectStitch, OmniPaint, or InsertAnything) using the inpainted background and the dimension-aware mask $M_t^*$ from Section~\ref{sec:dimension_mask}:
\begin{equation}
    I_{comp} = \text{Composite}(I_b^{\text{inp}}, I_p, M_t^*, M_p)
\end{equation}
Because occluders are absent from $I_b^{\text{inp}}$, the compositor generates a clean product integration with no occluder artefacts.

\textbf{Step 5: Recompose Cached Occluders.}
Alpha-composite the cached occluder regions onto the generated result:
\begin{equation}
    I_{final} = I_{comp} \odot (1 - \bigcup_i M_{O_i}) + \sum_i P_i
\end{equation}
The reinserted occluders preserve their original geometry, texture, lighting, and shadows by construction.


\section{Experiments}
\label{sec:experiments}

\subsection{Implementation Details}

All techniques are purely inference-time wrappers requiring no model fine-tuning or additional training. The dimension-aware mask computation is a closed-form geometric calculation; occlusion detection uses a single forward pass of EntitySeg~\cite{qi2022open}; recomposition uses pixel-level alpha blending. The combined overhead is negligible relative to the generative compositing step. All experiments use images at native resolution (typically 1024$\times$1024 or higher).
The threshold $\tau{=}0.06$ controls when dimension-aware mask computation is applied, balancing unnecessary mask expansion against missed adaptation cases.
The IoU threshold $\tau_{occ}{=}0.01$ controls occluder detection sensitivity, capturing minimal occlusions without including non-overlapping nearby objects.

\begin{figure*}[!tp]
    \centering
    \setlength{\tabcolsep}{0.5pt}
    \renewcommand{\arraystretch}{1.0}
    \scriptsize
    \fontsize{6.5}{7.2}\selectfont

    \newcommand{\dimleftbgwidth}{2.0cm}
    \newcommand{\dimleftbg}[1]{\includegraphics[width=\dimleftbgwidth,height=\dimleftbgwidth,keepaspectratio]{#1}}
    \newcommand{\dimleftbgmask}[1]{\includegraphics[width=\dimleftbgwidth,height=\dimleftbgwidth,keepaspectratio]{#1}}
    \newcommand{\dimrightimg}[1]{\includegraphics[width=\dimleftbgwidth,height=\dimleftbgwidth,keepaspectratio]{#1}}
    \newcommand{\dimoutwidth}{2.0cm}
    \newcommand{\dimoutimg}[1]{\includegraphics[width=\dimoutwidth,height=\dimoutwidth,keepaspectratio]{#1}}
    \newsavebox{\dimbgboxa}
    \newsavebox{\dimbgboxb}
    \newsavebox{\dimbgboxc}
    \sbox{\dimbgboxa}{\dimleftbg{figures/dimension/6/bg.jpg}}
    \sbox{\dimbgboxb}{\dimleftbg{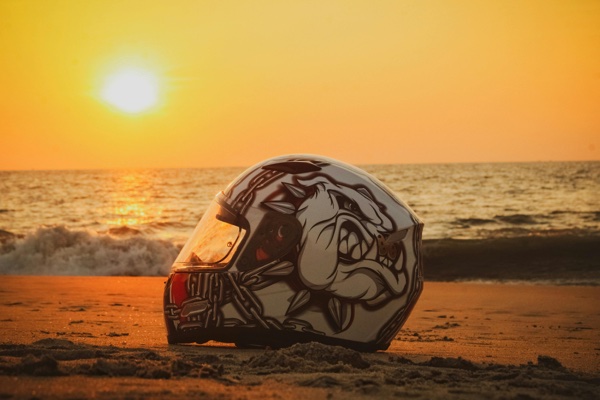}}
    \sbox{\dimbgboxc}{\dimleftbg{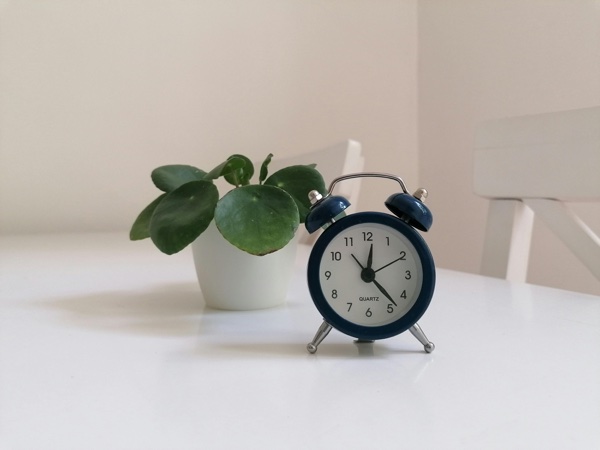}}

    \begin{tabular}{@{}c@{\hspace{0.8mm}}c@{\hspace{4mm}}c@{}}
        &
            {\fontsize{5.5}{6.5}\selectfont
            \begin{tabular}[b]{@{}*{3}{>{\centering\arraybackslash}m{\dimleftbgwidth}}@{}}
                \textbf{Background Image} & \textbf{Product Image} & \textbf{Overlayed Masks}
            \end{tabular}
        }
        &
            {\fontsize{5.5}{6.5}\selectfont
            \begin{tabular}[b]{@{}*{3}{>{\centering\arraybackslash}m{\dimoutwidth}}@{}}
                \makecell{\textbf{Result with}\\\textbf{Freeform Mask}} & \makecell{\textbf{Result with}\\\textbf{BBox}} & \makecell{\textbf{Result with Dim-}\\\textbf{Aware Mask (Ours)}}
            \end{tabular}
        } \\[-0.3mm]
        \textbf{(a)} &
            \begin{tabular}[c]{@{}*{3}{>{\centering\arraybackslash}m{\dimleftbgwidth}}@{}}
                \usebox{\dimbgboxa} &
                \dimrightimg{figures/dimension/6/obj.jpg} &
                \dimleftbgmask{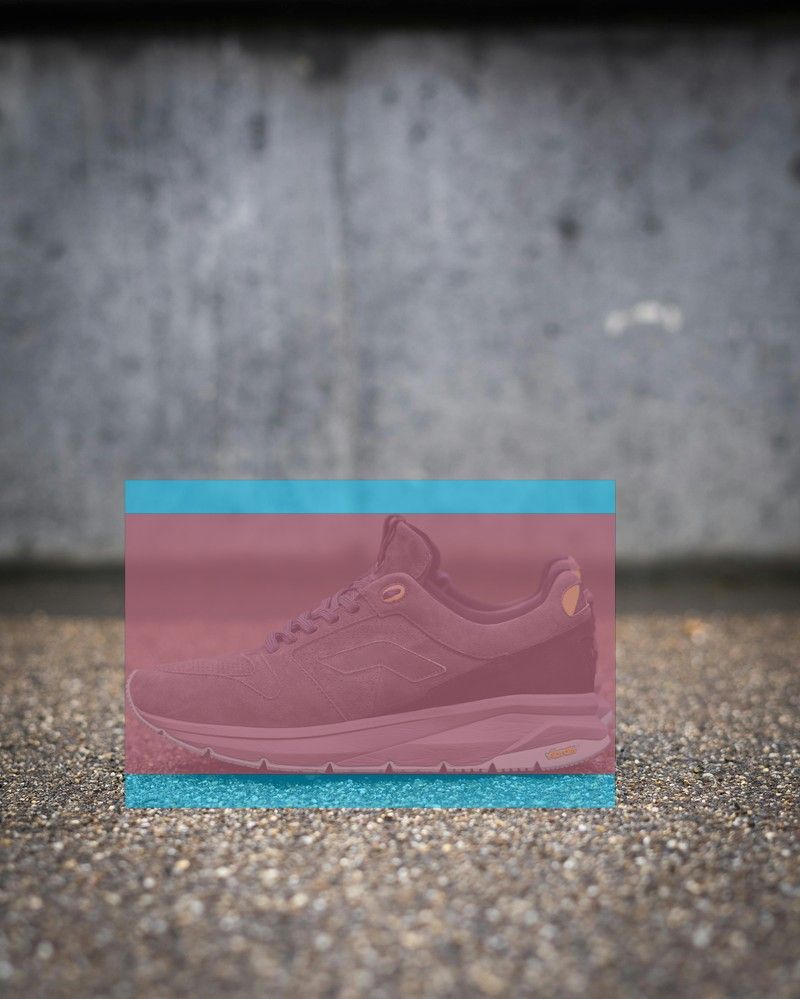} \\
            \end{tabular}
        &
            \begin{tabular}[c]{@{}*{3}{>{\centering\arraybackslash}m{\dimoutwidth}}@{}}
                \multicolumn{3}{c}{\fontsize{5.5}{6.5}\selectfont\textit{OmniPaint}} \\[0.2mm]
                \dimoutimg{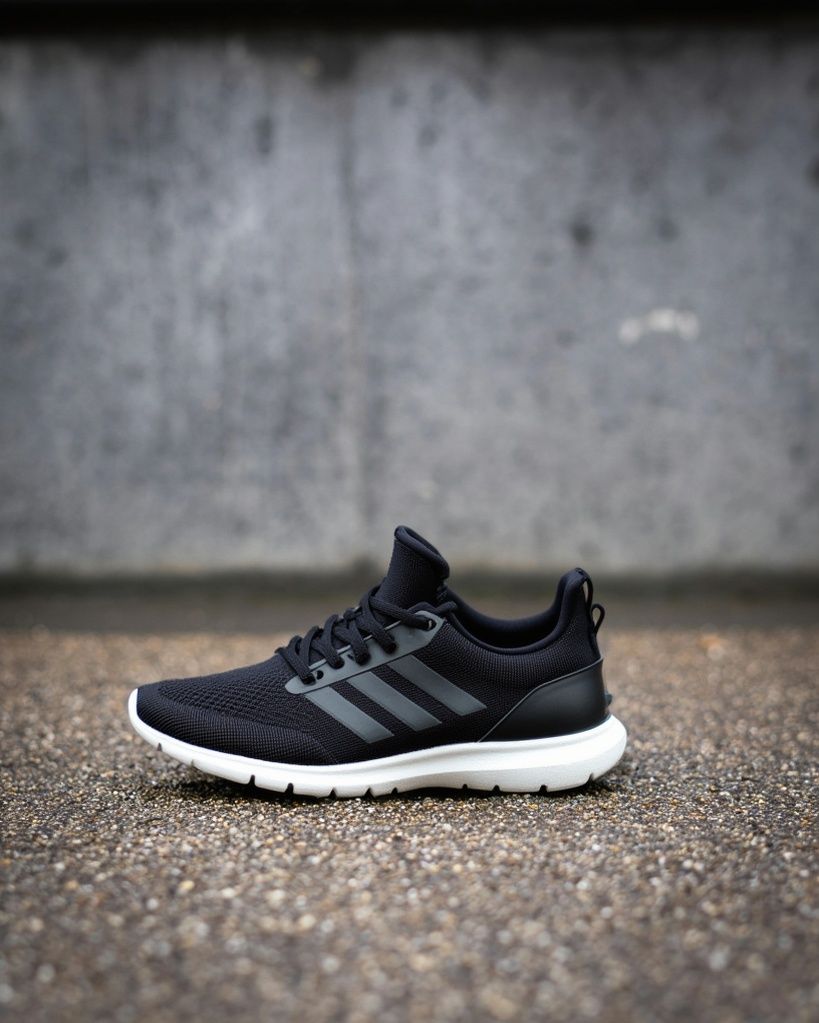} & \dimoutimg{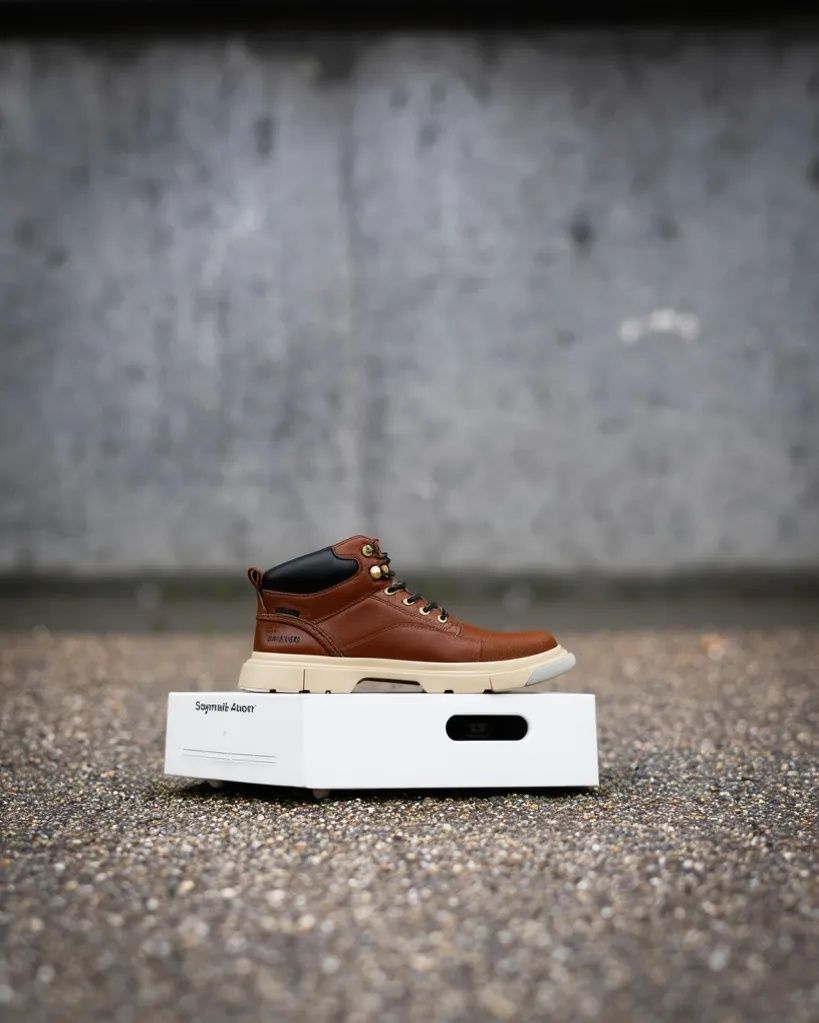} & \dimoutimg{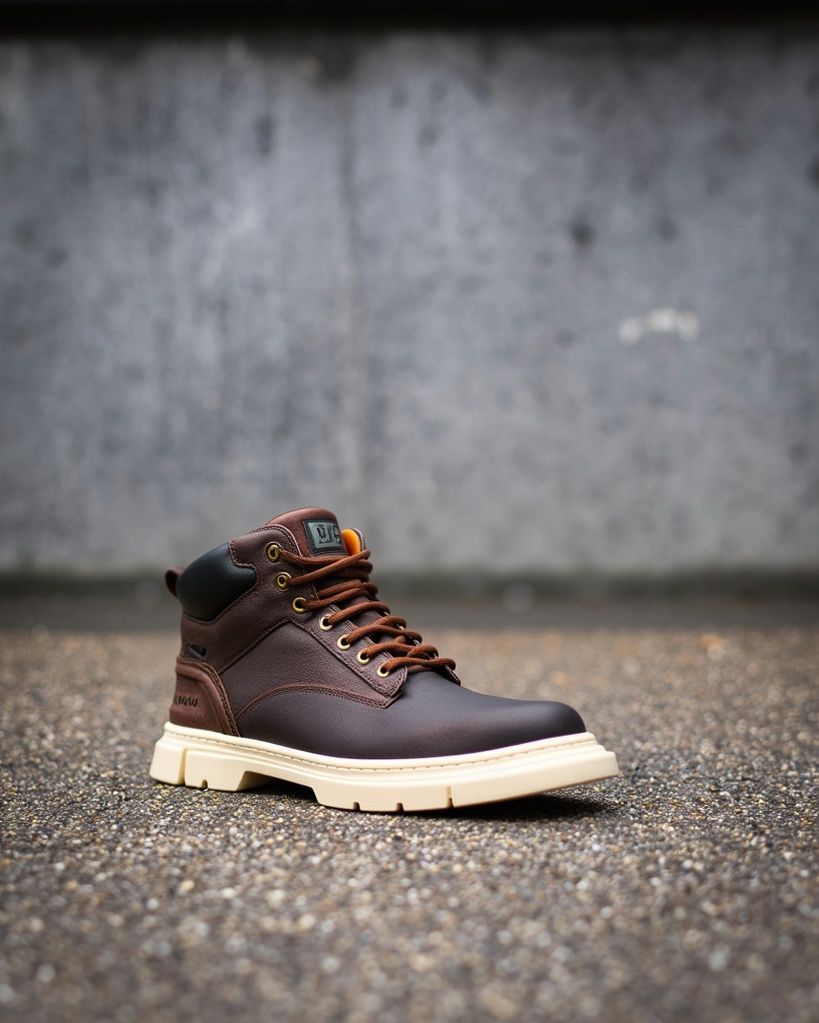} \\[0.2mm]
                \multicolumn{3}{c}{\fontsize{5.5}{6.5}\selectfont\textit{InsertAnything}} \\[0.2mm]
                \dimoutimg{figures/dimension/6/insertAnything_nonOptimal.jpg} & \dimoutimg{figures/dimension/6/insertAnything_bbox.jpg} & \dimoutimg{figures/dimension/6/insertAnything_optimal.jpg} \\[0.2mm]
                \multicolumn{3}{c}{\fontsize{5.5}{6.5}\selectfont\textit{ObjectStitch}} \\[0.2mm]
                \dimoutimg{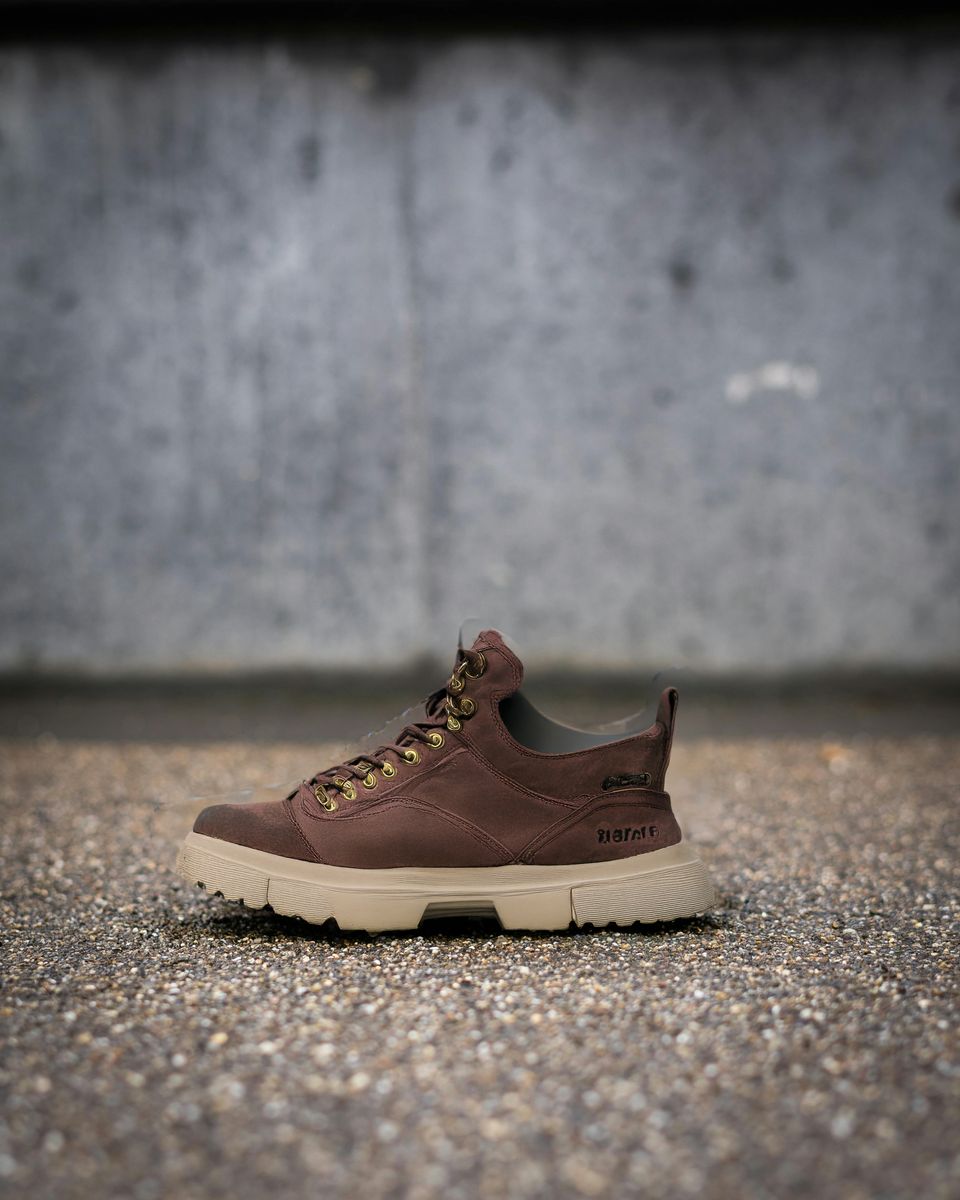} & \dimoutimg{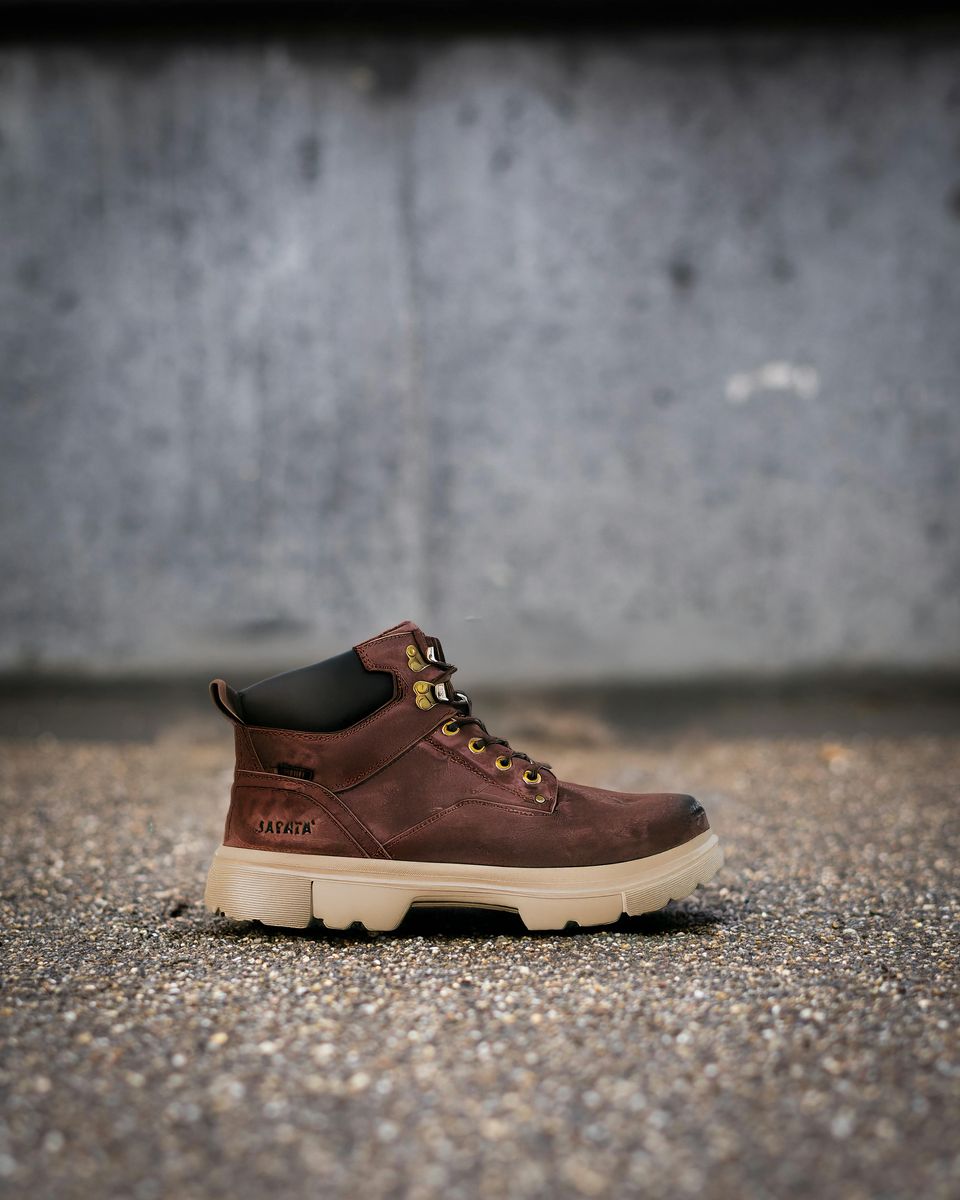} & \dimoutimg{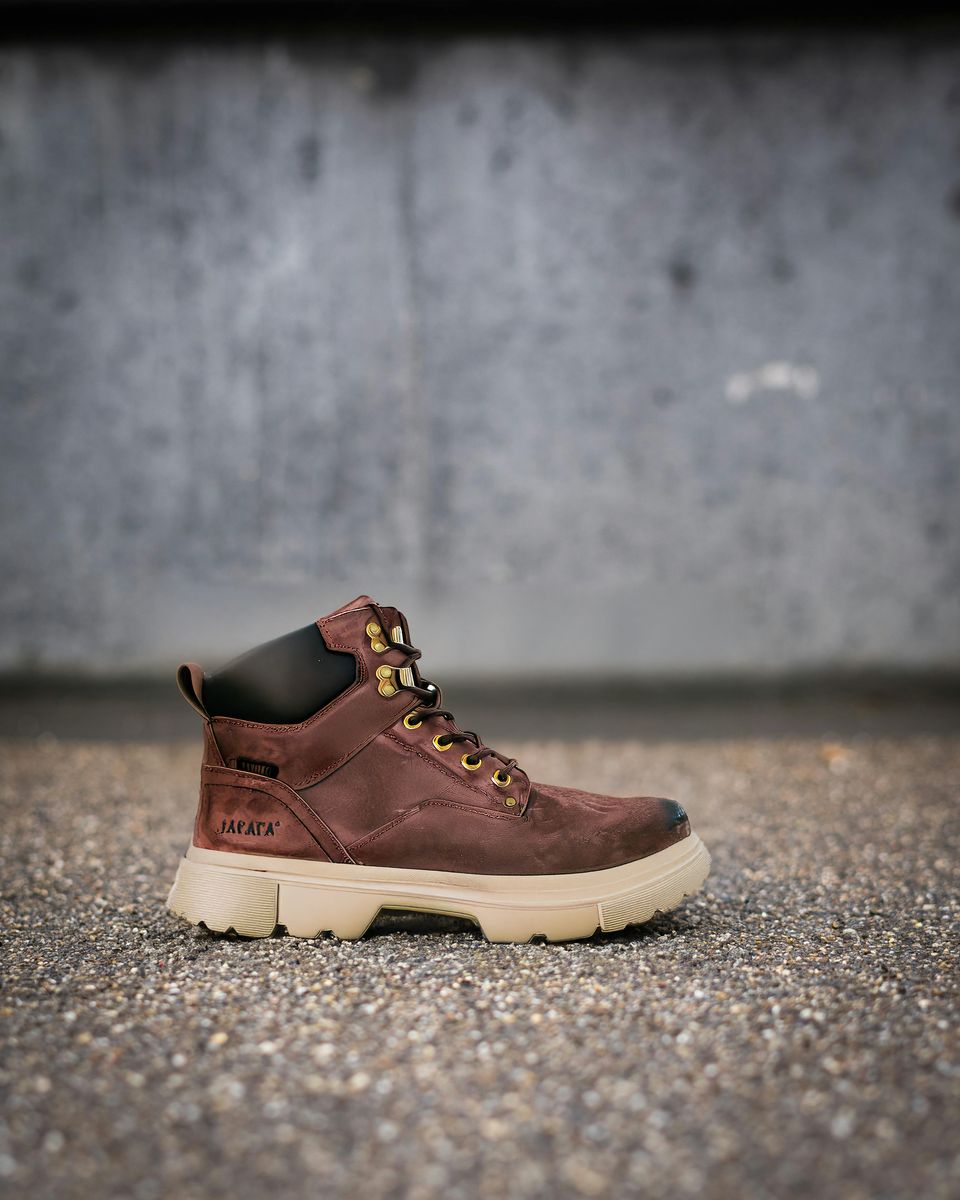} \\
            \end{tabular}
        \\[0.5mm]
        \textbf{(b)} &
            \begin{tabular}[c]{@{}*{3}{>{\centering\arraybackslash}m{\dimleftbgwidth}}@{}}
                \usebox{\dimbgboxb} &
                \dimrightimg{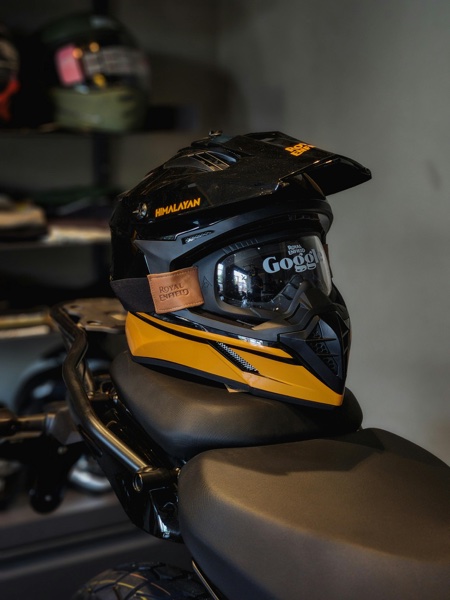} &
                \dimleftbgmask{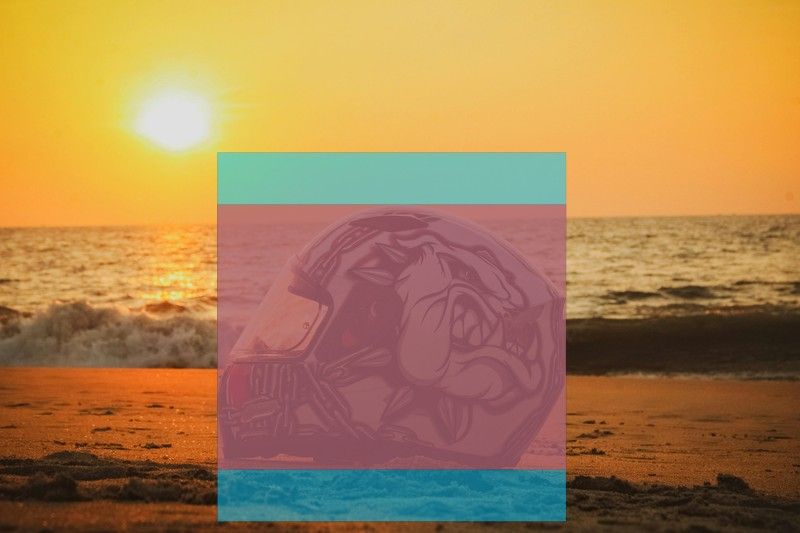} \\
            \end{tabular}
        &
            \begin{tabular}[c]{@{}*{3}{>{\centering\arraybackslash}m{\dimoutwidth}}@{}}
                \multicolumn{3}{c}{\fontsize{5.5}{6.5}\selectfont\textit{OmniPaint}} \\[0.2mm]
                \dimoutimg{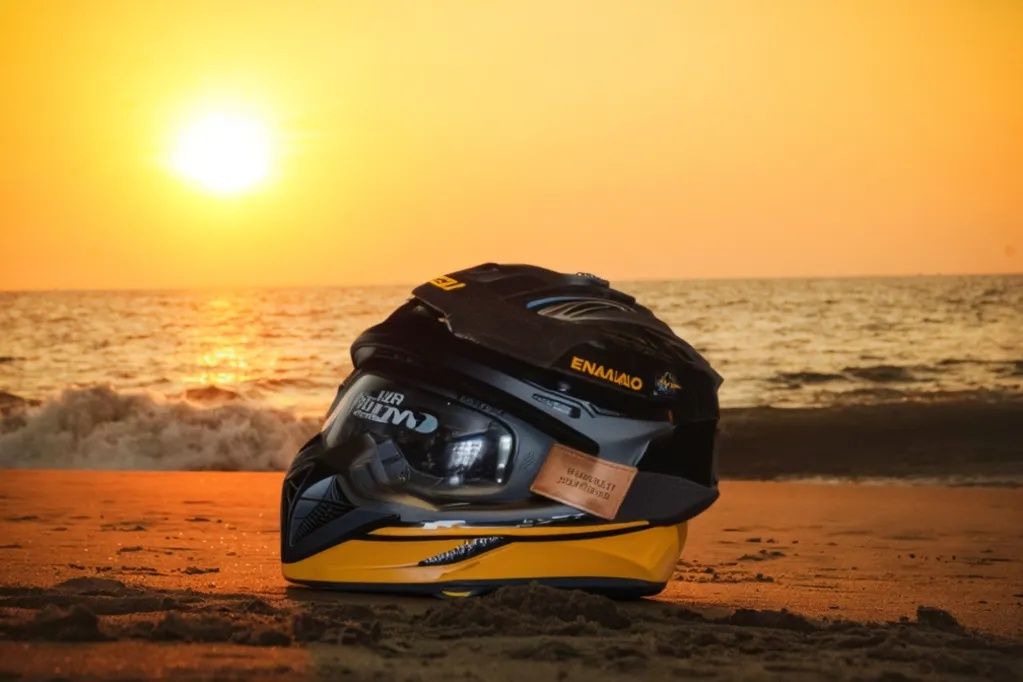} & \dimoutimg{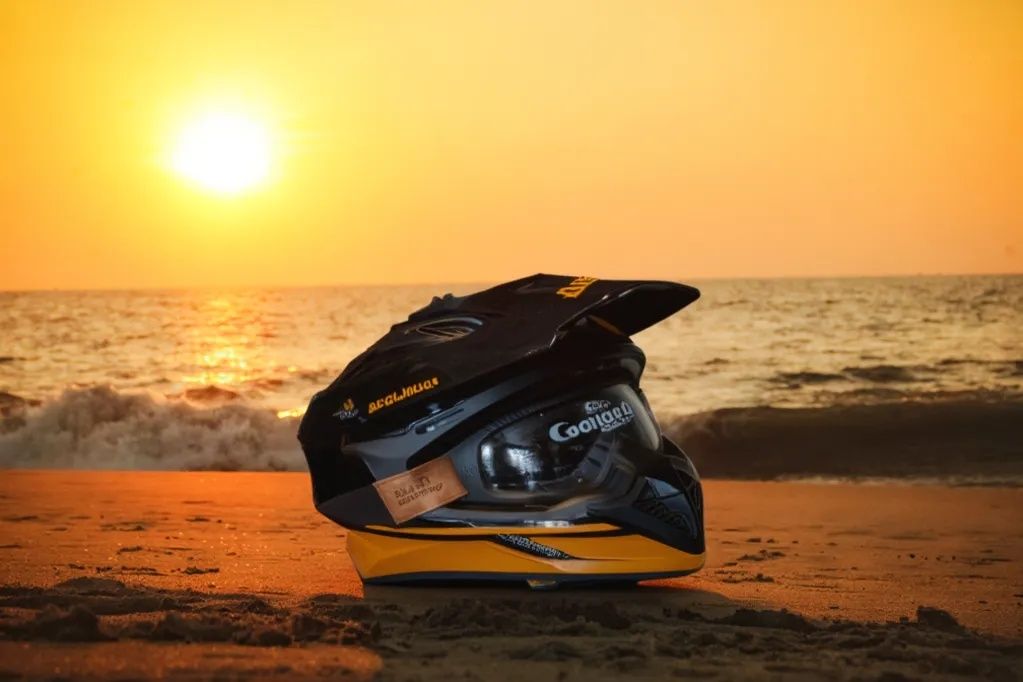} & \dimoutimg{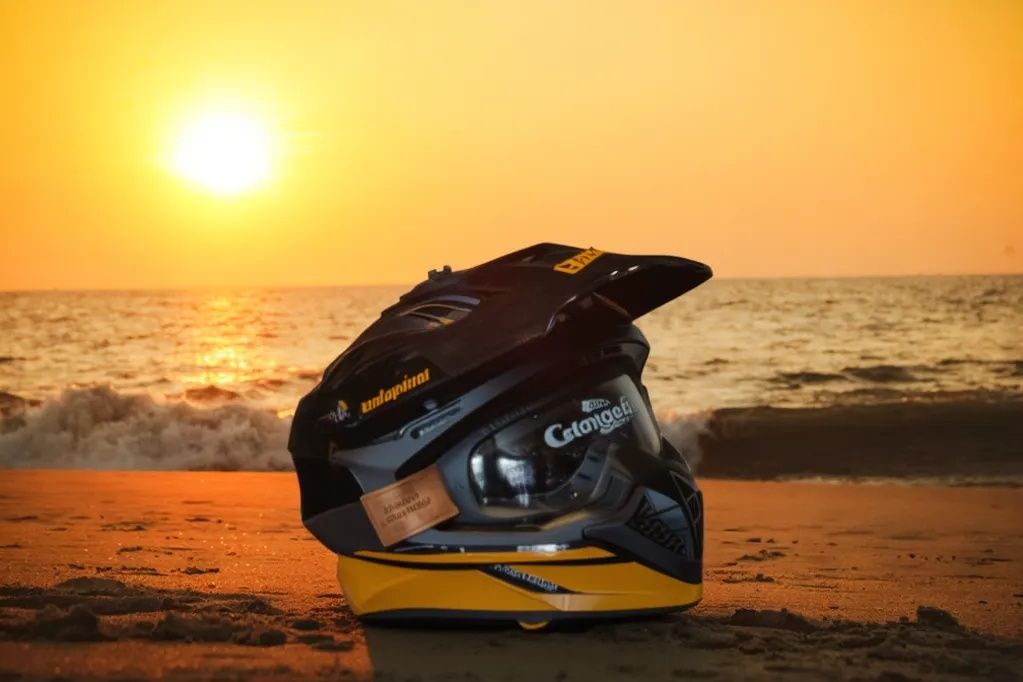} \\[0.2mm]
                \multicolumn{3}{c}{\fontsize{5.5}{6.5}\selectfont\textit{InsertAnything}} \\[0.2mm]
                \dimoutimg{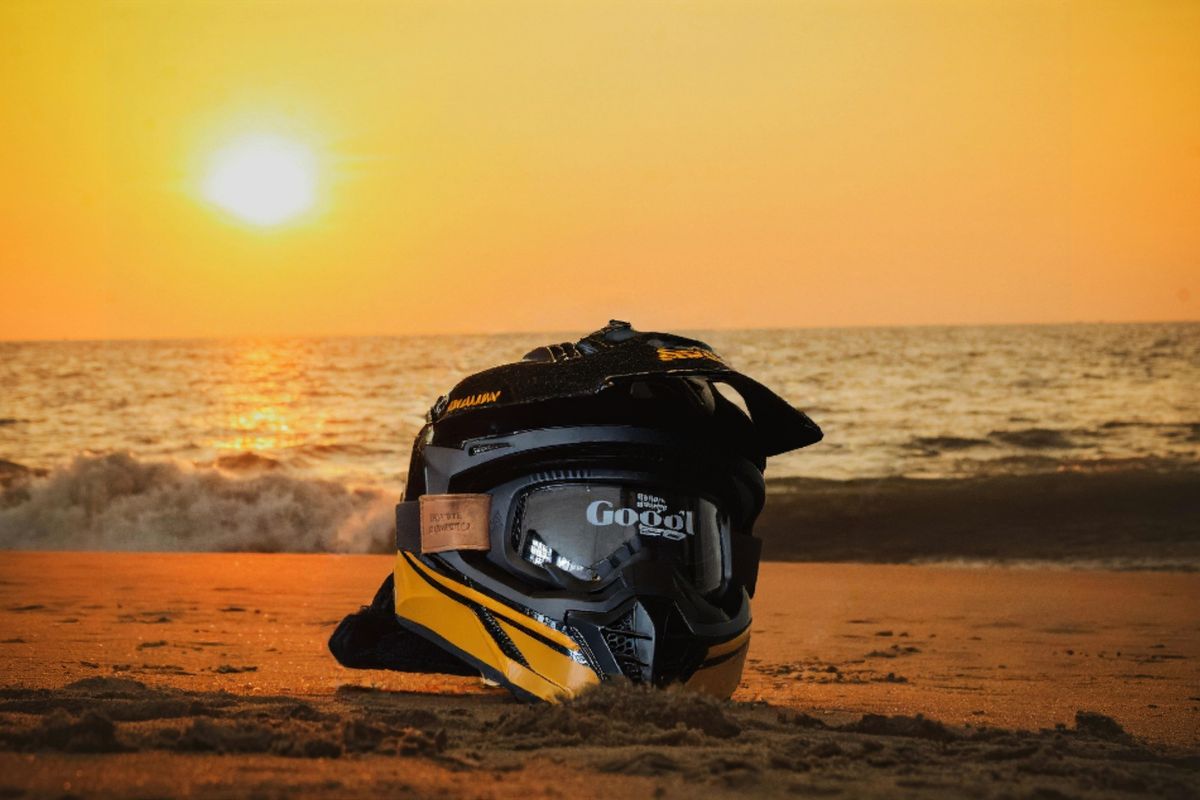} & \dimoutimg{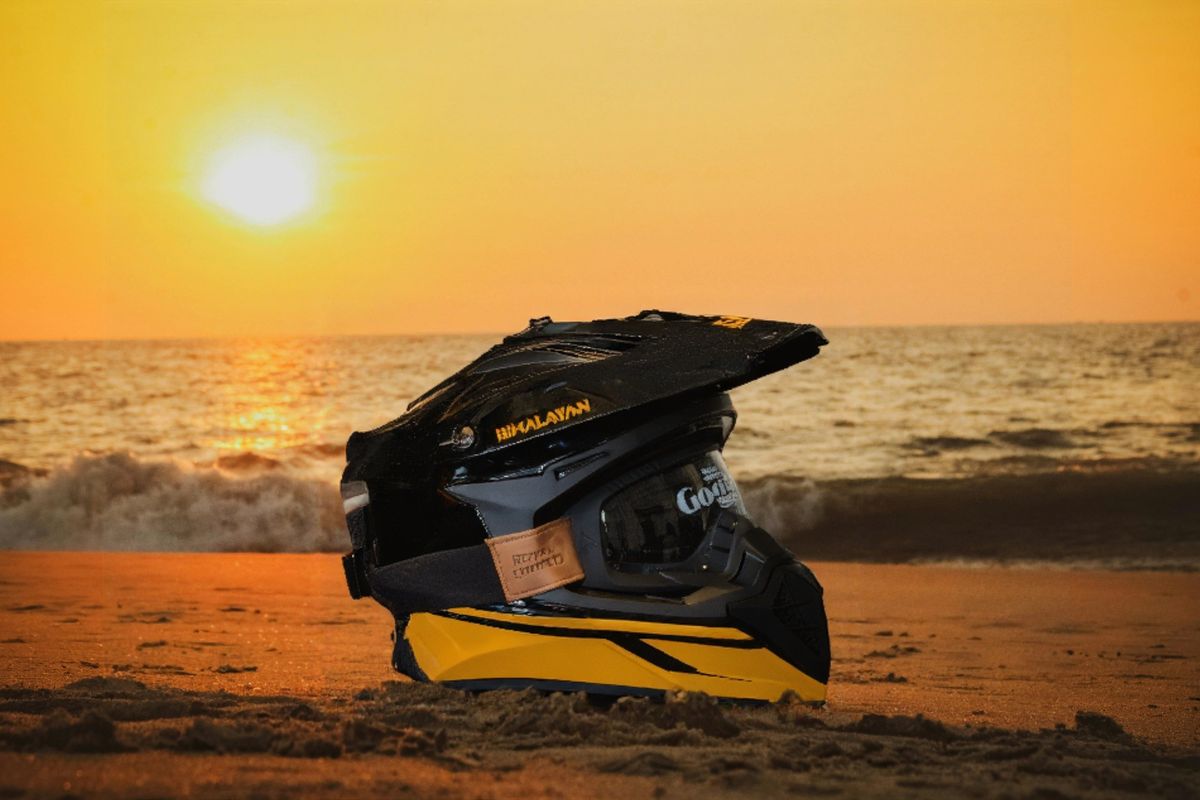} & \dimoutimg{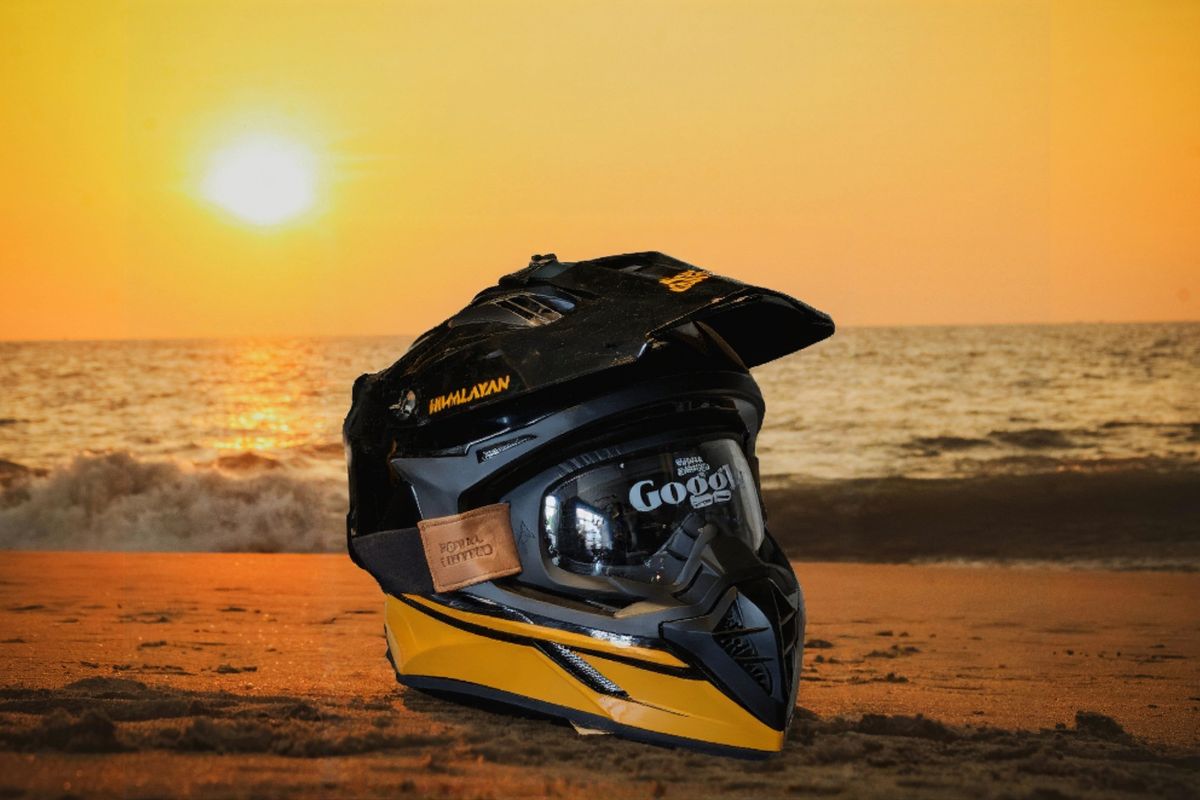} \\[0.2mm]
                \multicolumn{3}{c}{\fontsize{5.5}{6.5}\selectfont\textit{ObjectStitch}} \\[0.2mm]
                \dimoutimg{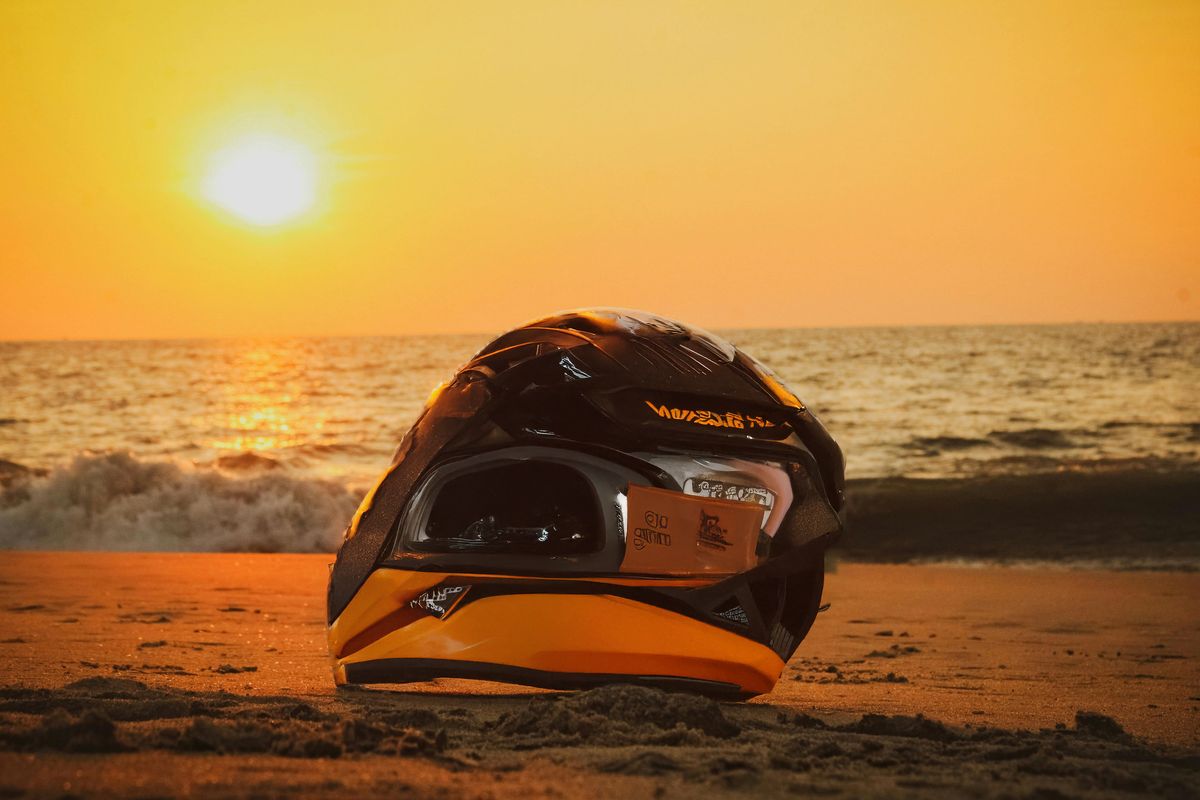} & \dimoutimg{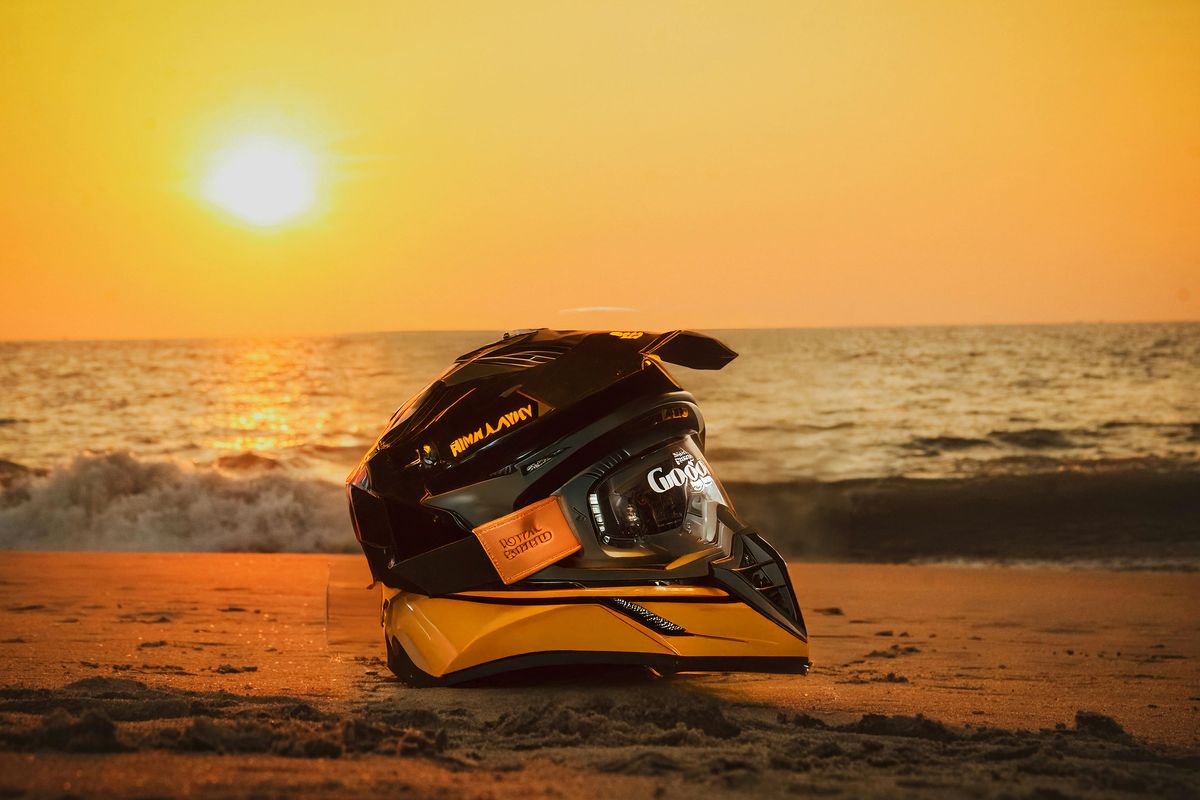} & \dimoutimg{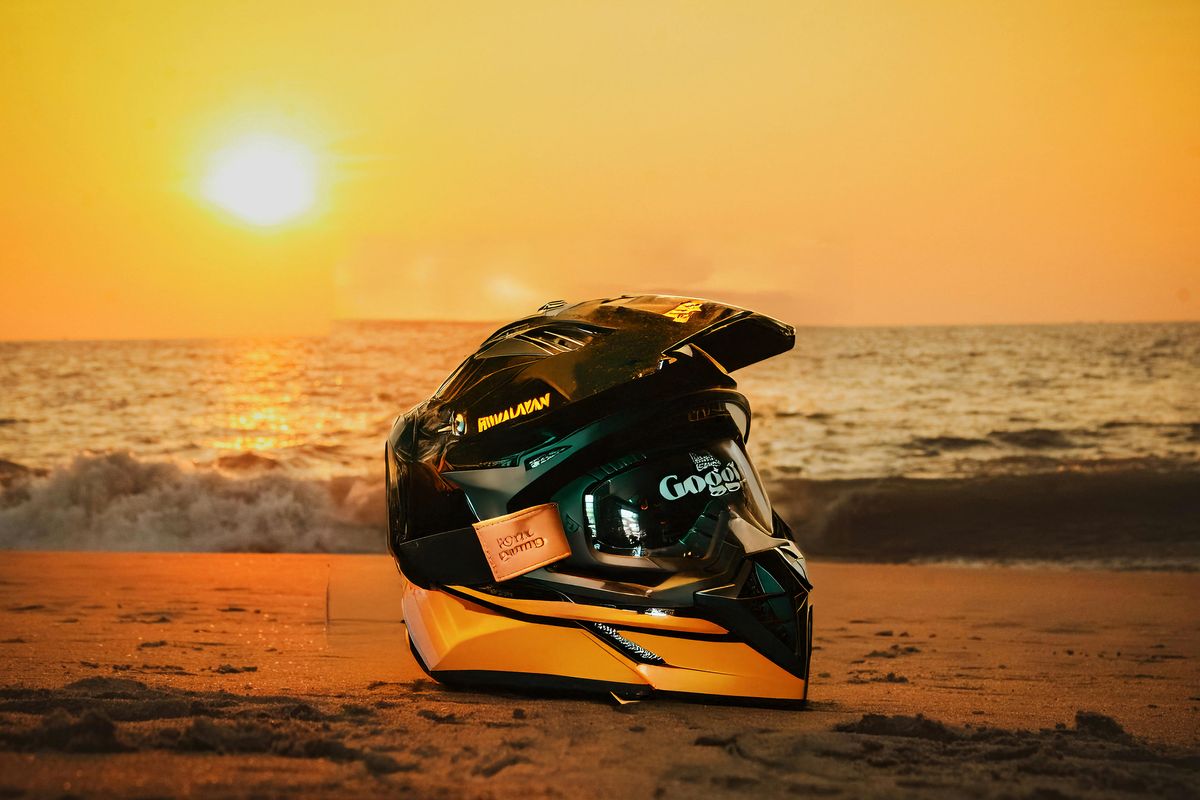} \\
            \end{tabular}
        \\[0.5mm]
        \textbf{(c)} &
            \begin{tabular}[c]{@{}*{3}{>{\centering\arraybackslash}m{\dimleftbgwidth}}@{}}
                \usebox{\dimbgboxc} &
                \dimrightimg{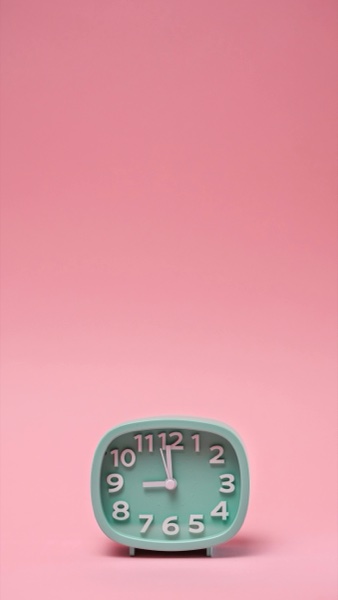} &
                \dimleftbgmask{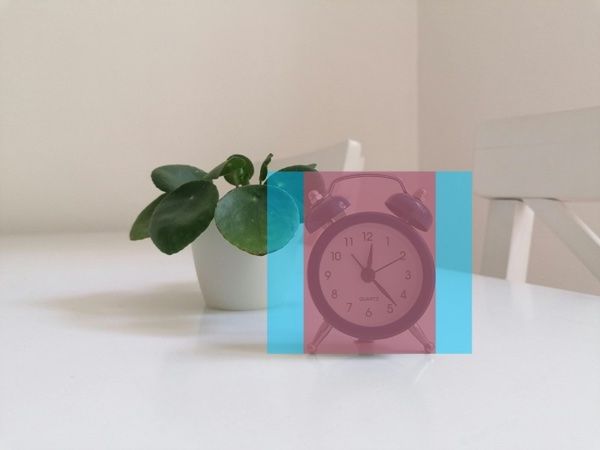} \\
            \end{tabular}
        &
            \begin{tabular}[c]{@{}*{3}{>{\centering\arraybackslash}m{\dimoutwidth}}@{}}
                \multicolumn{3}{c}{\fontsize{5.5}{6.5}\selectfont\textit{OmniPaint}} \\[0.2mm]
                \dimoutimg{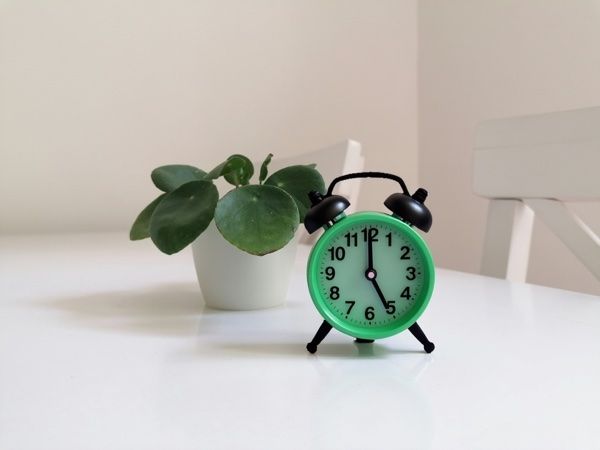} & \dimoutimg{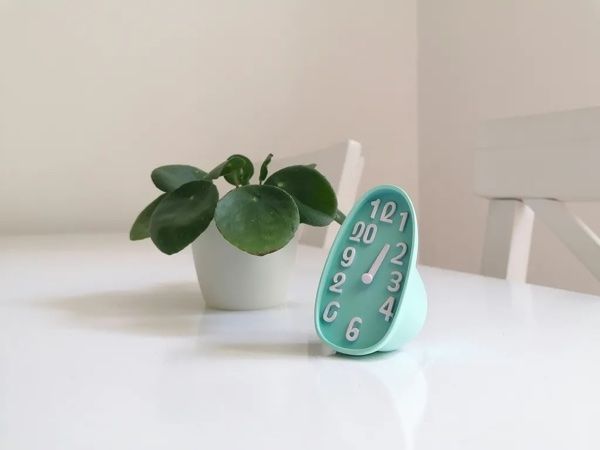} & \dimoutimg{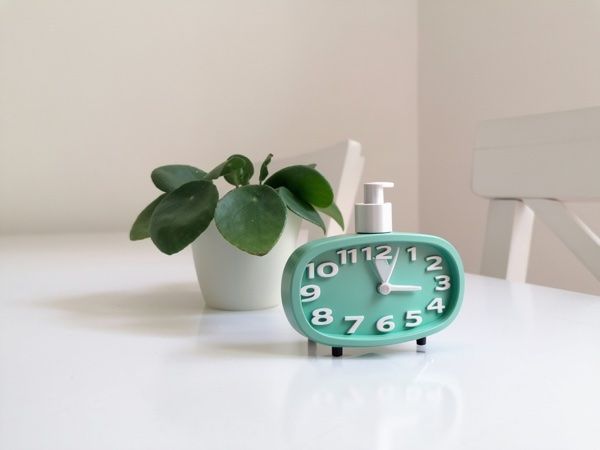} \\[0.2mm]
                \multicolumn{3}{c}{\fontsize{5.5}{6.5}\selectfont\textit{InsertAnything}} \\[0.2mm]
                \dimoutimg{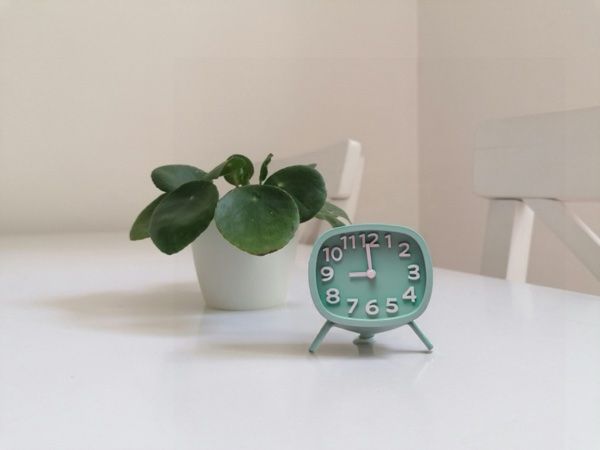} & \dimoutimg{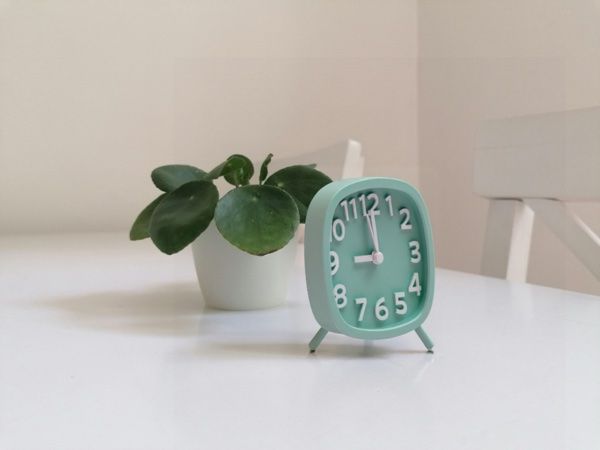} & \dimoutimg{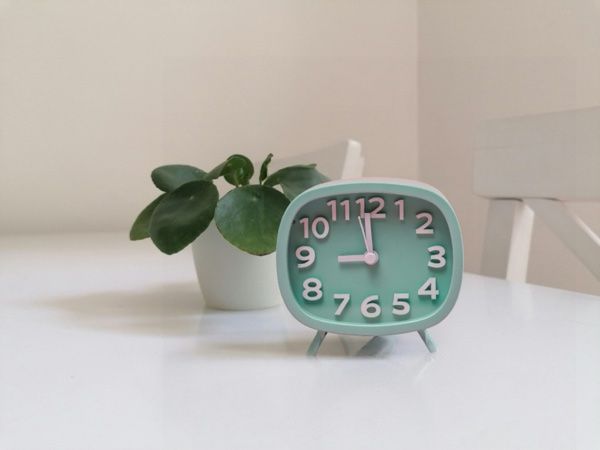} \\[0.2mm]
                \multicolumn{3}{c}{\fontsize{5.5}{6.5}\selectfont\textit{ObjectStitch}} \\[0.2mm]
                \dimoutimg{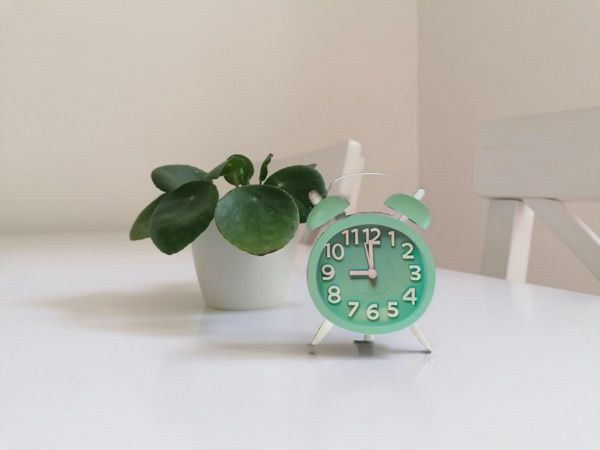} & \dimoutimg{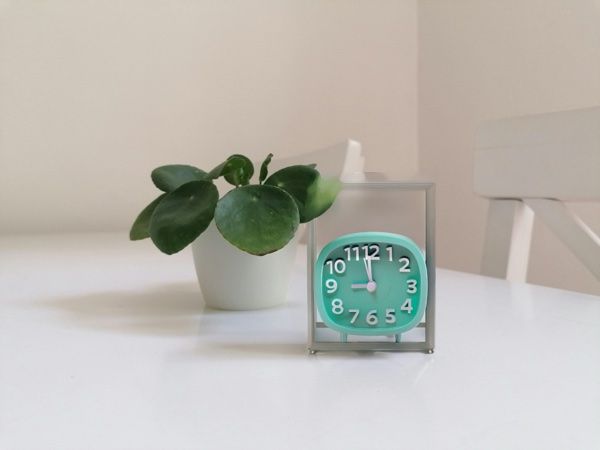} & \dimoutimg{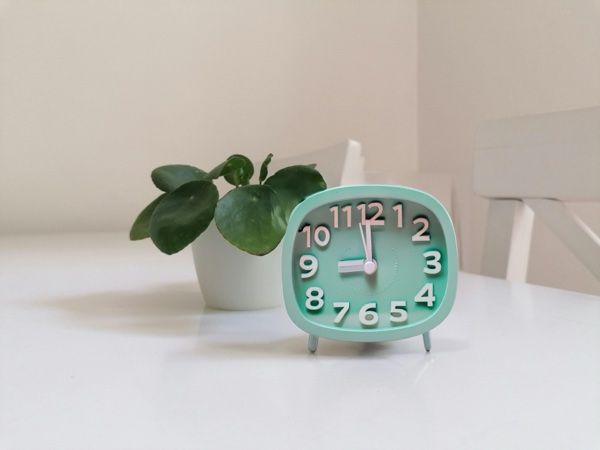} \\
            \end{tabular}
        \\
    \end{tabular}
    \caption{\textbf{Dimension-aware mask computation across models.} Left inputs: background and product. Mask overlay visualizes the dim-aware (blue) and bounding-box (pink) regions. Right: outputs under Freeform, BBox, and Dim-Aware masks for OmniPaint, InsertAnything, and ObjectStitch. Dim-aware masks preserve object proportions more reliably than Freeform and BBox masks across all models.}
    \label{fig:dimension_results}
\end{figure*}

\begin{figure*}[!tp]
    \centering
    \setlength{\tabcolsep}{0.5pt}
    \renewcommand{\arraystretch}{1.0}
    \scriptsize
    \fontsize{6.5}{7.2}\selectfont

    \newcommand{\occleftwidth}{1.40cm}
    \newcommand{\occleftasset}[1]{\includegraphics[width=\occleftwidth,height=\occleftwidth,keepaspectratio]{#1}}
    \newcommand{\occleftmask}[1]{\includegraphics[width=\occleftwidth,height=\occleftwidth,keepaspectratio]{#1}}
    \newcommand{\occoutwidth}{1.40cm}
    \newcommand{\occoutimg}[1]{\includegraphics[width=\occoutwidth,height=\occoutwidth,keepaspectratio]{#1}}
    \newsavebox{\occbgboxa}
    \newsavebox{\occbgboxb}
    \newsavebox{\occbgboxc}
    \sbox{\occbgboxa}{\occleftasset{figures/occlusion/6/bg.jpg}}
    \sbox{\occbgboxb}{\occleftasset{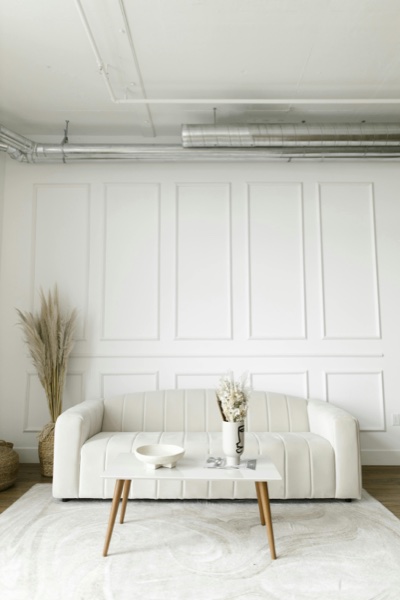}}
    \sbox{\occbgboxc}{\occleftasset{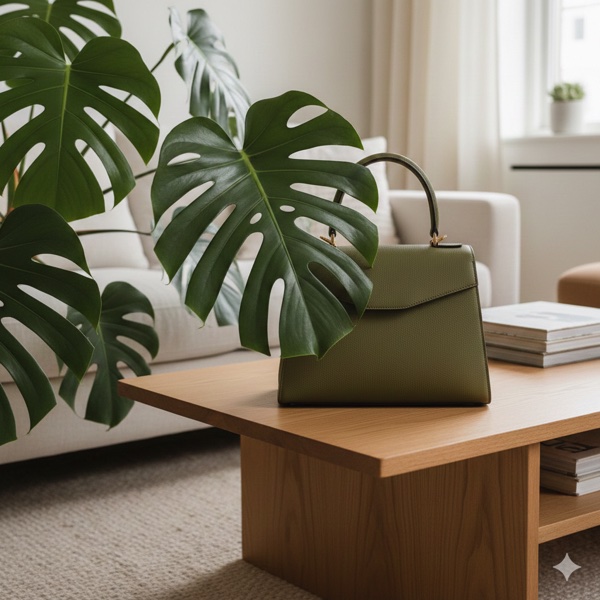}}

    \begin{tabular}{@{}c@{\hspace{0.8mm}}c@{\hspace{4mm}}c@{}}
        &
            {\fontsize{5.5}{6.5}\selectfont
            \begin{tabular}[b]{@{}*{4}{>{\centering\arraybackslash}m{\occleftwidth}}@{}}
                \makecell{\textbf{Background}\\\textbf{Image}} & \makecell{\textbf{Product}\\\textbf{Image}} & \makecell{\textbf{Overlayed}\\\textbf{Masks}} & \textbf{Occluders}
            \end{tabular}
        }
        &
            {\fontsize{5.5}{6.5}\selectfont
            \begin{tabular}[b]{@{}>{{\centering\arraybackslash}}m{0.28cm} *{3}{>{\centering\arraybackslash}m{\occoutwidth}}@{}}
                & \makecell{\textbf{Result with}\\\textbf{Freeform Mask}} & \makecell{\textbf{Result with}\\\textbf{BBox}} & \makecell{\textbf{Result with Dim-}\\\textbf{Aware Mask (Ours)}}
            \end{tabular}
        } \\[-0.3mm]
        \textbf{(a)} &
            \begin{tabular}[c]{@{}*{4}{>{\centering\arraybackslash}m{\occleftwidth}}@{}}
                \usebox{\occbgboxa} &
                \occleftasset{figures/occlusion/6/obj.jpg} &
                \occleftmask{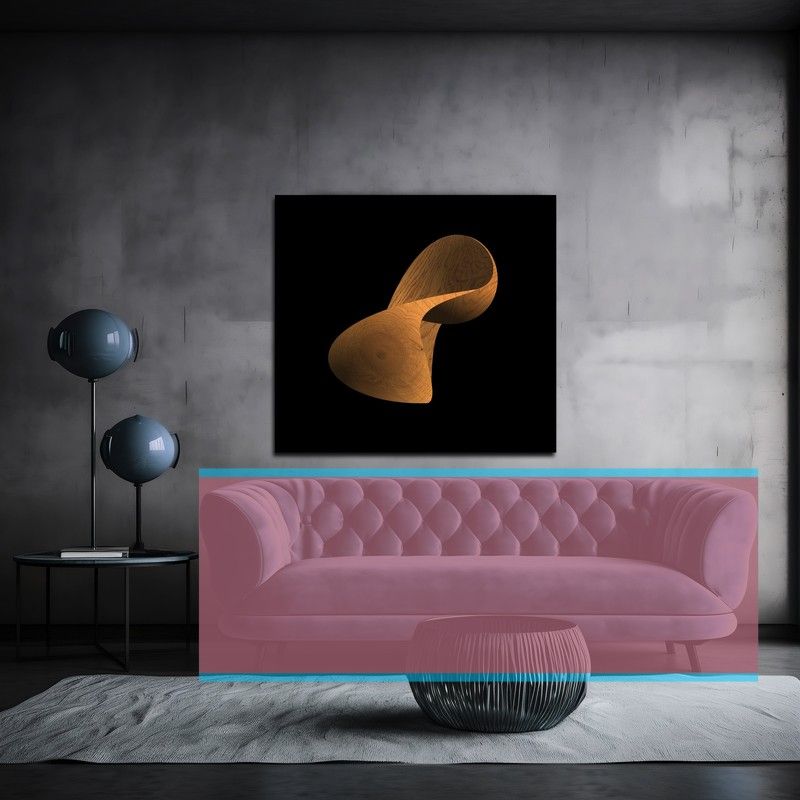} &
                \occleftmask{figures/occlusion/6/overlaps_composite.jpg} \\
            \end{tabular}
        &
            \begin{tabular}[c]{@{}>{\centering\arraybackslash}m{0.28cm} *{3}{>{\centering\arraybackslash}m{\occoutwidth}}@{}}
                \multicolumn{4}{c}{\fontsize{5.5}{6.5}\selectfont\textit{InsertAnything}} \\[0.2mm]
                \rotatebox[origin=c]{90}{\fontsize{4}{4.5}\selectfont\textbf{Before}} & \occoutimg{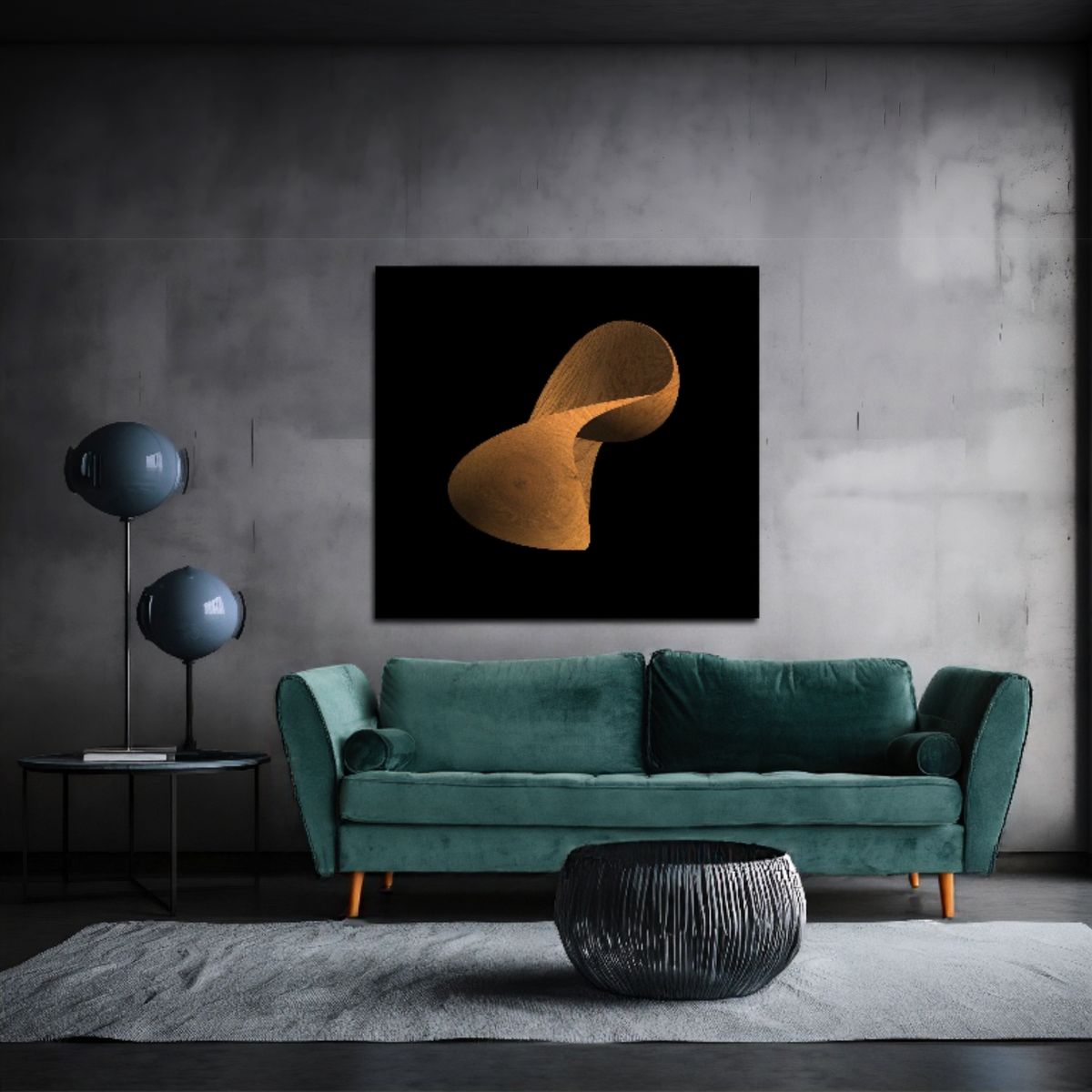} & \occoutimg{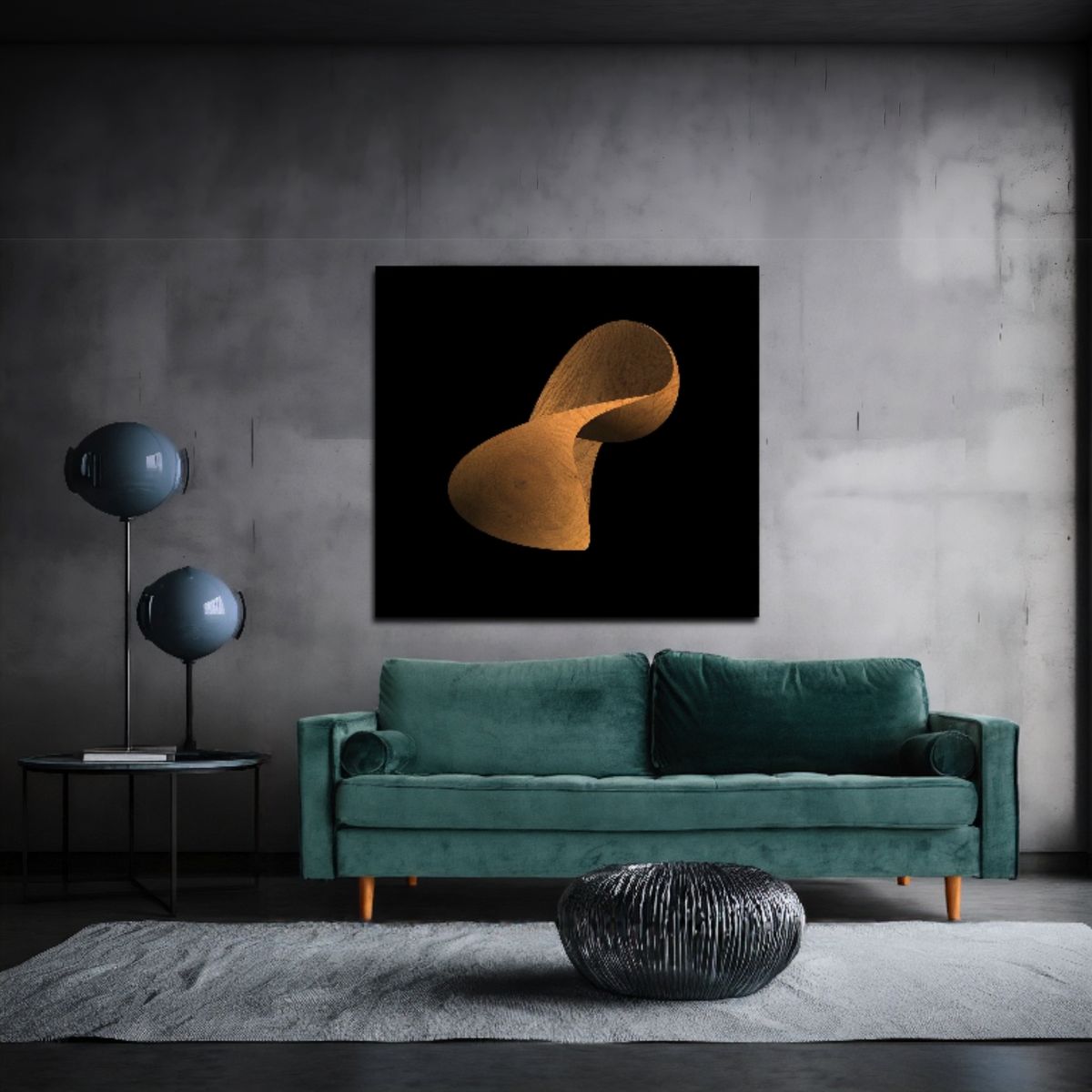} & \occoutimg{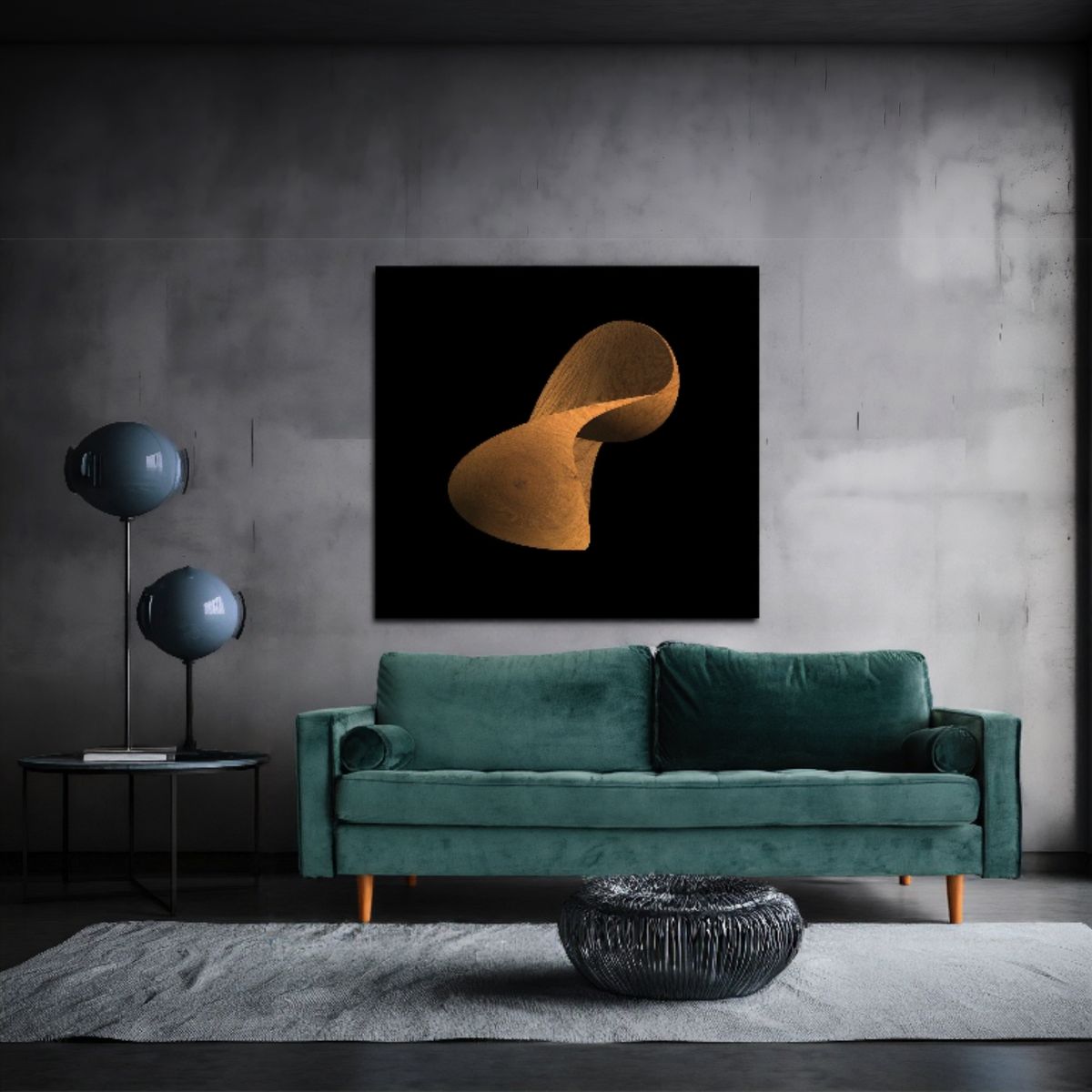} \\[0.2mm]
                \rotatebox[origin=c]{90}{\fontsize{4}{4.5}\selectfont\textbf{After}} & \occoutimg{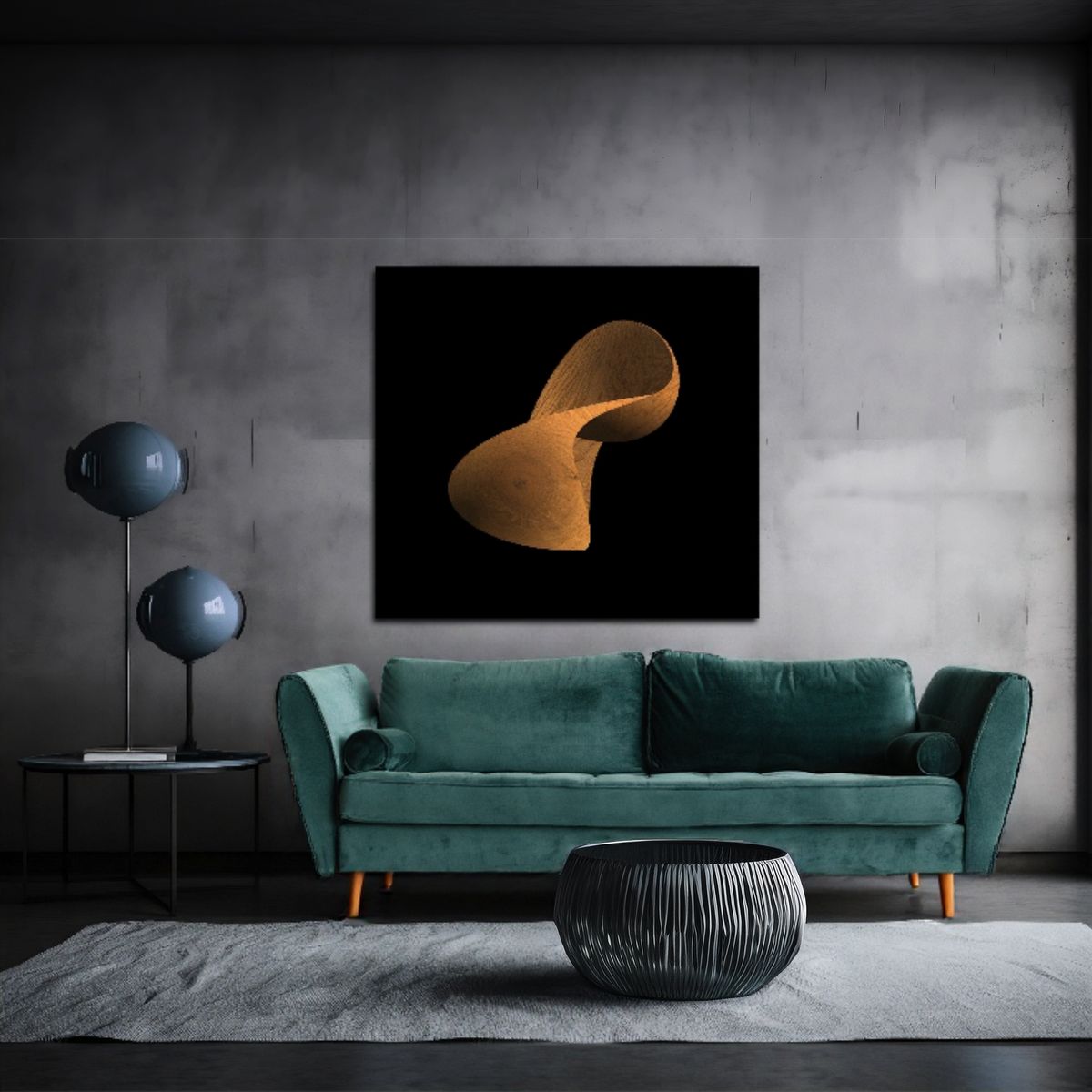} & \occoutimg{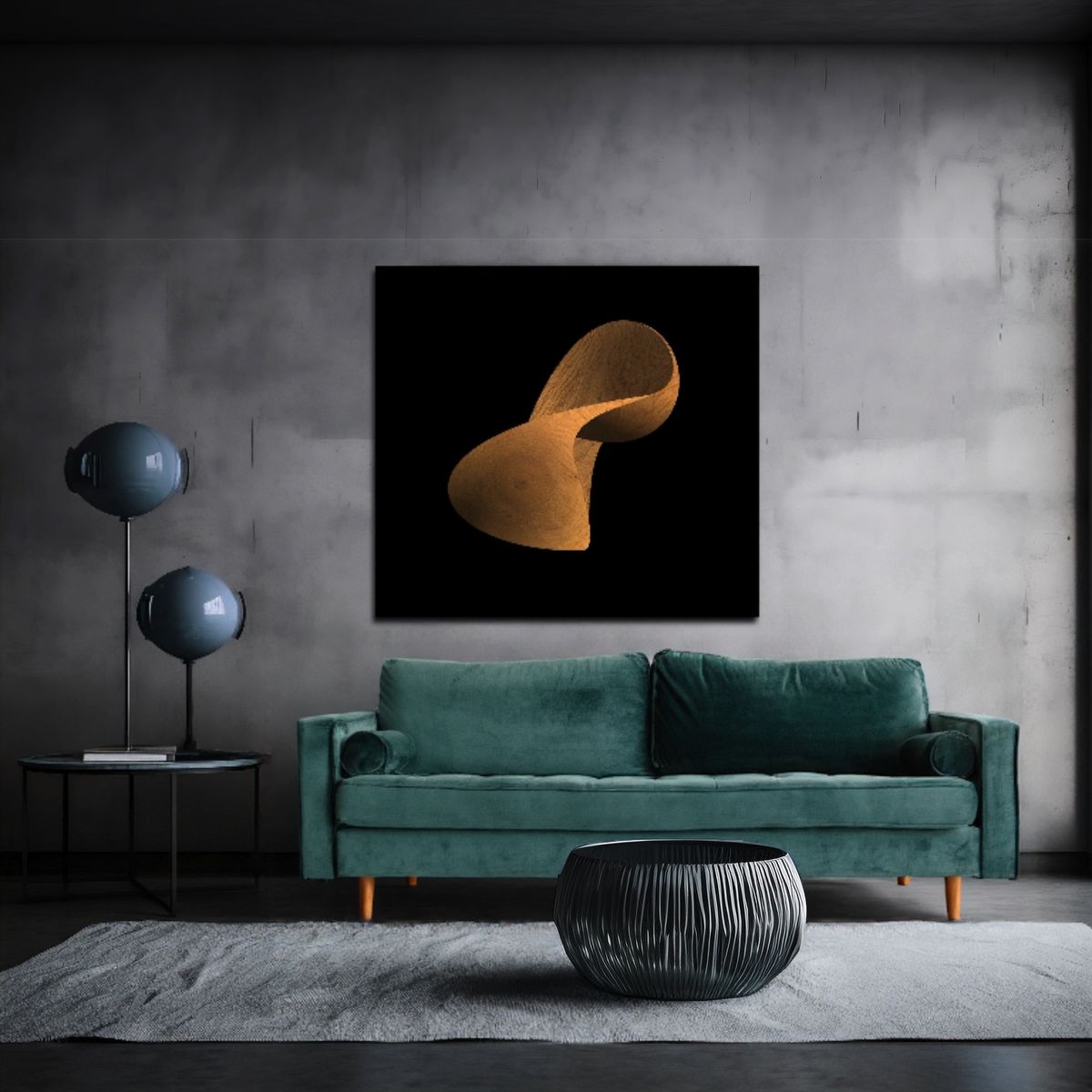} & \occoutimg{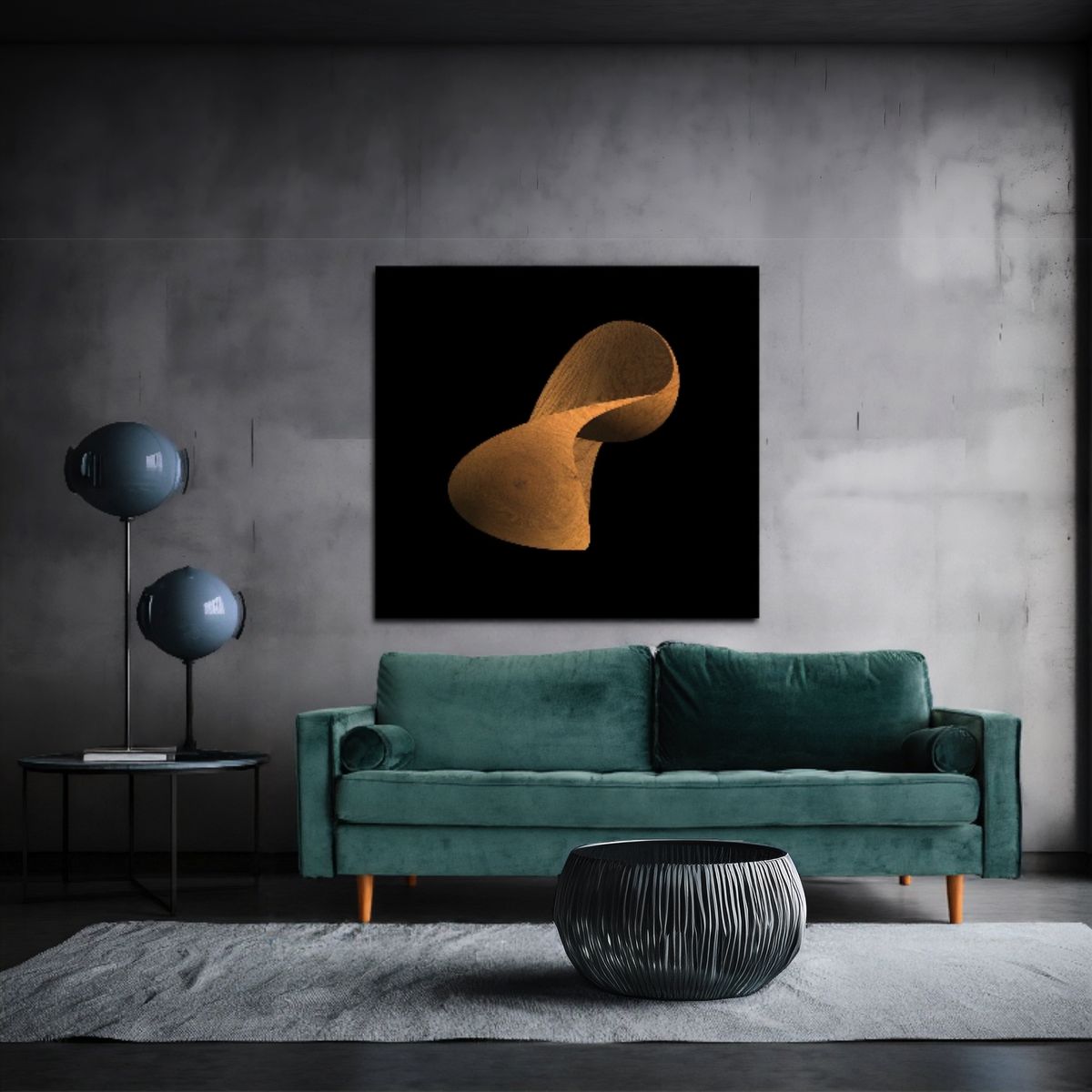} \\[0.2mm]
                \multicolumn{4}{c}{\fontsize{5.5}{6.5}\selectfont\textit{ObjectStitch}} \\[0.2mm]
                \rotatebox[origin=c]{90}{\fontsize{4}{4.5}\selectfont\textbf{Before}} & \occoutimg{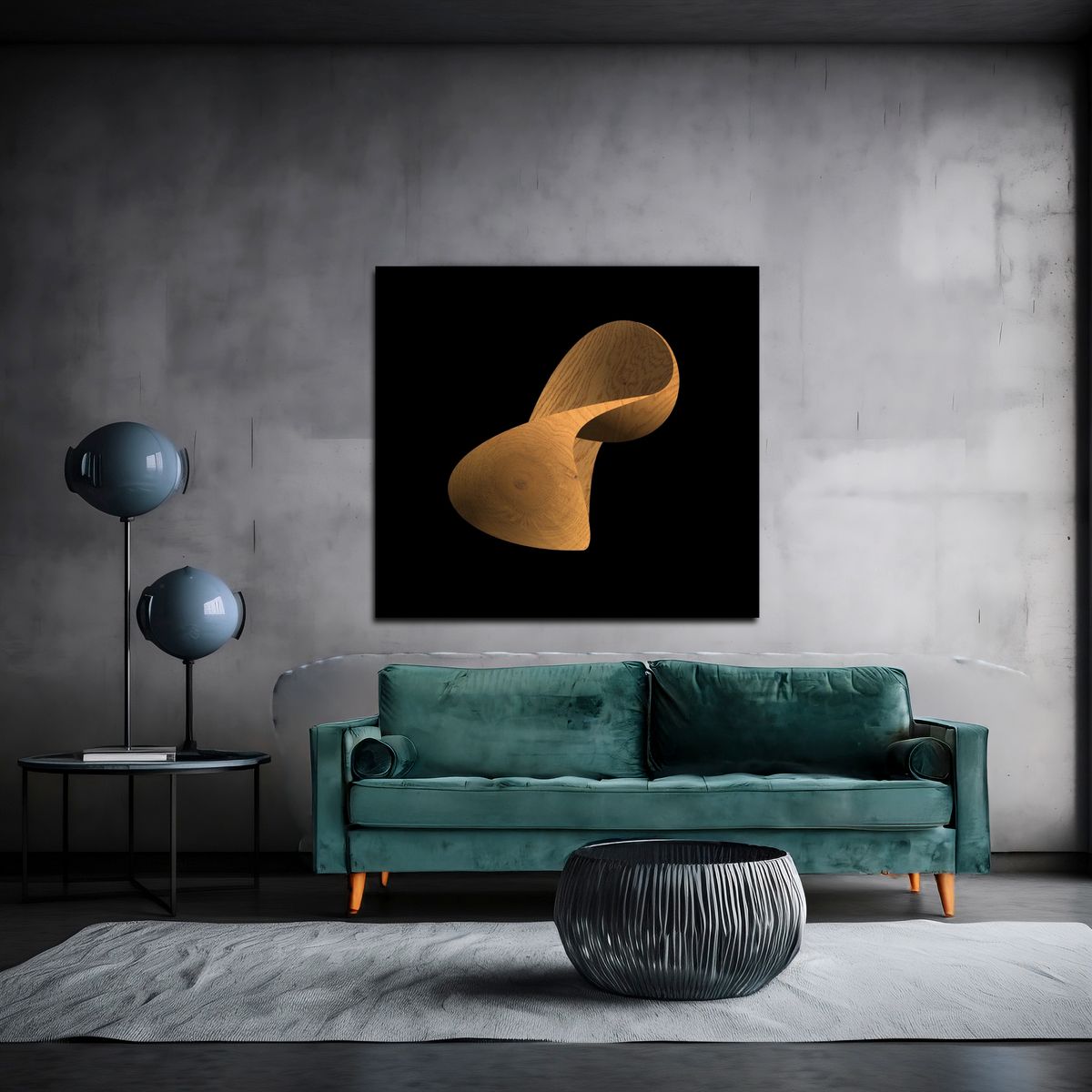} & \occoutimg{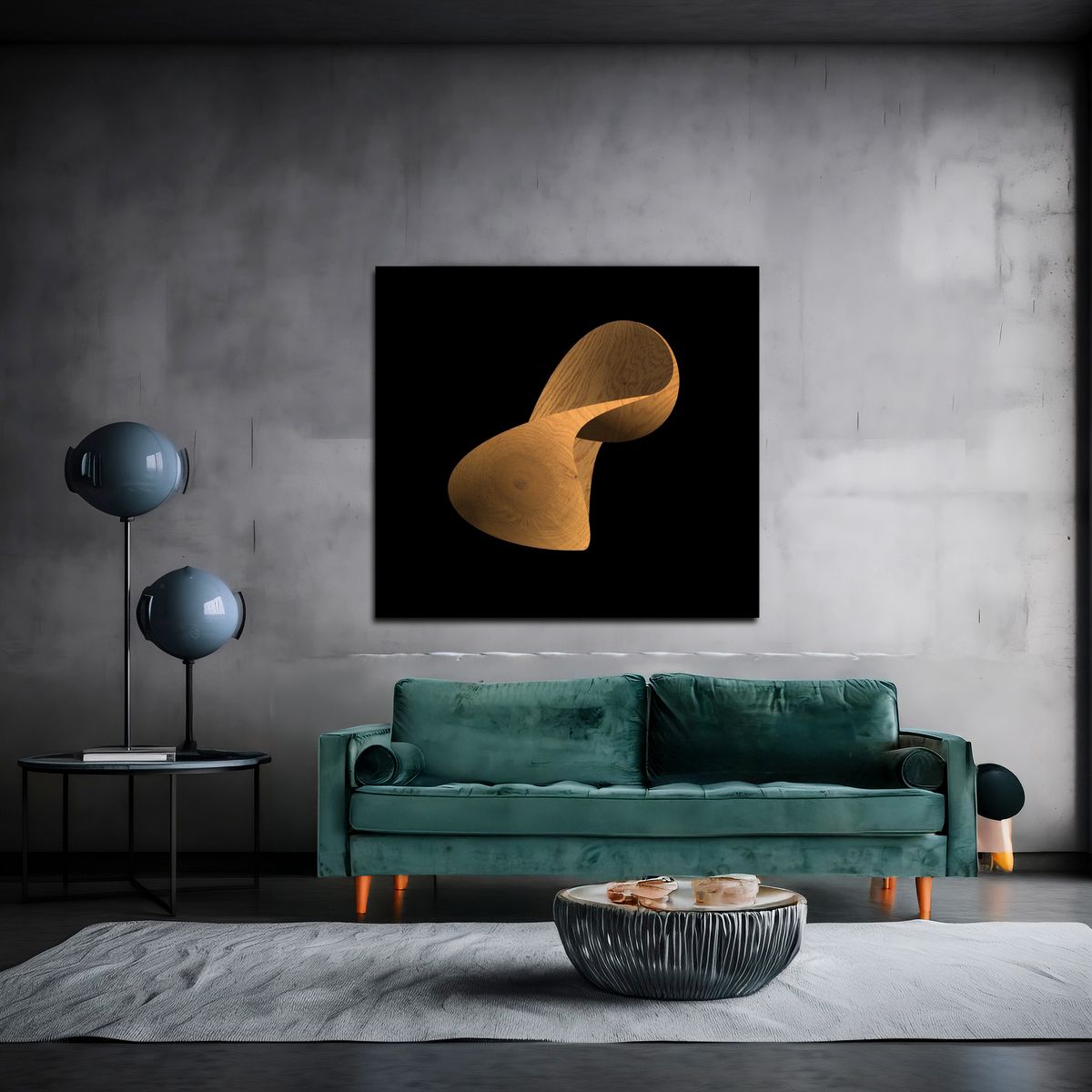} & \occoutimg{figures/occlusion/6/objectStitch_optimal_before.jpg} \\[0.2mm]
                \rotatebox[origin=c]{90}{\fontsize{4}{4.5}\selectfont\textbf{After}} & \occoutimg{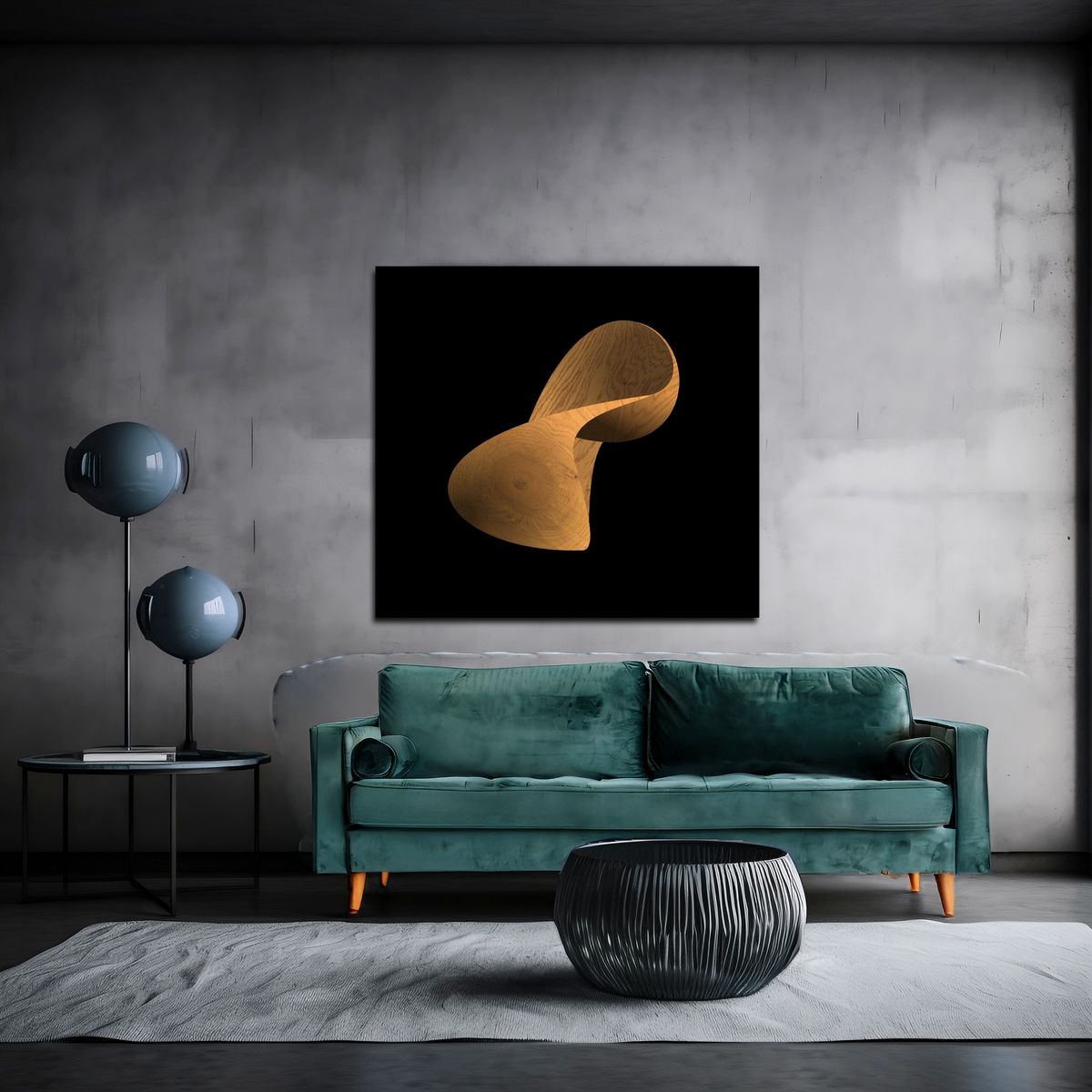} & \occoutimg{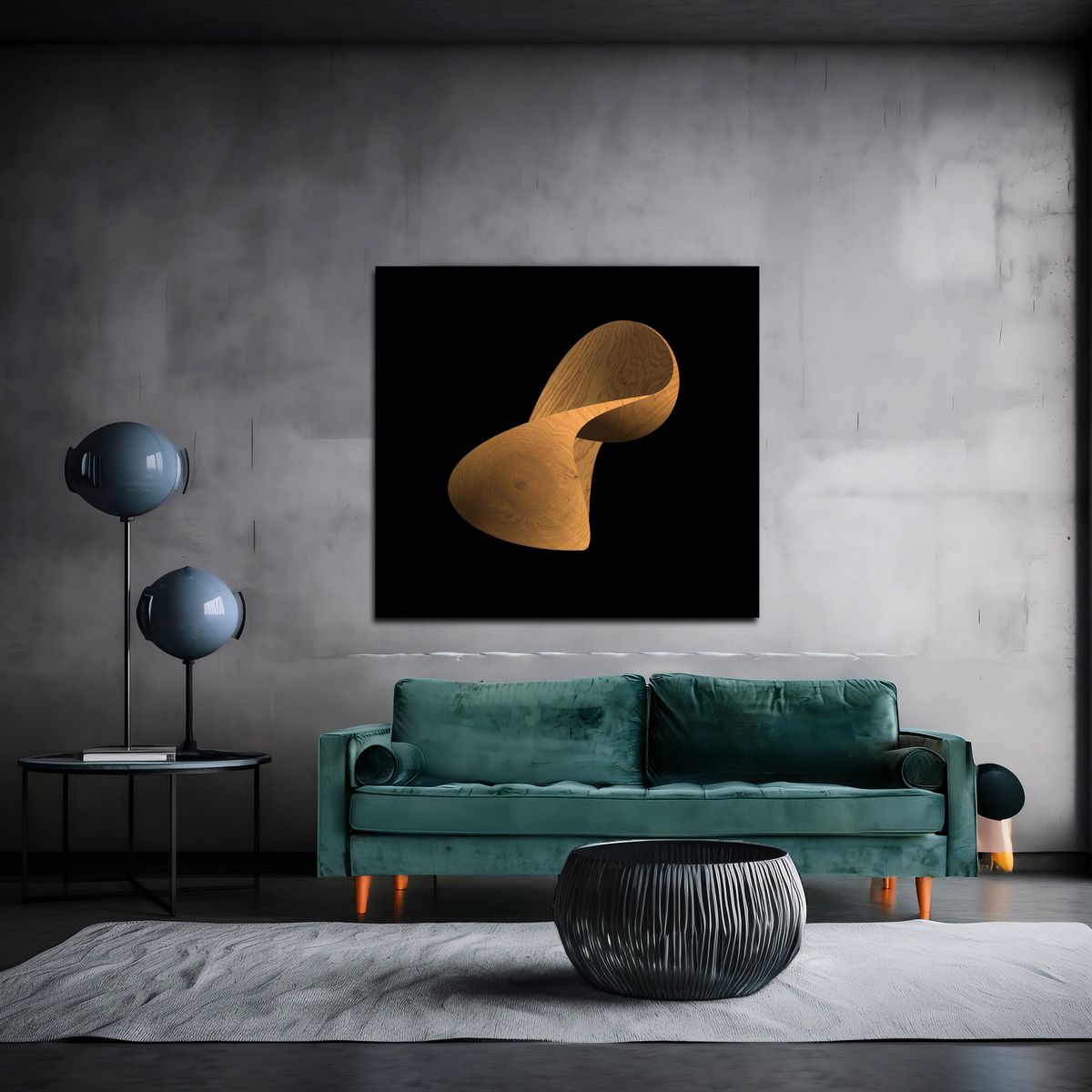} & \occoutimg{figures/occlusion/6/objectStitch_optimal_after.jpg} \\
            \end{tabular}
        \\[0.3mm]
        \textbf{(b)} &
            \begin{tabular}[c]{@{}*{4}{>{\centering\arraybackslash}m{\occleftwidth}}@{}}
                \usebox{\occbgboxb} &
                \occleftasset{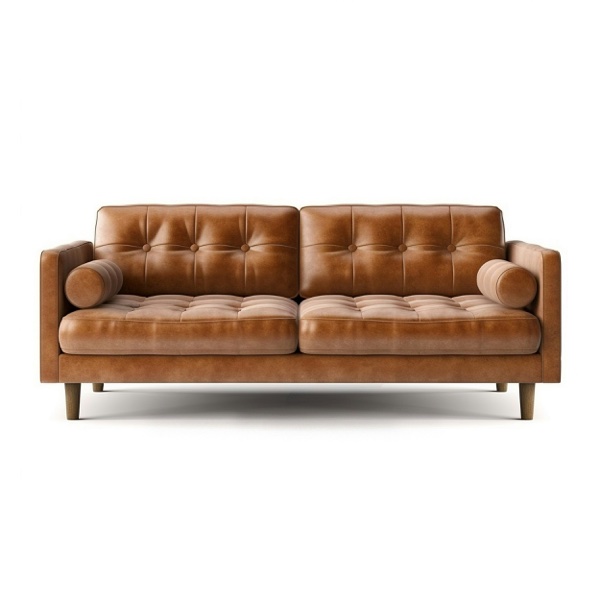} &
                \occleftmask{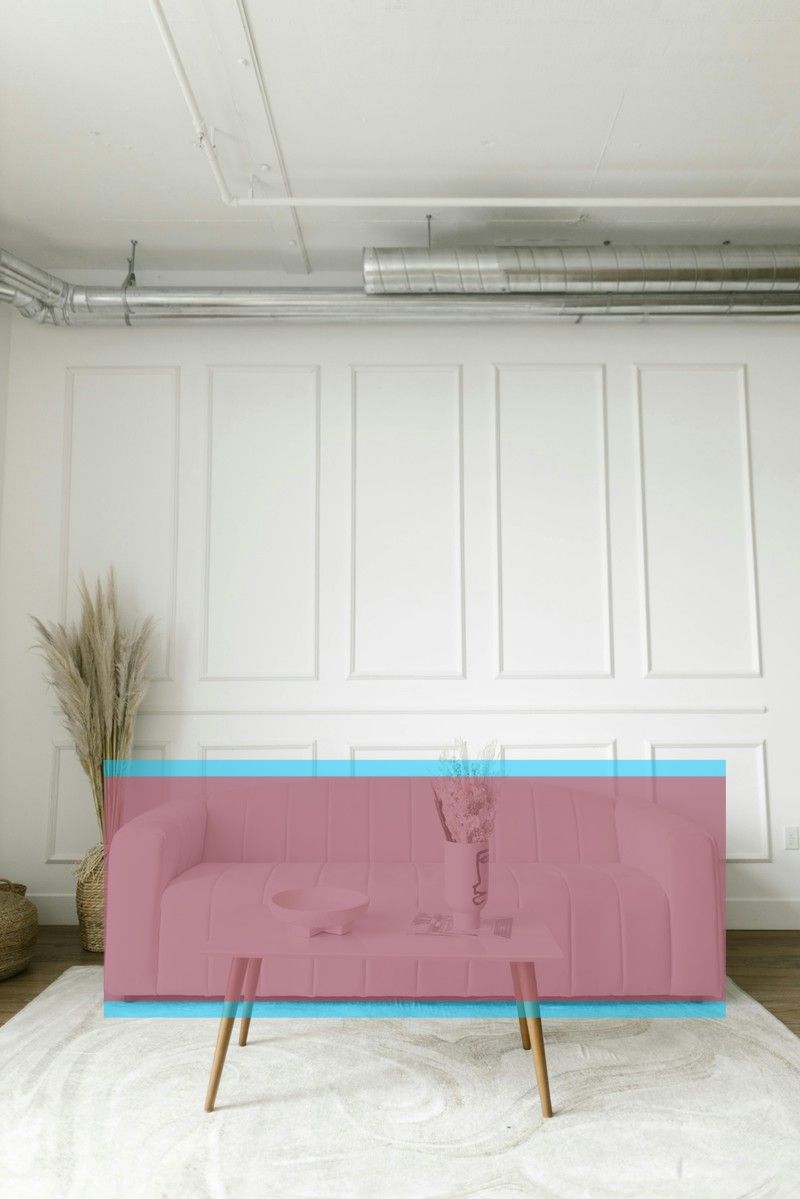} &
                \occleftmask{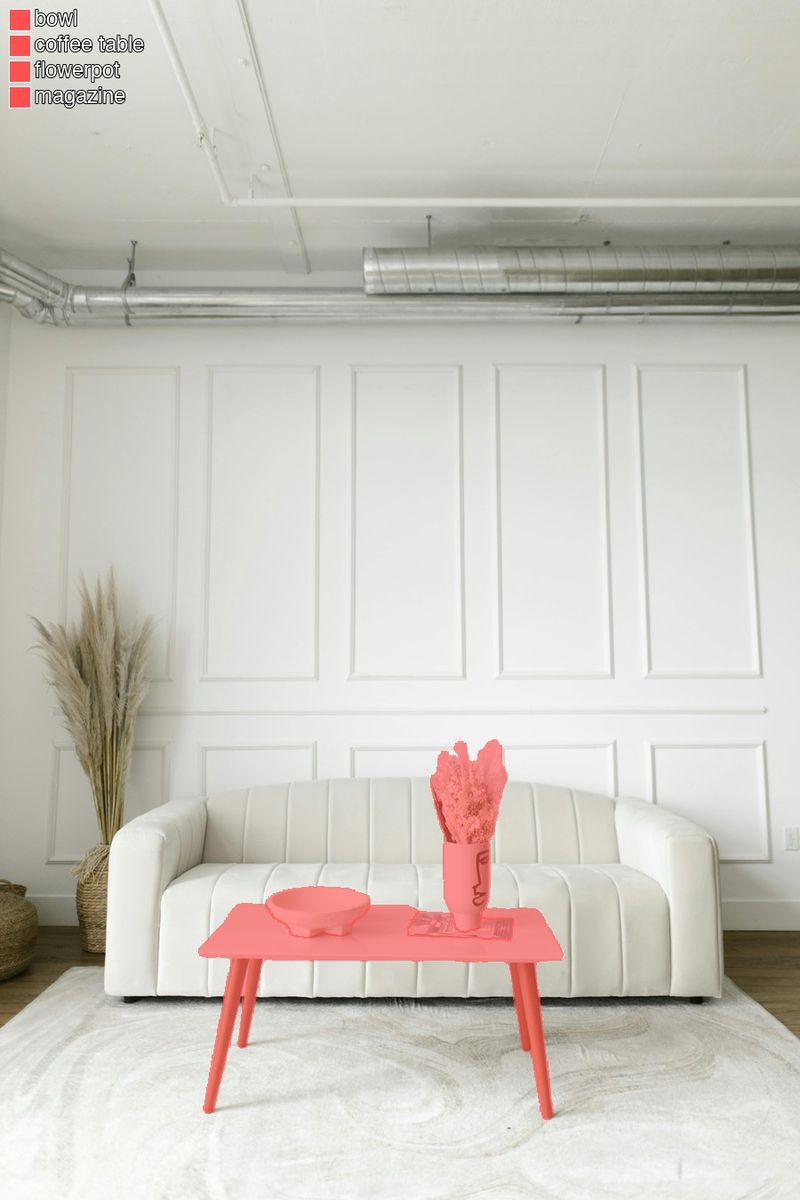} \\
            \end{tabular}
        &
            \begin{tabular}[c]{@{}>{\centering\arraybackslash}m{0.28cm} *{3}{>{\centering\arraybackslash}m{\occoutwidth}}@{}}
                \multicolumn{4}{c}{\fontsize{5.5}{6.5}\selectfont\textit{InsertAnything}} \\[0.2mm]
                \rotatebox[origin=c]{90}{\fontsize{4}{4.5}\selectfont\textbf{Before}} & \occoutimg{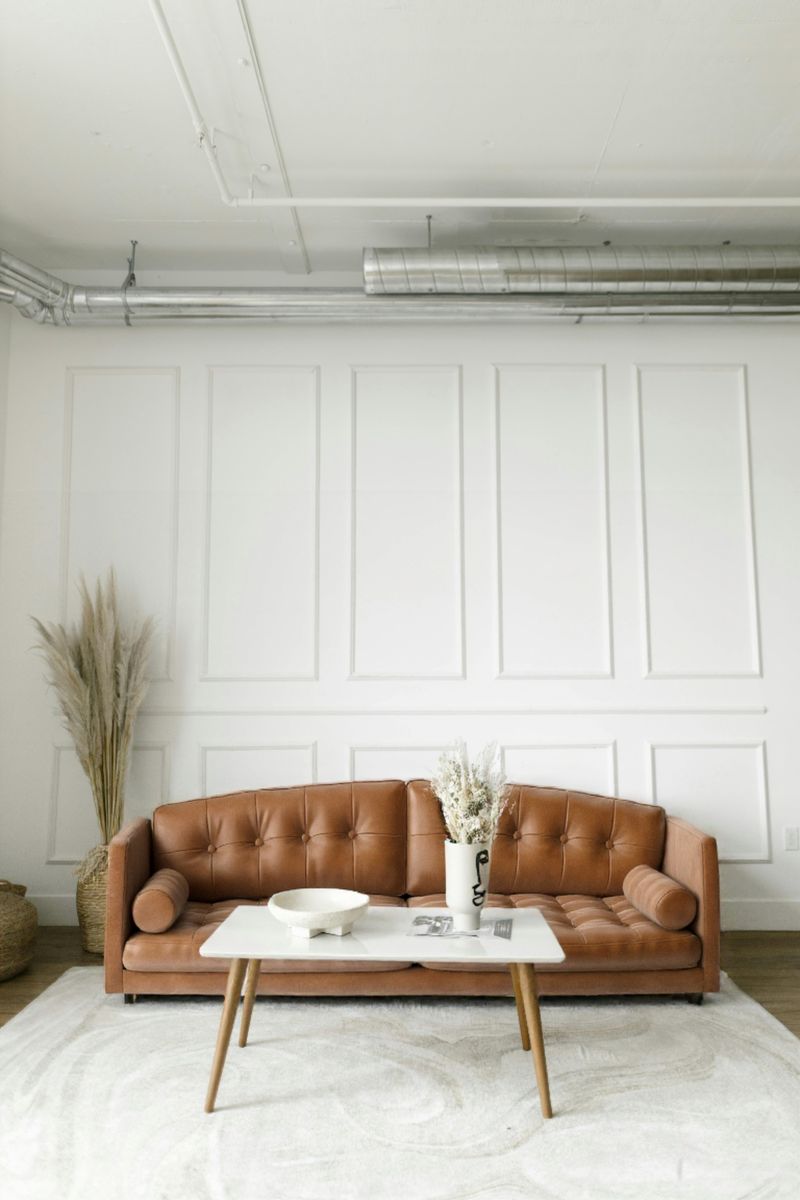} & \occoutimg{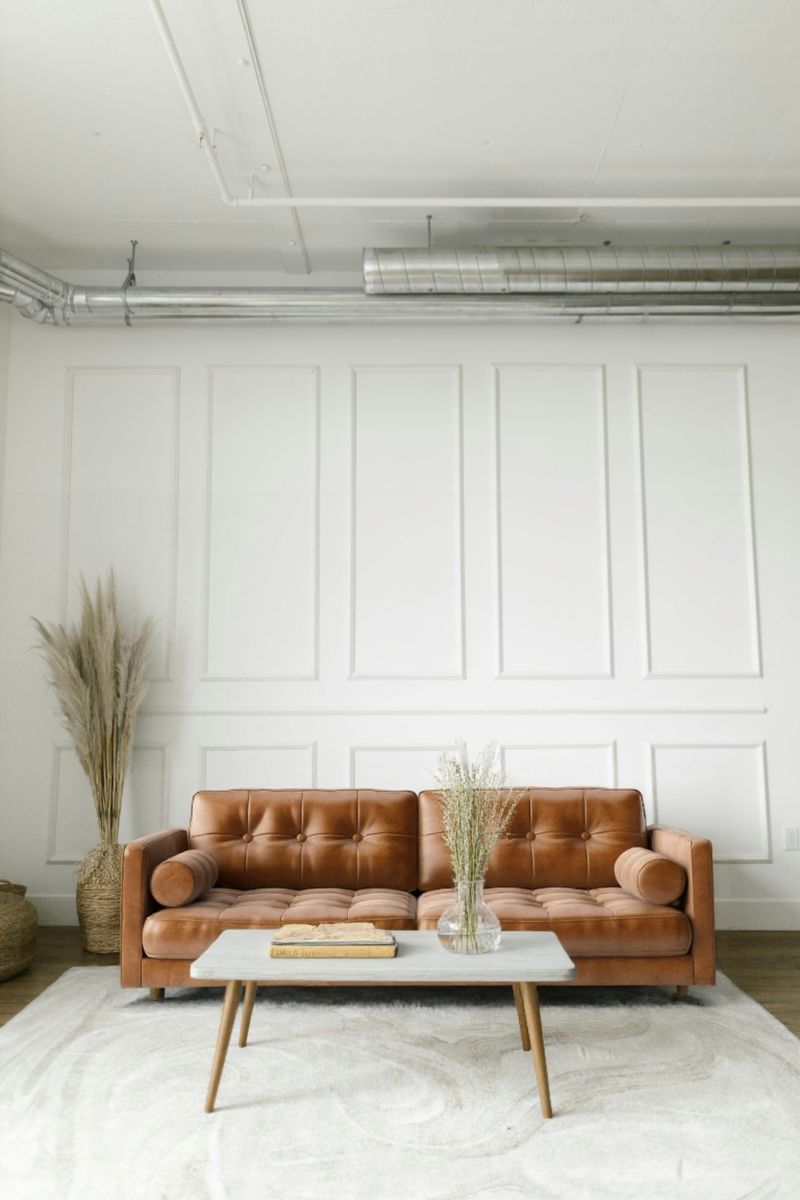} & \occoutimg{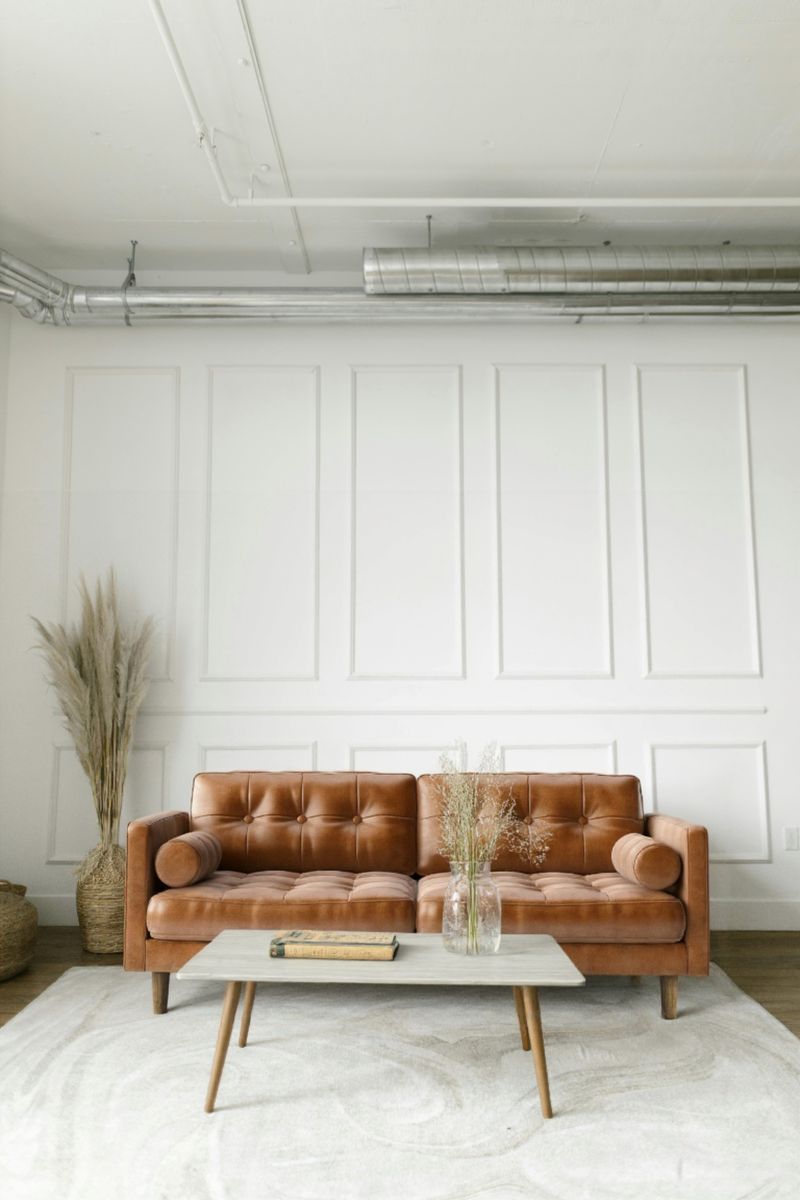} \\[0.2mm]
                \rotatebox[origin=c]{90}{\fontsize{4}{4.5}\selectfont\textbf{After}} & \occoutimg{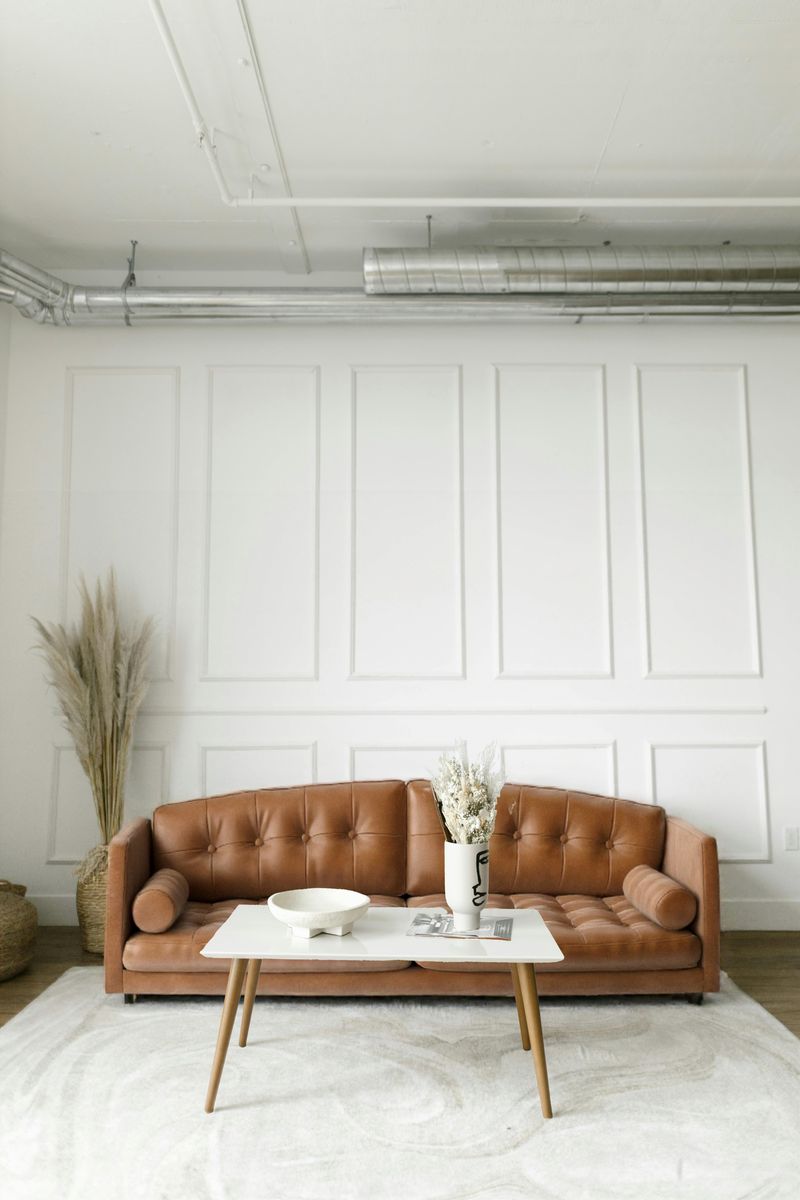} & \occoutimg{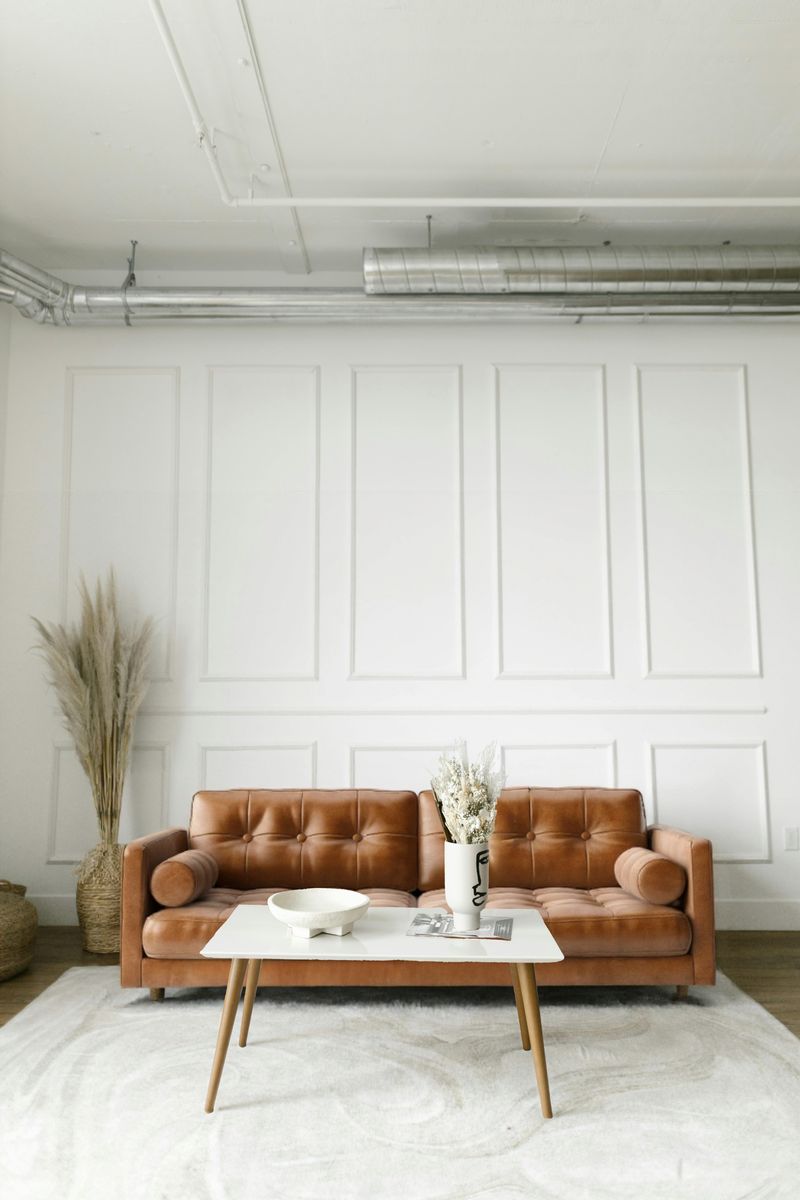} & \occoutimg{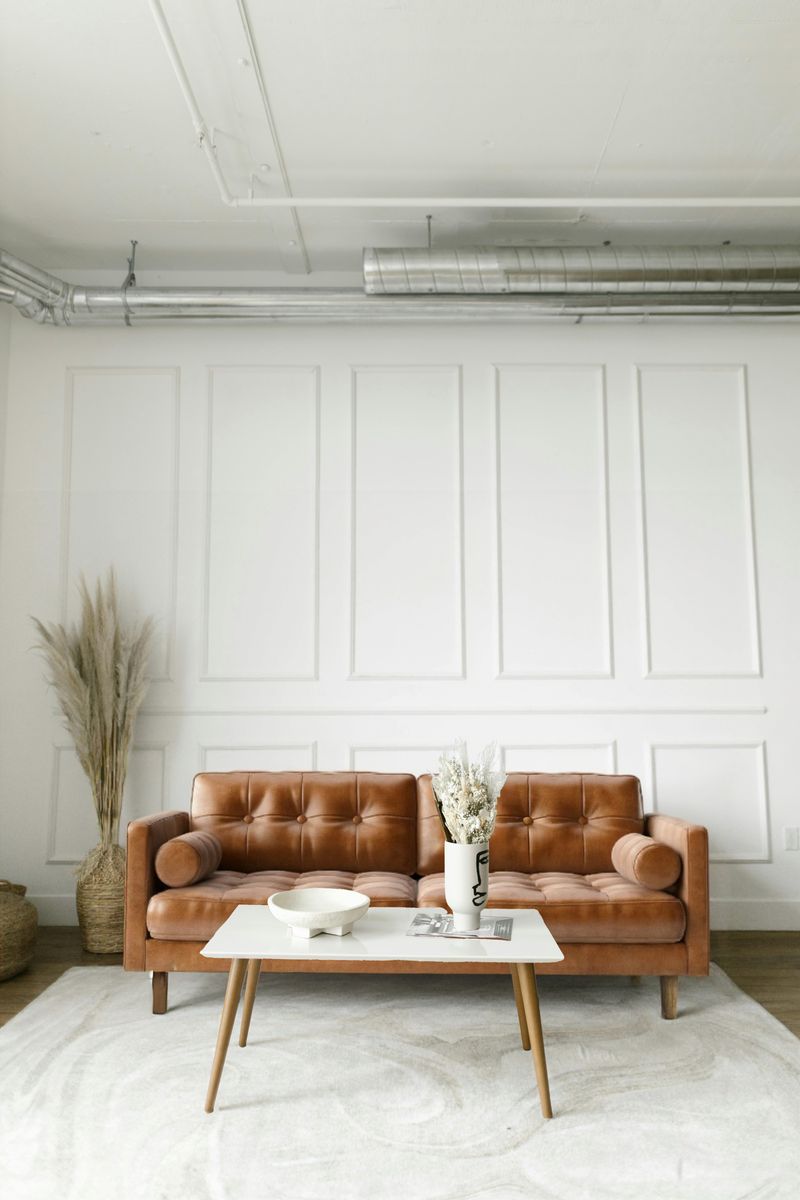} \\[0.2mm]
                \multicolumn{4}{c}{\fontsize{5.5}{6.5}\selectfont\textit{ObjectStitch}} \\[0.2mm]
                \rotatebox[origin=c]{90}{\fontsize{4}{4.5}\selectfont\textbf{Before}} & \occoutimg{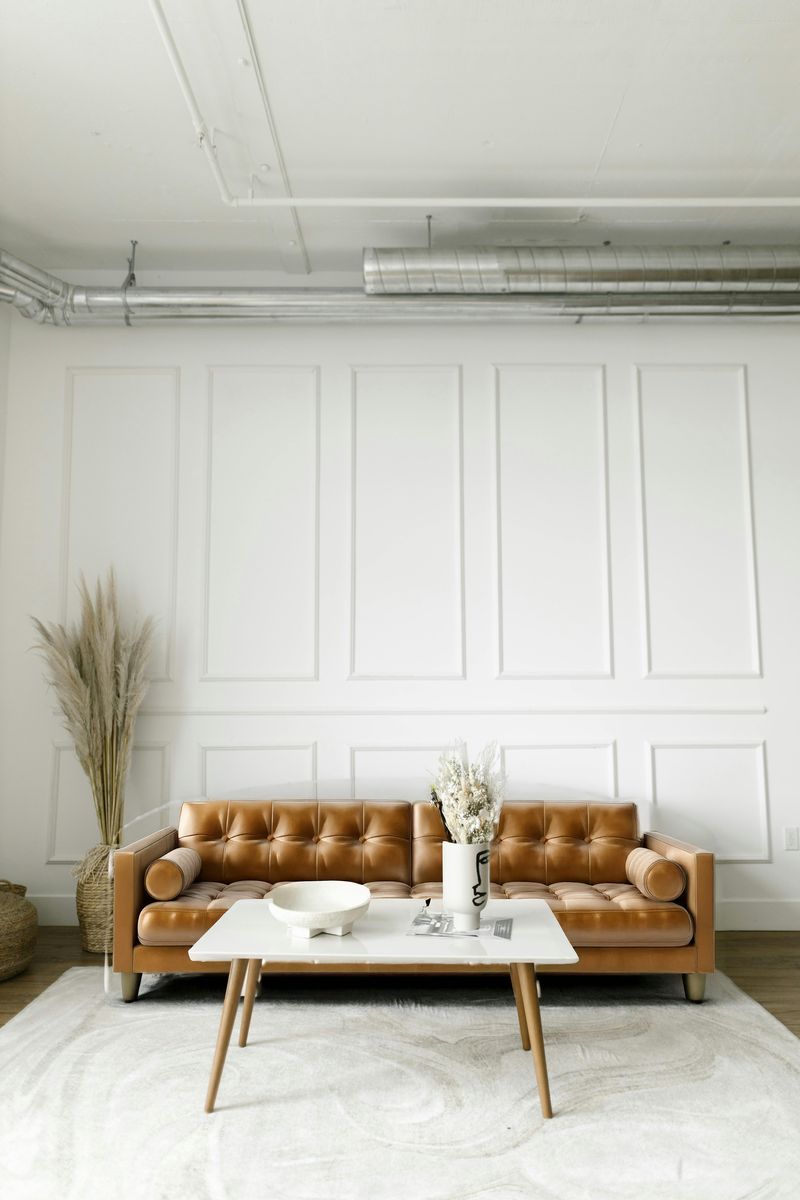} & \occoutimg{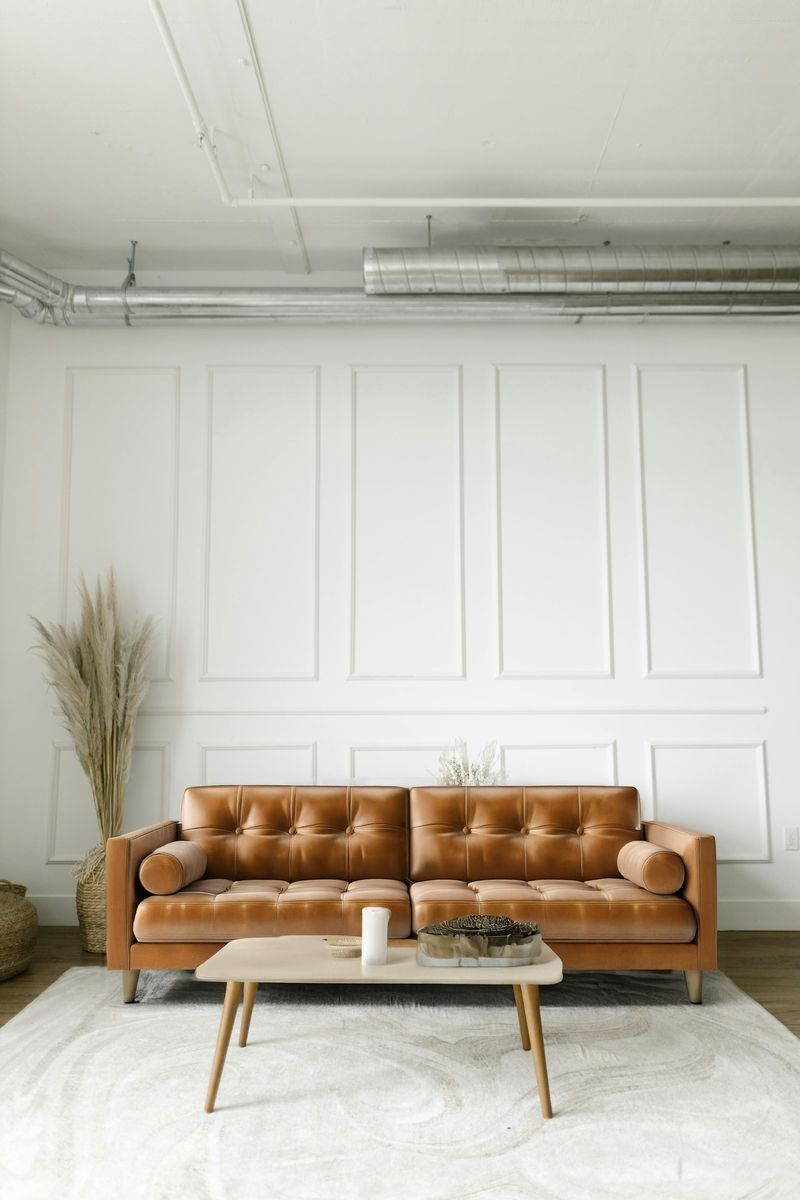} & \occoutimg{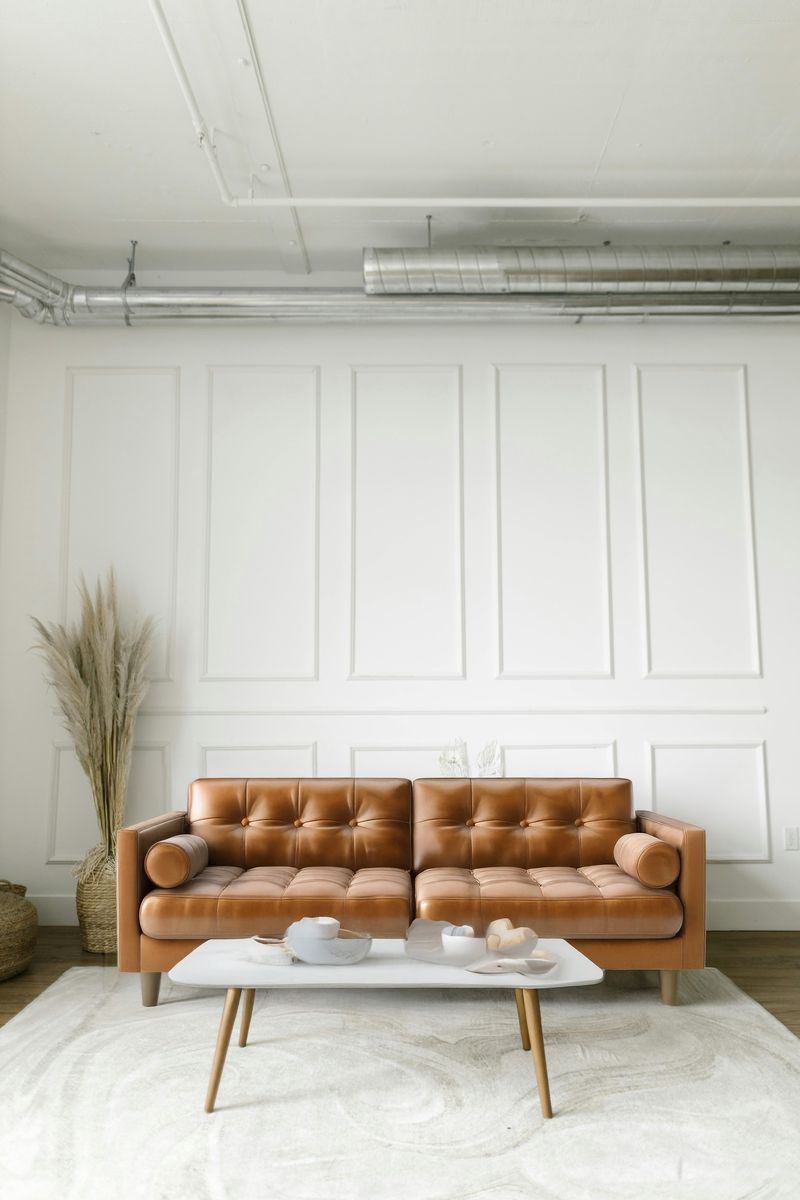} \\[0.2mm]
                \rotatebox[origin=c]{90}{\fontsize{4}{4.5}\selectfont\textbf{After}} & \occoutimg{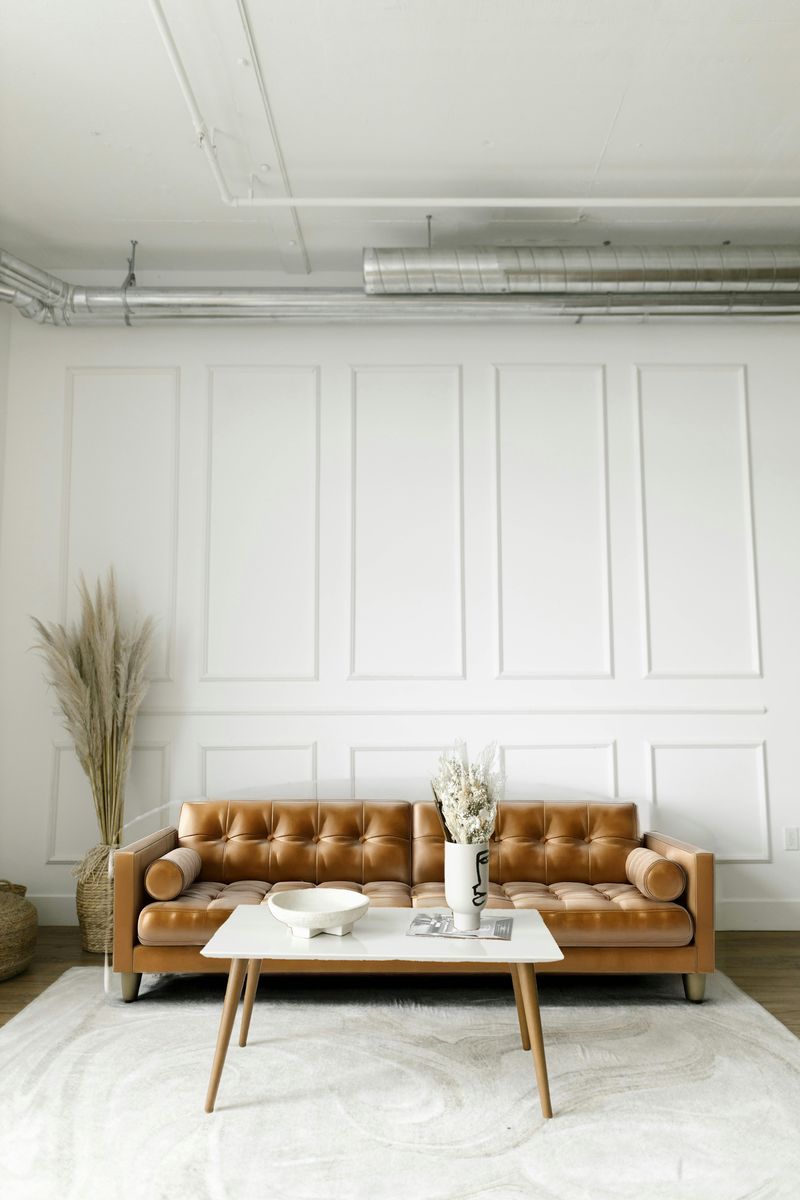} & \occoutimg{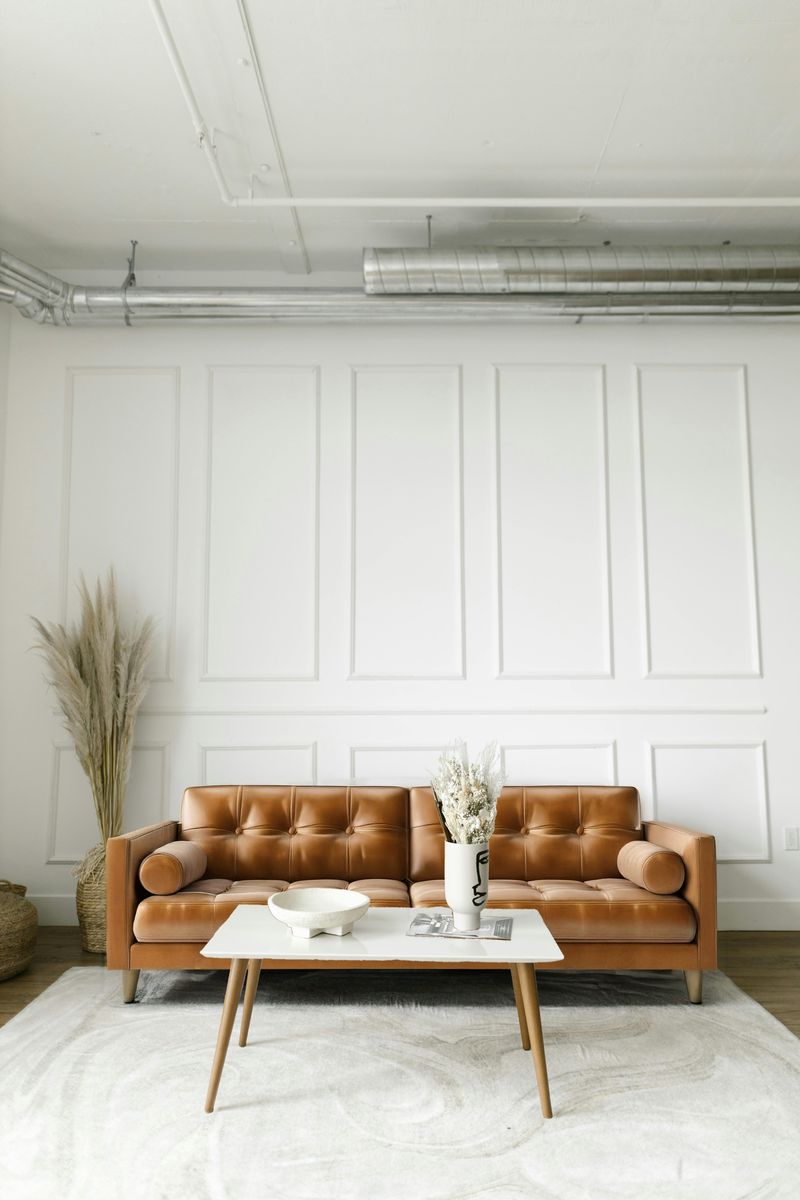} & \occoutimg{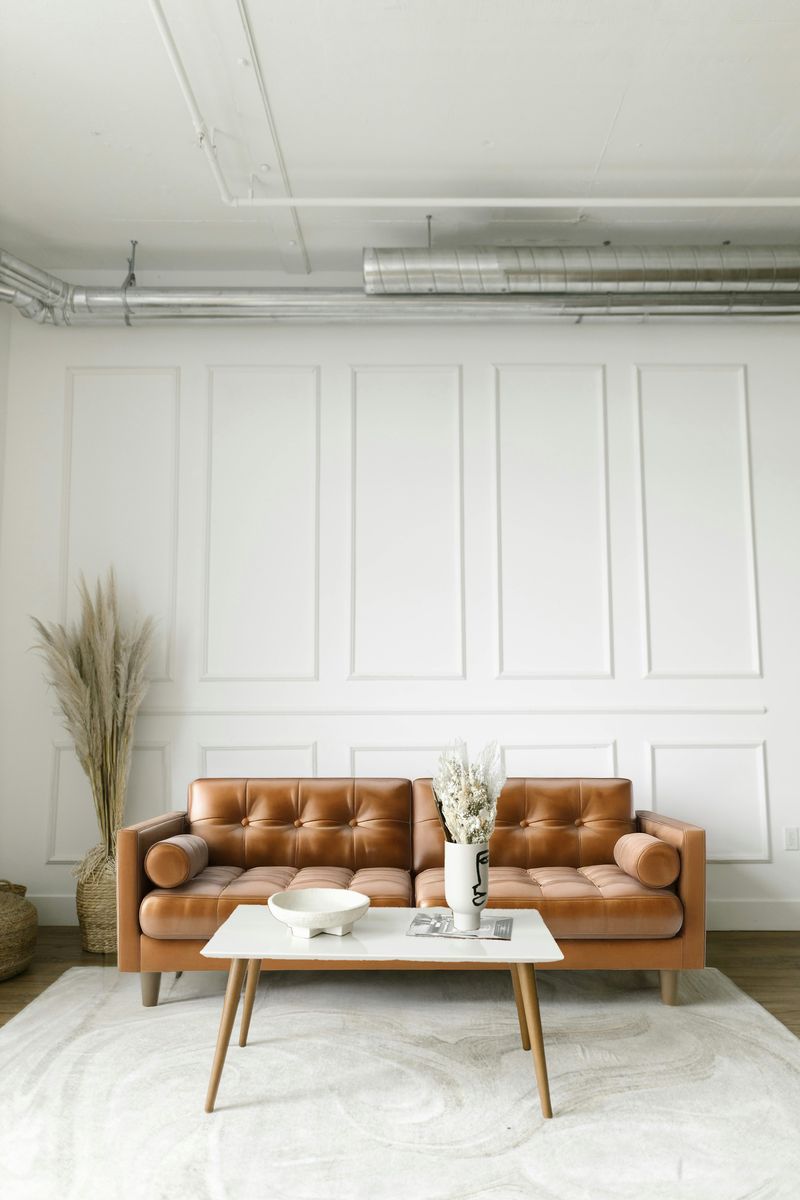} \\
            \end{tabular}
        \\[0.3mm]
        \textbf{(c)} &
            \begin{tabular}[c]{@{}*{4}{>{\centering\arraybackslash}m{\occleftwidth}}@{}}
                \usebox{\occbgboxc} &
                \occleftasset{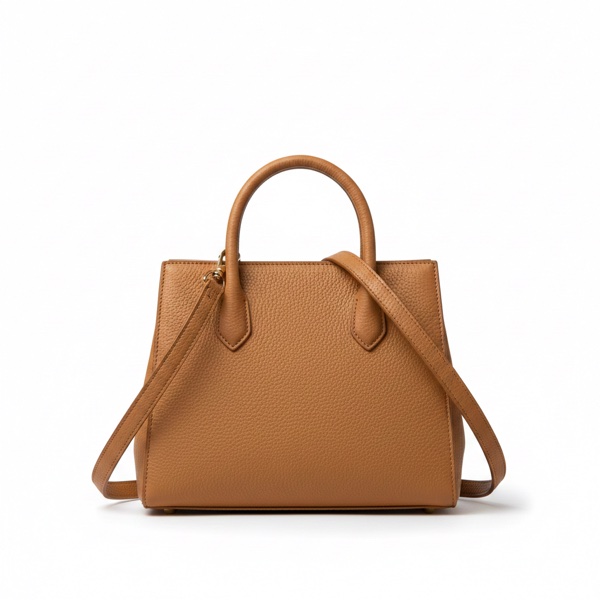} &
                \occleftmask{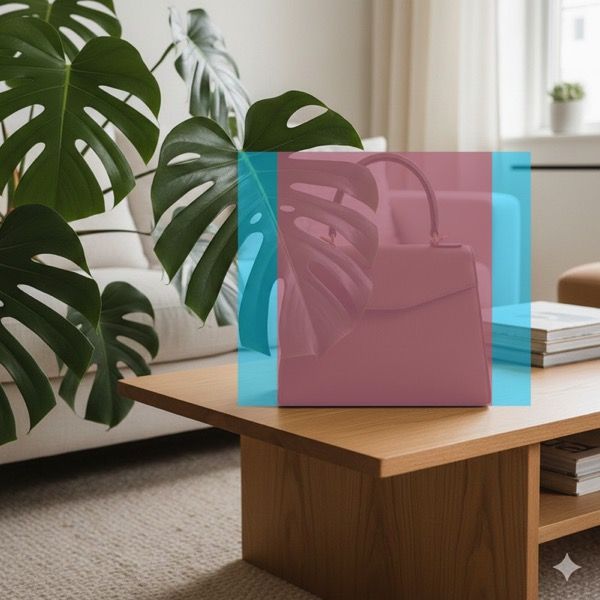} &
                \occleftmask{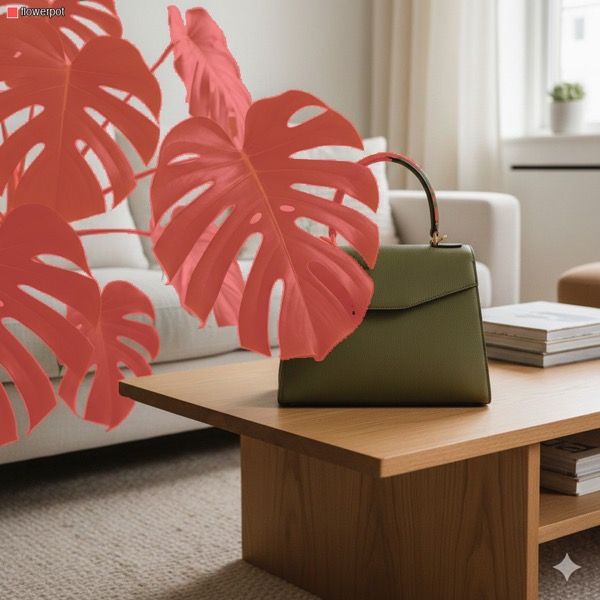} \\
            \end{tabular}
        &
            \begin{tabular}[c]{@{}>{\centering\arraybackslash}m{0.28cm} *{3}{>{\centering\arraybackslash}m{\occoutwidth}}@{}}
                \multicolumn{4}{c}{\fontsize{5.5}{6.5}\selectfont\textit{InsertAnything}} \\[0.2mm]
                \rotatebox[origin=c]{90}{\fontsize{4}{4.5}\selectfont\textbf{Before}} & \occoutimg{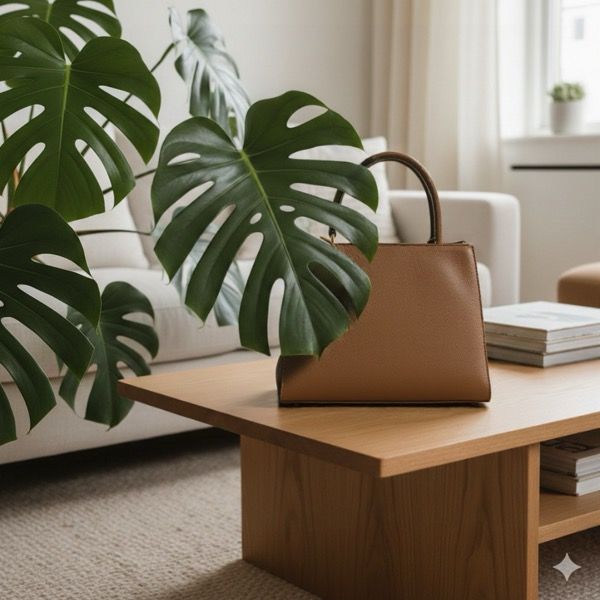} & \occoutimg{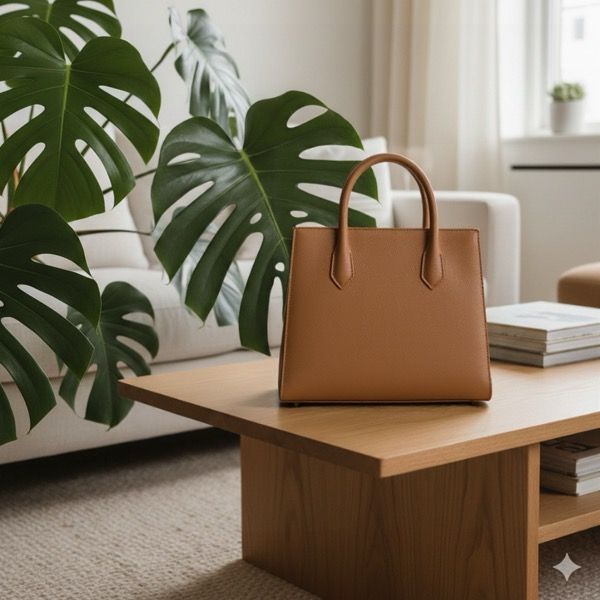} & \occoutimg{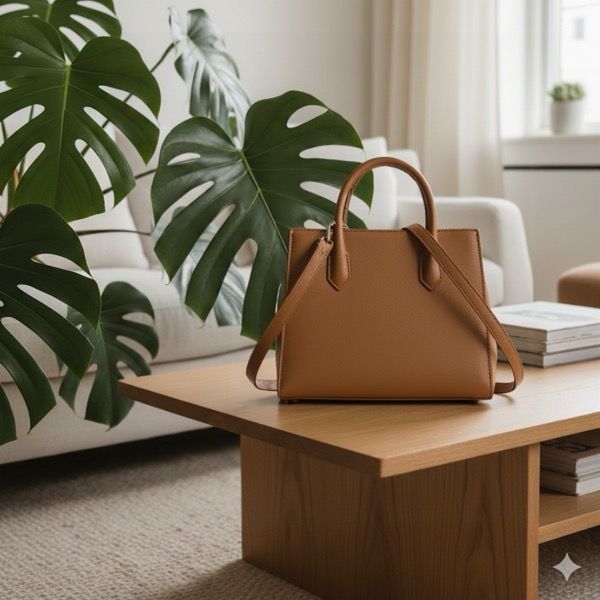} \\[0.2mm]
                \rotatebox[origin=c]{90}{\fontsize{4}{4.5}\selectfont\textbf{After}} & \occoutimg{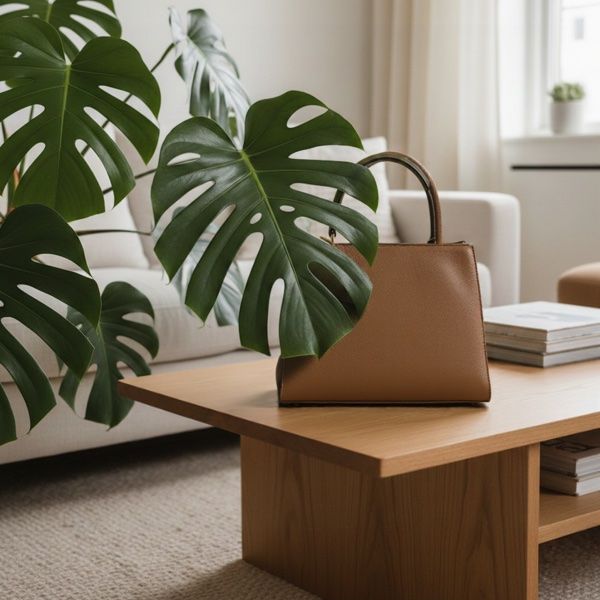} & \occoutimg{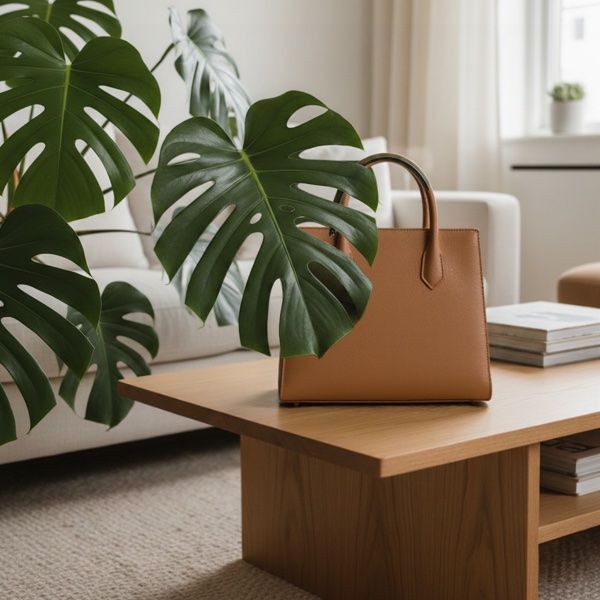} & \occoutimg{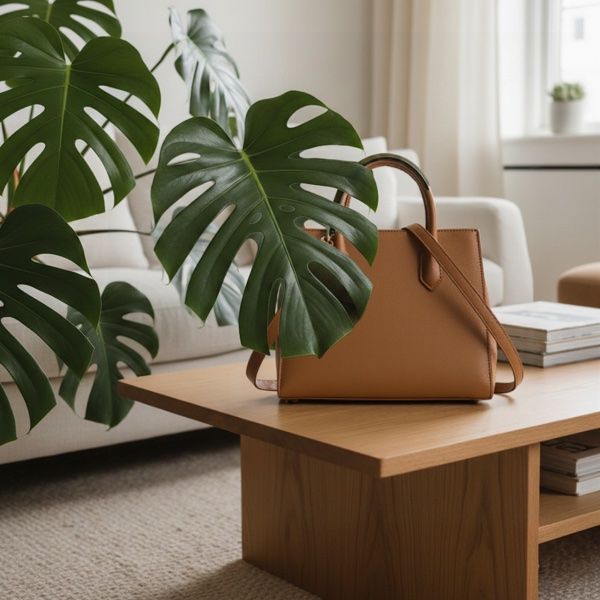} \\[0.2mm]
                \multicolumn{4}{c}{\fontsize{5.5}{6.5}\selectfont\textit{ObjectStitch}} \\[0.2mm]
                \rotatebox[origin=c]{90}{\fontsize{4}{4.5}\selectfont\textbf{Before}} & \occoutimg{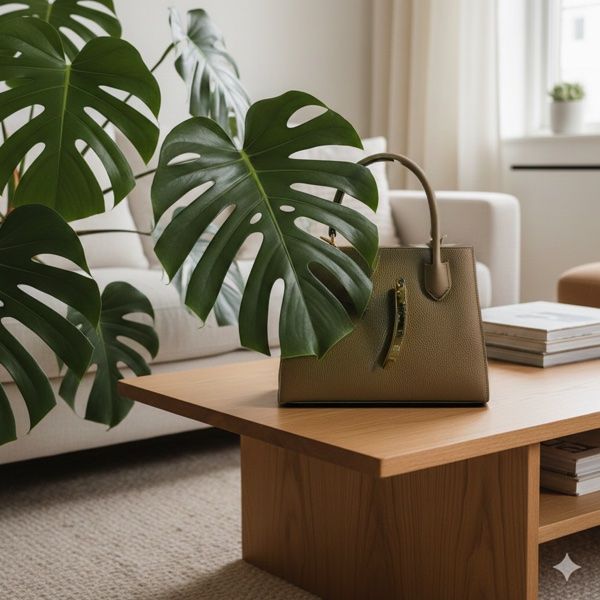} & \occoutimg{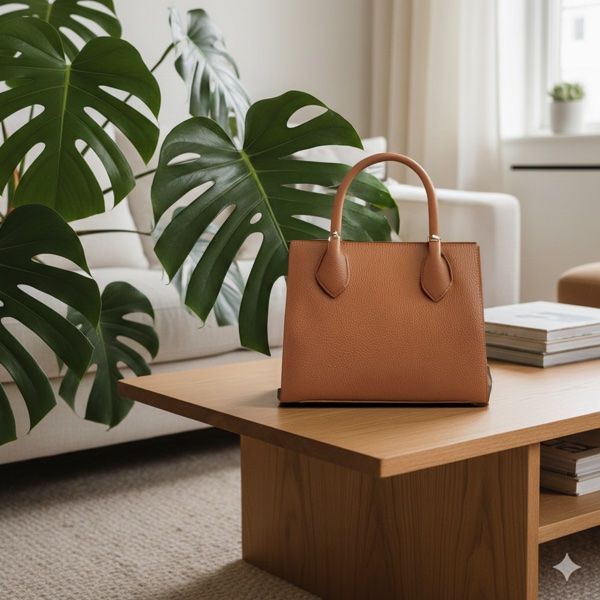} & \occoutimg{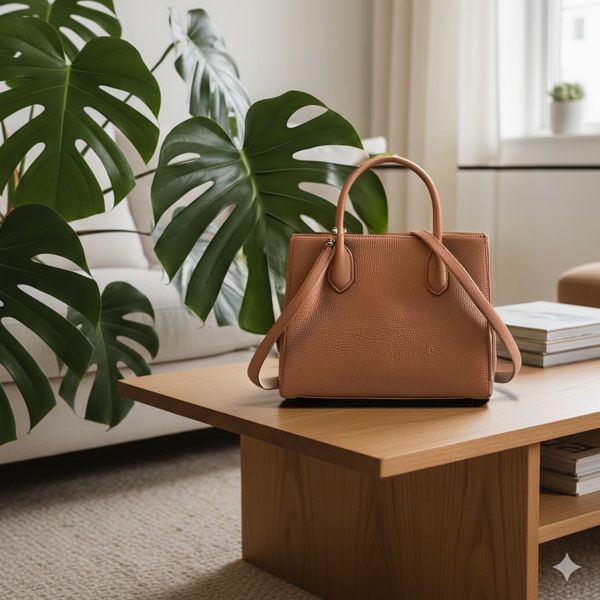} \\[0.2mm]
                \rotatebox[origin=c]{90}{\fontsize{4}{4.5}\selectfont\textbf{After}} & \occoutimg{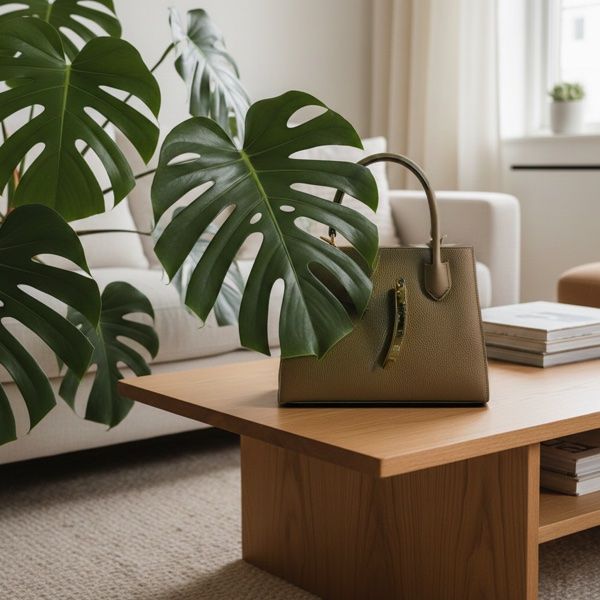} & \occoutimg{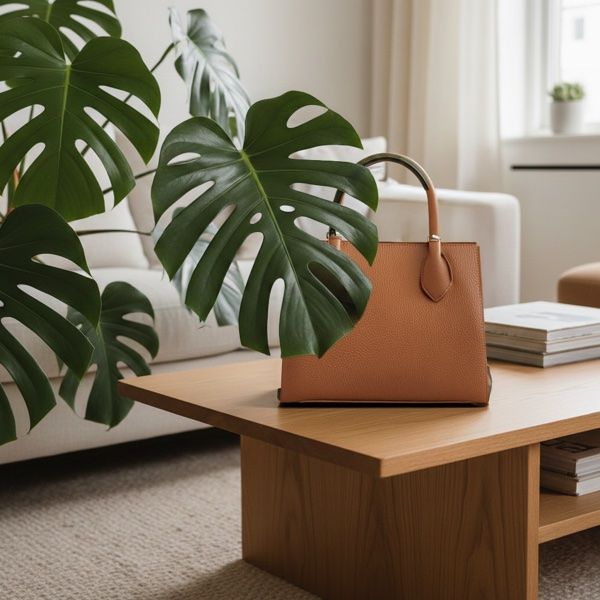} & \occoutimg{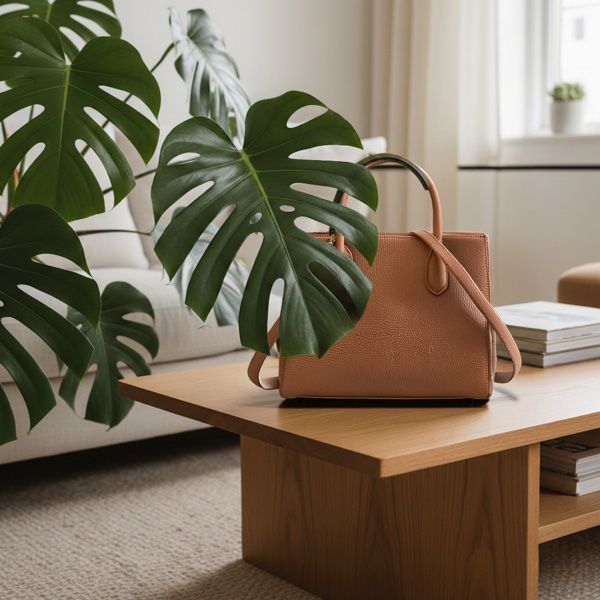} \\
            \end{tabular}
        \\
    \end{tabular}
    \caption{\textbf{Occlusion-aware hybrid restoration across models.} Left inputs: background and product. Mask overlay and occluder composite are visualizations of the occlusion regions. Right: Freeform, BBox, and Dim-Aware outcomes for InsertAnything and ObjectStitch, shown before and after occluder restoration. Our pipeline segments occluders, caches their pixels, inpaints them away to expose a clean background for compositing, and finally pastes the cached pixels back---preserving overlapping structures more faithfully than the corresponding before-restoration outputs.}
    \label{fig:occlusion_results}
\end{figure*}

\subsection{Experimental Setup}

\noindent\textbf{Evaluation Dataset.}
We curated \textbf{CatalogStitch-Eval}, a challenging evaluation dataset for catalog image compositing consisting of two subsets:
\begin{itemize}
    \item \textbf{Dimension Mismatch Set:} 35 image pairs where the replacement product has significantly different aspect ratio than the target region (e.g., tall lamp replacing wide table, square bag replacing rectangular clutch).
    \item \textbf{Occlusion Set:} 23 images featuring products partially occluded by 1--2 foreground elements (plants, vases, side tables, lamps).
\end{itemize}
The dataset package (metadata, pre-computed masks, method outputs, and HTML viewers) is publicly released\footnote{\url{https://github.com/adobe-research/CatalogStitch}}. Full qualitative results for all 35 dimension-aware and 23 occlusion examples are provided in the supplementary material (\texttt{additional\_results\_dimension\_aware.pdf} and \texttt{additional\_results\_occlusion.pdf}).

\noindent\textbf{Baseline Methods.}
We evaluate our techniques as preprocessing/postprocessing steps applied to three state-of-the-art compositing models:
\begin{itemize}
    \item \textbf{ObjectStitch}~\cite{song2023objectstitch}: Content adaptor-based approach for reference-guided inpainting.
    \item \textbf{OmniPaint}~\cite{yu2025omnipaint}: Unified insertion-removal inpainting framework with disentangled editing.
    \item \textbf{InsertAnything}~\cite{song2025insertanything}: In-context editing approach for reference-based object insertion via DiT.
\end{itemize}

For each baseline, we compare: (1) the original method with standard masks, and (2) the method enhanced with our CatalogStitch techniques.

\subsection{Qualitative Results}

\noindent\textbf{Dimension-Aware Mask Results.}
Figure~\ref{fig:dimension_results} shows representative high-mismatch examples across OmniPaint, InsertAnything, and ObjectStitch. The supplementary material contains all 35 benchmark cases for all three compositors.

Key observations:
\begin{itemize}
    \item Without dimension-aware mask adaptation, products are stretched, squashed, or cropped to fit the original target region.
    \item With the adapted mask, products retain their native proportions while staying aligned with the original scene layout.
    \item The same mask computation generalizes across all three baseline compositors, supporting a model-agnostic interface improvement rather than a model-specific fix.
\end{itemize}

\noindent\textbf{Occlusion Handling Results.}
Figure~\ref{fig:occlusion_results} shows representative occlusion-heavy scenes where foreground objects are visibly restored after compositing. The supplementary material provides before/after results for all 23 benchmark examples.

Key observations:
\begin{itemize}
    \item Without occlusion handling, foreground elements are destroyed, distorted, or inconsistently hallucinated.
    \item With hybrid restoration, overlapping objects recover their exact geometry, texture, and illumination because the original pixels are pasted back.
    \item The benefit is most visible for fine structures and hard boundaries that diffusion-based models struggle to reconstruct reliably.
\end{itemize}

\subsection{Quantitative Evaluation}

Table~\ref{tab:quantitative_main} reports quantitative results for the six method variants. Baselines use standard bbox masks; ``+ Ours'' uses dimension-aware masks (and for occlusion, exact-pixel restoration).
We evaluate using five metrics: \textbf{AR Error} measures aspect-ratio preservation of the inserted product (lower is better). \textbf{FID}~(Fr\'{e}chet Inception Distance) captures the distributional similarity between input-product and generated-object crops---lower values indicate more realistic composites. \textbf{CLIP-score} measures semantic alignment between the composited region and the reference product using CLIP embeddings (higher is better). \textbf{DINO-score} evaluates structural similarity via self-supervised DINO features, capturing fine-grained appearance fidelity (higher is better). \textbf{Occluder PSNR} measures pixel-level restoration quality over occluder regions (higher is better).

Across all baselines, adding our proposed components consistently improves every metric: AR Error drops from $\sim$30\% to $\sim$4--5\%, FID decreases substantially, and both CLIP and DINO scores increase. Among all configurations, \textbf{InsertAnything + Ours achieves the best results across all five metrics}, reaching the lowest FID (77.72), highest CLIP-score (92.68), highest DINO-score (88.30), lowest AR Error (3.92), and highest Occluder PSNR (27.54).

\noindent\textbf{Aspect Ratio Preservation.}
For the 35 dimension-mismatch examples, we measure:
\begin{equation}
    \text{AR Error} = \frac{|AR_{output} - AR_{input}|}{AR_{input}} \times 100\%
\end{equation}
where $AR_{output}$ is measured from the generated object region and $AR_{input}$ from the input product.

\noindent\textbf{Occluder Fidelity.}
For the 23 occlusion examples (InsertAnything/ObjectStitch), we report masked PSNR over restoration regions estimated by the before/after delta; higher values indicate stronger recovery toward the source scene.

\begin{table}[!tb]
    \centering
    \scriptsize
    \setlength{\tabcolsep}{3pt}
    \resizebox{\columnwidth}{!}{%
    \begin{tabular}{lccccc}
        \toprule
        Method & AR Error $\downarrow$ & Occluder PSNR $\uparrow$ & FID $\downarrow$ & CLIP-score $\uparrow$ & DINO-score $\uparrow$ \\
        \midrule
        OmniPaint & 31.07 & - & 137.15 & 83.03 & 68.50 \\
        \textbf{OmniPaint + Ours} & 4.57 & - & 135.79 & 86.11 & 72.99 \\
        ObjectStitch & 30.97 & 11.60 & 101.55 & 90.27 & 85.76 \\
        \textbf{ObjectStitch + Ours} & 5.05 & 26.84 & 91.52 & 90.62 & 88.09 \\
        InsertAnything & 29.98 & 13.33 & 105.99 & 90.23 & 82.63 \\
        \textbf{InsertAnything + Ours} & \textbf{3.92} & \textbf{27.54} & \textbf{77.72} & \textbf{92.68} & \textbf{88.30} \\
        \bottomrule
    \end{tabular}%
    }
    \caption{\textbf{Quantitative comparison on CatalogStitch-Eval.} AR Error, CLIP, and DINO scores are computed on the 35 dimension examples; Occluder PSNR on the 23 occlusion examples (-- for OmniPaint). CLIP and DINO are cosine similarity $\times 100$; FID is between input-product and generated-object crop distributions.}
    \label{tab:quantitative_main}
\end{table}

\subsection{Discussion}

\noindent\textbf{Cross-Model Consistency.}
Our techniques yield consistent improvements across all compositor architectures. AR Error drops from $\sim$30\% to $\sim$4--5\% uniformly across OmniPaint, InsertAnything, and ObjectStitch, confirming that dimension mismatch and occluder destruction are \textit{systematic} failure modes shared across generative pipelines rather than artifacts of any single architecture.

\noindent\textbf{Complementarity of Techniques.}
The dimension-aware and occlusion-aware modules address orthogonal failure modes and compose naturally. In the 23 occlusion examples, many also exhibit dimension mismatch; applying both modules in sequence yields the strongest results across every metric (Table~\ref{tab:quantitative_main}).

\noindent\textbf{Failure Cases.}
When aspect-ratio mismatch is very large, the expanded mask may cover important scene context, degrading background coherence. For occlusion restoration, translucent occluders may introduce subtle compositing seams at alpha boundaries. Under-segmentation of closely packed objects can also leave restore regions incomplete.


\section{Conclusion}
\label{sec:conclusion}

We presented CatalogStitch, a set of model-agnostic techniques that bridge the gap between research-grade object compositing methods and production catalog image generation requirements. Our dimension-aware mask computation algorithm addresses the dimension mismatch problem by automatically adapting target regions to accommodate products with different aspect ratios. Our occlusion-aware hybrid restoration method guarantees pixel-perfect preservation of foreground elements that would otherwise be destroyed during compositing.

By automating tedious manual corrections, our techniques transform the human-AI interaction model for compositing: users focus on high-level creative decisions (which product, which background) while the system handles low-level technical adjustments. This simplification enables non-expert users to leverage powerful generative models without specialized skills.

We demonstrated our techniques across three state-of-the-art compositing models (ObjectStitch, OmniPaint, InsertAnything), showing consistent improvements on challenging catalog scenarios involving dimension mismatches and occlusions.

A central insight of our work is that \textit{hybrid approaches}, combining generative AI for harmonization and shadow generation with deterministic methods for mask computation and pixel restoration, achieve both the flexibility of AI compositing and the reliability required for production quality. Fully generative approaches cannot guarantee the fidelity needed for professional catalog imagery.

\noindent\textbf{Limitations and Future Work.}
Our current approach assumes occluders are in front of the target product; handling complex multi-layer occlusion relationships is left for future work. The dimension-aware mask computation uses rectangular bounding boxes; exploring shape-aware mask adaptation could further improve results. Additionally, broader quantitative and human-preference evaluation across larger datasets and more compositors would strengthen the empirical validation.

{
    \small
    \bibliographystyle{ieeenat_fullname}
    \bibliography{main}
}

\end{document}